\PassOptionsToPackage{sort&compress}{natbib} 
\documentclass[review]{elsarticle}
\usepackage[utf8]{inputenc}

\usepackage[dvipsnames]{xcolor}
\usepackage{microtype}
\usepackage{graphicx}
\usepackage{booktabs} 

\makeatletter
 
\newskip\@bigflushglue \@bigflushglue = -100pt plus 1fil

\def\bigcentering{\let\\\@centercr\rightskip\@bigflushglue%
\leftskip\@bigflushglue
\parindent\z@\parfillskip\z@skip}

\makeatother

\usepackage[english]{babel} 
\usepackage{subfig}
\usepackage{tikz}
\usepackage{amsmath}
\usepackage{amssymb}
\usepackage{bbm, dsfont}
\usepackage{multicol}
\usepackage{stmaryrd}
\usepackage{multirow}
\usepackage{stfloats}
\usepackage{tabularx}
\usepackage{siunitx}
\usepackage{algorithm}
\usepackage{algorithmic}
\usepackage{xargs}
\usepackage{natbib}
\usepackage{bm}
\newtheorem{remark}{Remark}[section]
\usepackage{rotating}
\usepackage{hyperref}
\usepackage{mathtools}

\usepackage[normalem]{ulem}
\usepackage{soul}
\usepackage{cancel}

\newcommand{\algorithmicinput}{\textbf{input}}
\newcommand{\algorithmicoutput}{\textbf{output}}
\newcommand{\INPUT}{\item[\algorithmicinput]}
\newcommand{\OUTPUT}{\item[\algorithmicoutput]}

\usepackage[colorinlistoftodos,prependcaption,textsize=tiny]{todonotes}

\newcommand{\changecolor}{Black}
\newcommand{\change}[1]{{\color{\changecolor} {#1}}}
\newcommand{\remove}[1]{} 
\newcommand{\removeeq}[1]{} 

\newcommand{\changecolorb}{Black}
\newcommand{\changeb}[1]{{\color{\changecolorb} {#1}}}
\newcommand{\removeb}[1]{}
\newcommand{\removeeqb}[1]{}

\journal{Aerospace Science and Technology}

\title{Upper Trust Bound Feasibility Criterion for Mixed Constrained Bayesian Optimization with Application to Aircraft Design}
\author[on,isae]{R. Priem}
\author[on]{N. Bartoli}
\author[isae]{Y. Diouane}
\author[on,isae]{A. Sgueglia}
\address[on]{ONERA, DTIS, Université de Toulouse, Toulouse, France}
\address[isae]{ISAE-SUPAERO, Université de Toulouse, Toulouse, 31055 Cedex 4, France}

\begin{document}

\begin{frontmatter}
\begin{abstract}
Bayesian optimization methods have been successfully applied to \textit{black box} optimization problems that are \textit{expensive to evaluate}. In this paper, we adapt the so-called \textit{super efficient global optimization} algorithm to solve more accurately mixed constrained problems. The proposed approach handles constraints by means of \textit{upper trust bound}, the latter encourages exploration of the feasible domain by combining the mean prediction and the associated uncertainty function given by the Gaussian processes. On top of that, a refinement procedure, based on a learning rate criterion, is introduced to enhance the exploitation and exploration trade-off. We show the good potential of the approach on a set of numerical experiments. Finally, we present an application to conceptual aircraft configuration upon which we show the superiority of the proposed approach compared to a set of the state-of-the-art black box optimization solvers.
\end{abstract}

\end{frontmatter}


\section{Introduction}

In this paper, we are interested in the following mixed constrained optimization problem
\begin{equation}
    \min\limits_{\bm{x} \in \Omega} \left\{ f(\bm{x}) ~~\mbox{s.t.}~~ \bm{g}(\bm{x})\geq 0 ~\mbox{and}~ \bm{h}(\bm{x})=0 \right\}
    \label{eq:opt_prob}
\end{equation}
where $f:\mathbb{R}^d \mapsto \mathbb{R}$ is the objective function, \hbox{$\bm{g}:\mathbb{R}^d \mapsto \mathbb{R}^m$} gives the inequality constraints, and $\bm{h}:\mathbb{R}^d \mapsto \mathbb{R}^p$ returns the equality constraints.
The design space $\Omega \subset \mathbb{R}^d$ is a bounded domain.
The functions $f$, $\bm{g}$, and $\bm{h}$ are typically simulations which possess no exploitable properties such as derivatives (i.e., black box).
They take also a long time to evaluate.
In some cases, these functions may be multimodal and the resulting feasible domain more complex to define.

Many practical optimization problems are in a black box form, expensive to evaluate and present mixed multimodal constraints.
For instance, for \textit{multidisciplinary design optimization} (MDO) problems \cite{RaymerAircraftDesignConceptual2018}, one has to conceive the best product regarding specific performances.
The design domain is restricted by several mixed constraints coming from the requirements, the different disciplines involved (e.g., structure, aerodynamic and propulsion in aircraft design) and their interconnections.
In this context, \textit{Bayesian optimization} (BO) is a powerful strategy for solving problem~(\ref{eq:opt_prob}).

Most of the work on BO \cite{frazier2018tutorial,ShahriariTakingHumanOut2016,WangMaxvalueentropysearch2017} focuses on unconstrained black box optimization problems using a sequential enrichment of surrogate models.
In fact, using \textit{Gaussian processes} (GPs) \cite{Krigestatisticalapproachbasic1951,RasmussenGaussianprocessesmachine2006} to define response surfaces, the sequential enrichment is performed by maximizing a given acquisition function \cite{frazier2018tutorial,ShahriariTakingHumanOut2016}. 
The latter is meant to model a compromise between exploration of new zones in the design space and exploitation (i.e. minimization) of the GPs.

Extensions of the BO framework have been developed to handle constraints \change{\cite{frazier2018tutorial,ShahriariTakingHumanOut2016,FeliotBayesianapproachconstrained2017,gelbartBayesianOptimizationUnknown2014a,lethamConstrainedBayesianOptimization2019}}.
In this case, both the objective and the constraints functions are modeled with GPs, and an optimization sub-problem based on the infill criterion leads the enrichment process. Existing constrained BO methods can be split into two main categories.
The first one, which is the largest, addresses only inequality constrained problems \change{\cite{frazier2018tutorial,ShahriariTakingHumanOut2016,FeliotBayesianapproachconstrained2017,gelbartBayesianOptimizationUnknown2014a,lethamConstrainedBayesianOptimization2019}} and the references therein.
We note that most of the existing works recommend to handle equality constraints by transformation of the initial optimization problem.
For instance, equality constraints of the type $\bm{h}(\bm{x})=0$ are changed into two inequality constraints of the type $\bm{h}(\bm{x})\ge 0$ and $-\bm{h}(\bm{x})\ge 0$.
In general, such transformation turns out to be harmful as it increases the number of constraints and also introduces antagonist requirements leading to constraint qualification issues \cite{NocedalNumericaloptimization2006}.
Alternatively, other approaches proposed to address mixed constrained problems without any transformation \cite{SasenaExplorationmetamodelingsampling2002,PichenyBayesianoptimizationmixed2016}.
\remove{To the best of our knowledge, none of these approaches is directly tackling mixed general constrained problems.}

ALBO \cite{PichenyBayesianoptimizationmixed2016} is the state-of-the-art solver in BO to handle mixed constrained problems.
It combines an unconstrained BO framework with the classical \textit{augmented Lagrangian} (AL) framework \cite{NocedalNumericaloptimization2006}.
ALBO was originally designed for the equality constraints problems \cite{GramacyModelingaugmentedLagrangian2016} and then extended to inequality constraints by means of the slack variables.
The ALBO procedure is the same as the AL framework except that the minimization of the AL function is replaced by the maximization of an acquisition function.
The new acquisition function is not given explicitly but only through an estimation method.
Despite of the introduced effort in reducing the computational cost of the acquisition function estimation, the ALBO process is still not adapted to solve large scale problems in a reasonable time.

The \textit{super efficient global optimization} (SEGO) framework \cite{SasenaExplorationmetamodelingsampling2002} is an extension of the well-known unconstrained efficient global optimization framework \cite{JonesEfficientglobaloptimization1998} to handle mixed constrained optimization problems.
The SEGO enrichment process is led by a constrained optimization sub-problem where the objective function is given by an acquisition function and the constraints by the GPs mean predictions of the constraints functions (without inclusion of the uncertainties provided by the GPs).
By doing so, the constraints can be badly approximated by the GPs and the optimization can be misled especially when  constraints are hard to approximate.
The main advantage of the SEGO framework, compared to others, is related to the fact that it scales well when solving large scale constrained optimization problems, e.g., \citet{BouhlelEfficientglobaloptimization2018}.

To tackle the issue of badly modeled  constraints, including the uncertainties (provided by the GP models) has been shown to be very useful \cite{lam2015multifidelity,lam2017lookahead,priem2019use}.
For instance, the authors in \cite{lam2015multifidelity,lam2017lookahead} introduced a scalar fixed \textit{upper trust bound} (UTB) to handle the constraints. Their proposed approach was designed to allow the exploration of a larger feasible domain (but relaxed) rather than being restricted to a small feasible one.
Recently, unlike in \cite{lam2015multifidelity,lam2017lookahead} where only a fixed scalar UTB was investigated, \citet{priem2019use} used a dynamic adaptive strategy for updating the UTB during the optimization process.
The main idea was to include uncertainties on GPs but only during specific stages of the optimization procedure.
This led to a good compromise between exploration (of the design space) and exploitation (i.e., minimization of the objective function).
It provided encouraging results on difficult optimization toy problem.
We stress that all the approaches \cite{lam2015multifidelity,lam2017lookahead,priem2019use} were designed to handle only inequality constraints and thus are not adapted to solve mixed constrained optimization problems of the form~(\ref{eq:opt_prob}).
We note also that all the works \cite{lam2015multifidelity,lam2017lookahead,priem2019use} were only validated using toy problems.
Thus, confirming the potential of all these approaches using extensive and practical numerical tests can be very useful.

In this paper, in the context of the SEGO framework, we propose to improve the existing constraints handling strategies \cite{lam2015multifidelity,priem2019use} by (a) including equality constraints, (b) using a better adaptive mechanism for the update of the UTB during the optimization process, and (c) performing extensive and practical numerical tests on a large test set of problems.
In our proposed approach, the UTB is controlled using a learning rate vector that helps in managing the trade-off between exploration of badly modeled domain and exploitation of the known feasible domain predicted by the surrogate models.
In fact, the extended version of SEGO for equality constraints using the \textit{upper trust bound}, called SEGO-UTB, is shown to ensure a better exploration of the entire feasible domain.
Its superiority, compared to SEGO and other solvers, is confirmed on 29 mixed constrained problems using different test strategies including data profiles.
Finally, SEGO-UTB is applied to solve an MDO problem where the goal is to  optimize a ``tube \& wing" hybrid aircraft configuration with a distributed electric propulsion \cite{sguegliafasthybrid2018}.

The outline of the paper is as follows.
In Section~\ref{sec:rev_sego}, a detailed review of the constrained Bayesian optimization framework is given.
The adaptive UTB \remove{on the constraints} as well as different constraints learning rate strategies are given in Section~\ref{sec:utb}.
Section~\ref{sec:test} presents our academical tests. 
The MDO test case is commented in Section~\ref{sec:FAST}.
Conclusions and perspectives are finally drawn in Section~\ref{sec:clc}.

\section{Bayesian optimization and the SEGO framework}
\label{sec:rev_sego}

\subsection{Bayesian optimization framework}
Starting from an initial \textit{design of experiments} (DoE) using a first set of $l$ sample points chosen in the design domain $\Omega$, constrained BO framework builds surrogate models using GPs~\cite{RasmussenGaussianprocessesmachine2006,Krigestatisticalapproachbasic1951} of the objective $f$, inequality $\bm{g}$ and equality $\bm{h}$ constraints functions.
The surrogate models are then iteratively enriched in order to locate the optimum of the constrained optimization problem.
The search strategy balances the exploration of the design space $\Omega$ and the exploitation of the surrogate models by solving a maximization mixed-constrained sub-problem.
The objective function of the sub-problem is expressed by an acquisition function while the constraints use GPs to replace $\bm{g}$ and $\bm{h}$.  
Solving the sub-problem is assumed to be computationally inexpensive as one uses only GPs information.
Iteratively, the solution of the sub-problem is evaluated on $f$, $\bm{g}$, $\bm{h}$, and added to the respective DoE.
The same process is repeated until a maximum number of iterations is reached. The main steps of the BO framework, when applied to problem~\eqref{eq:opt_prob}, are summarized by Algorithm~\ref{alg:BO}.
\begin{algorithm}[ht!]
     \begin{algorithmic}[1]
        \INPUT{: Objective and constraints functions, initial DoEs for objective and constraints, a maximum number of iterations max\_nb\_it\;}
        \FOR{\change{$l = 0$} \TO \change{\mbox{max\_nb\_it} - 1}}
            \STATE {Build the surrogate models using GPs\;}
            \STATE {Find $\bm{x}^{(l+1)}$ a solution of the enrichment maximization sub-problem\;}
            \STATE {Evaluate the objective and constraints functions at $\bm{x}^{(l+1)}$\;}
            \STATE {Update the DoE\;}
        \ENDFOR
        \OUTPUT{: The best point found in the DoE\;}
    \end{algorithmic}
    \caption{The Bayesian optimization framework.}
    \label{alg:BO}
\end{algorithm}
The next two subsections describe \change{the information provided by the GPs} as well as the maximization sub-problem choices within the constrained BO framework.

\subsection{Gaussian Process}

Scalar \change{output} GPs \cite{RasmussenGaussianprocessesmachine2006} are fully defined by a mean function $\mu$ and a standard deviation function $\sigma$. 
The mean function describes the global behaviour of the GP whereas the standard deviation function depicts the GP uncertainty of each sample on the entire domain.

A \remove{detailed} description of GPs can be as follows.
Let $s: \mathbb{R}^d \mapsto \mathbb{R}$ be a scalar function for which a GP is built using a DoE of $l$ points $\mathcal{D}_s^{(l)}=\{\bm{x}^{(k)},y_{s}^{(k)}\}_{k = 1, \ldots, l}$ where $\bm{x}^{(k)} \in \Omega$ and $y_{s}^{(k)}=s(\bm{x}^{(k)}) \in \mathbb{R}$.
For clarity reasons, in the context of our optimization problem \eqref{eq:opt_prob}, $s$ can represent the objective function (i.e., $s=f$) or a given component constraint function (i.e., $s=g_j$ or $h_j$ for a given constraint component $j$).

The GP model related to $s$ using $l$ sample points is a family of functions defined by a mean function $\mu_s^{(l)}$ and a standard deviation $\sigma_s^{(l)}$.
Namely, at each point $\bm{x}$ of the bounded domain $\Omega$, the GP of $s$ is defined with a multivariate Gaussian distribution $\mathcal{N}(\mu_s^{(l)}(\bm{x}), \sigma_s^{(l)}(\bm{x}))$.

\change{Note that the mean $\mu_s^{(l)}$ and $\sigma_s^{(l)}$ are computed thanks to a correlation function chosen by the user. 
In fact,}\remove{We note that} the definition of the \change{correlation function} \remove{kernel function $k$} depends on a set of hyper-parameters that are in general estimated by maximizing a likelihood function. 
Unfortunately, such maximization can be computationally challenging for large scale functions or with a large DoE.
Practical approaches to estimate the hyper-parameters can be found in \cite{KandasamyHighdimensionalBayesian2015,BouhlelEfficientglobaloptimization2018}.

\subsection{The enrichment optimization sub-problem}

The BO framework combines the surrogate models provided by GPs and the enrichment strategy driven by the maximization of the sub-problem.
In fact, for a given iteration $l$, the GPs of the objective $f$ and each component of the constraints $g$, $h$ are built using the current DoE ($\mathcal{D}^{(l)}_f$,  $\mathcal{D}^{(l)}_{g_i}$, $\mathcal{D}^{(l)}_{h_j}$) for $i= 1,\ldots, m$ and $j= 1,\ldots,p$.
The corresponding GPs mean and standard deviation functions are $\mu^{(l)}_f$, $\mu^{(l)}_{g_i}$, $\mu^{(l)}_{h_j}$, $\sigma^{(l)}_f$, $\sigma^{(l)}_{g_i}$ and $\sigma^{(l)}_{h_j}$ for $i= 1,\ldots, m$ and $j= 1,\ldots,p$.
For clarity reasons, we will use the following vector notations for inequality constraints $\bm{g}$, i.e., $\bm{\mu}^{(l)}_{\bm{g}}=[\mu^{(l)}_{g_1}, \ldots, \mu^{(l)}_{g_m} ]^{\top} \in \mathbb{R}^m$, $\bm{\sigma}^{(l)}_{\bm{g}}=[\mu^{(l)}_{g_1}, \ldots, \sigma^{(l)}_{g_m} ]^{\top} \in \mathbb{R}^m$, and similarly for equality constraints $\bm{h}$, i.e., $\bm{\mu}^{(l)}_{\bm{h}}=[\mu^{(l)}_{h_1}, \ldots, \mu^{(l)}_{h_p} ]^{\top} \in \mathbb{R}^p$, $\bm{\sigma}^{(l)}_{\bm{h}}=[\mu^{(l)}_{h_1}, \ldots, \sigma^{(l)}_{h_p} ]^{\top} \in \mathbb{R}^p$.

Originally, an unconstrained BO framework is led by an acquisition function $\alpha^{(l)}_f$ modeling the trade-off between exploration of new areas in the design space (i.e., areas with high value of $\sigma^{(l)}_f$) and exploitation (i.e., minimization of  $\mu^{(l)}_f$).
The most promising point is given by maximization of the acquisition function:
\begin{equation*}
    \bm{x}^{(l+1)} = \arg \max\limits_{\bm{x}\in \Omega} \alpha_f^{(l)}(\bm{x}).
\end{equation*}
The acquisition functions, that have been developed, are either explicit or implicit \cite{frazier2018tutorial,ShahriariTakingHumanOut2016,WangMaxvalueentropysearch2017,Bartoliadaptivemodeling2019}. The computational cost of the implicit ones forbids their use for large scale optimization problems.

In the context of constrained optimization problems, for a given outer iteration $l$, the BO framework has been extended in two following ways.
The first way is by using a merit type function $\alpha^{(l)}_m : \mathbb{R}^d \mapsto \mathbb{R}$ where one combines the objective and the constraints.
The new enrichment point is thus computed by maximizing the merit function $\alpha^{(l)}_m$ on the design space $\Omega$, i.e.,
\begin{equation*}
    \bm{x}^{(l+1)} = \arg \max\limits_{\bm{x}\in \Omega} \alpha_m^{(l)}(\bm{x}).
\end{equation*}
Several methods based of this merit function have been proposed in the literature \changeb{\cite{frazier2018tutorial,ShahriariTakingHumanOut2016, lethamConstrainedBayesianOptimization2019, gelbartBayesianOptimizationUnknown2014a}}, ALBO is belonging to this class of methods.

The second way of handling constraints, in the context of BO, consists in solving a mixed constrained maximization sub-problem, i.e.,
\begin{equation}
    \bm{x}^{(l+1)} = \arg \max\limits_{\bm{x} \in \Omega_{\bm{g}}^{(l)} \cap \Omega_{\bm{h}}^{(l)}} \alpha_f^{(l)}(\bm{x}),
    \label{eq:inner_loop_segre}
\end{equation}
where $\alpha^{(l)}_f: \mathbb{R}^d \mapsto \mathbb{R}$ is a given acquisition function related to the objective function $f$ (similarly to the unconstrained case), $\Omega_{\bm{g}}^{(l)}$ and $\Omega_{\bm{h}}^{(l)}$ are respectively the approximated feasible domains defined by the feasibility criteria $\bm{\alpha}_{\bm{g}}^{(l)}:  \mathbb{R}^d \mapsto \mathbb{R}^m$ and $\bm{\alpha}_{\bm{h}}^{(l)}: \mathbb{R}^d \mapsto \mathbb{R}^p$. We note that the feasibility criteria $\bm{\alpha}_{\bm{g}}^{(l)}$ and $\bm{\alpha}_{\bm{h}}^{(l)}$ are not necessarily of the same form as $\bm{g}$ and $\bm{h}$.
For instance, equality constraints can be expressed as inequality approximated constraints, in this case, the approximated feasible domain $\Omega_{\bm{h}}^{(l)}$ related to the constraints $\bm{h}$ will be of the form $\{\bm{x} \in \Omega ~;~ \bm{\alpha}_{\bm{h}}^{(l)}(\bm{x}) \geq 0\}$.

To the best of our knowledge, a multitude of existing approaches \cite{frazier2018tutorial,ShahriariTakingHumanOut2016,SasenaExplorationmetamodelingsampling2002,lam2015multifidelity,lam2017lookahead,priem2019use} uses a constrained maximization sub-problem of the form (\ref{eq:inner_loop_segre}), but only SEGO \cite{SasenaExplorationmetamodelingsampling2002} was designed to solve mixed constrained optimization problems.
In this context, SEGO sets the feasibility criteria functions $\bm{\alpha}_{\bm{g}}^{(l)}$ and $\bm{\alpha}_{\bm{h}}^{(l)}$  to be  equal to the prediction of the GP models of the constraints $\bm{\mu_g}^{(l)}$ and $\bm{\mu_h}^{(l)}$ on the following way  $\Omega_{\bm{g}}^{(l)} = \{\bm{x}\in \Omega ~;~ \bm{\mu_g}^{(l)}(x) \geq 0 \}$ and $\Omega_{\bm{h}}^{(l)} = \{\bm{x}\in \Omega ~;~ \bm{\mu_h}(x) = 0\}$.
For SEGO, by using  only the mean functions of the GPs to model the constraints, all the functions involved in the mixed constrained maximization sub-problem are explicit and computationally inexpensive. \change{On the contrary, the} implicit methods, where typically Monte-Carlo estimators are used for each evaluation of the acquisition functions\change{, are expensive to compute.}

Despite the good results that SEGO has shown \cite{SasenaExplorationmetamodelingsampling2002,Bartoliadaptivemodeling2019}, the use of only the mean functions of the GPs to model the constraints of the sub-problem (\ref{eq:inner_loop_segre}) can mislead the optimization process and impact it badly.
In fact, during its early stage, the DoE is still poor as it does not provide enough information to build accurate GPs.
Due to the large uncertainties on the GPs defining $\bm{g}$ and $\bm{h}$, using only the mean functions $\bm{\mu_g}^{(l)}$ and $\bm{\mu_h}^{(l)}$, may consider that most of the design space is unfeasible.
Hence a large part of the feasible domain may not be explored.
In this case, the enrichment process gets very local and may ignore other feasible areas of the design space.

\begin{figure*}[htb!]
    \centering
    \subfloat[SEGO \label{fig:mc_feas_dom}]{\includegraphics[width=0.45\textwidth]{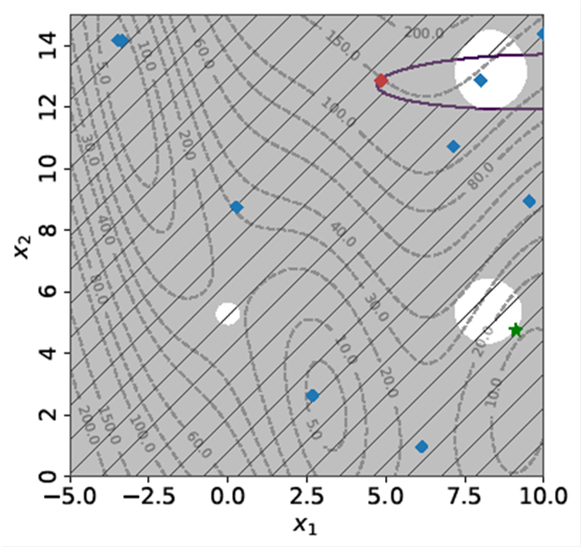}}
    \subfloat[SEGO-UTB\label{fig:utb_feas_dom}]{\includegraphics[width=0.45\textwidth]{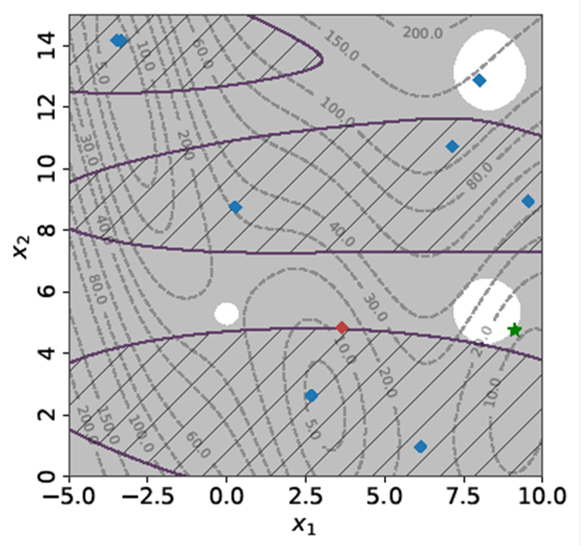}}
    \caption{A representation of the SEGO/SEGO-UTB  feasible domain (as predicted at a given iteration) for the modified Branin problem. The hatched area is the SEGO/SEGO-UTB unfeasible domain, the grey area shows the true unfeasible domain and the dashed curves are the contour plots of the objective function. The blue squares represent the current DoE whereas the green star indicates the global minimum of the problem. The red square is the new point to add in the DoE.}
    \label{fig:feas_dom}
\end{figure*}

As example, in Figure~\ref{fig:mc_feas_dom}, we show the feasible domain (the true and the predicted one) and the contour plots of the objective function for the modified Branin problem \cite{ParrReviewefficientsurrogate2010}.
The true unfeasible domain is represented with the grey color (the feasible domain is formed by three disjoint balls) and the predicted unfeasible domain at the first iteration is represented by the hatched area.
One can \change{clearly} see that the feasible area, as predicted by SEGO, is not covering two of the three true feasible domains (white areas).
Due to that, during the maximization sub-problem, SEGO will only provide enrichment points within the predicted feasible domain.
Consequently, the true feasible domain, where the global optimum is located, will never be explored by SEGO.
To overcome this issue, we will introduce in the next section a new feasibility criterion that explores more efficiently the design domain whenever the provided GPs for the constraints are inaccurate.

\section{Mixed constrained BO by using upper trust bounds}
\label{sec:utb} 

\subsection{On the use of upper trust bounds for constraints estimation}

At the end of Section \ref{sec:rev_sego}, we explain that SEGO may mislead the optimization process. \change{This is} due to a lack of accuracy of the constraints GPs, as it only uses the mean of the predicted values by the GPs without taking into account the level of accuracy associated to such estimation.

In the context of unconstrained BO optimization, a combination of both functions $\mu^{(l)}_s$ and $\sigma^{(l)}_s$ was used to provide an \textit{upper confidence bound} (UCB) \cite{SrinivasGaussianprocessoptimization2010} on the acquisition function, and shown to lead to a more robust model approximation for the objective function (up to a confidence level).
Similarly, we mimic the UCB strategy to better estimate the constraints within the SEGO framework.
The key idea is as follows: for a given scalar function $s$, using a learning rate $\tau_s^{(l)} \geq 0$, the functions $\mu^{(l)}_s \pm \tau_s^{(l)}\sigma^{(l)}_s$ approximate the targeted function $s$ with a trust level that is related to $\tau_s^{(l)}$.
For instance, when $\tau_s^{(l)}=3$ the trust interval can be expressed with $99\%$ \changeb{confidence for a single point $x$ sampled in $\Omega$}.
Outside of this zone, the value cannot be trusted as a reliable sample of the GP.
Based on this observation, we will try to model the constraints ($\bm{h}$ and $\bm{g}$) in a more robust way by introducing  an \textit{upper trust bound} (UTB) on the constraints.
The proposed UTB mechanisms depend on the nature of the regarded constraint (i.e., equality or inequality).

Concerning the inequality constraints $\bm{g}$, the UTB mechanism, by means of $\bm{\alpha_g}^{(l)}$, tries to relax the predicted feasible domain so that it includes the true feasible domain with a high probability.
The given trust level is governed by $\bm{\tau}^{(l)}_{\bm{g}}=[\tau^{(l)}_{g_1}, \ldots, \tau^{(l)}_{g_m} ]^{\top} \in \mathbb{R}^m_{+}$, namely, one has
\begin{eqnarray}
    \label{eq:UTB_g}
        \Omega_g^{(l)} &= & ~  \left \{ \bm{x} \in \Omega ~;~ \bm{\mu}^{(l)}_{\bm{g}}(\bm{x}) + \bm{\tau}^{(l)}_{\bm{g}}\change{\odot}\bm{\sigma}^{(l)}_{\bm{g}}(\bm{x}) \ge 0 \right \}
\end{eqnarray}
where the operator ``$\change{\odot}$'' denotes the element-wise multiplication.

Figure~\ref{fig:utb_feas_dom} shows the trusted feasible zone at 99\% \changeb{for a single point $\bm{x}$ sampled in $\Omega$} (that can be found with a learning rate equals to 3) of the modified Branin problem.
Compared to SEGO, see Figure~\ref{fig:mc_feas_dom}, the use of UTB, in the constraints formulation of the SEGO maximization sub-problem, leads to a bigger predicted feasible domain that, here, includes all the true feasible areas.
This allows a better exploration of the feasible domain and hence finding the global optimum. 

For the equality constraints $\bm{h}$, the UTB feasibility criterion is less straightforward.
It expresses the best constraints approximation (with the smallest violation) within the trusted domain delimited by the vectors $\bm{\mu}^{(l)}_{\bm{h}}(\bm{x}) - \bm{\tau_h}^{(l)} \change{\odot} \bm{\sigma}^{(l)}_{\bm{h}}$ and $\bm{\mu}^{(l)}_{\bm{h}}(\bm{x}) + \bm{\tau_h}^{(l)}\change{\odot}\bm{\sigma}^{(l)}_{\bm{h}}$ where \hbox{$\bm{\tau_h}^{(l)}=[\tau^{(l)}_{h_1}, \ldots, \tau^{(l)}_{h_p}]^{\top} \in \mathbb{R}^p_{+}$} is the related trust level. In this context, 
for each $i=1,\ldots,p$, the approximated feasible domain is given by $\alpha_{h_i}^{(l)} \change{=} \tau_{h_i}^{(l)}\sigma_{h_i}^{(l)}-|\mu_{h_i}^{(l)}|\ge 0$, meaning that we allow to violate the approximated equality constraint  $\mu_{h_i}^{(l)}$ up to the confidence level $\tau_{h_i}^{(l)}\sigma_{h_i}^{(l)}$.
In other words, the equality constraints are approximated by a set of inequality constraints as follows:
\begin{eqnarray}
    \label{eq:UTB_h}
        \Omega_h^{(l)} &= & ~  \left \{ \bm{x} \in \Omega ~;~ \bm{\tau_h}^{(l)}\change{\odot}\bm{\sigma_h}^{(l)}(\bm{x}) - \left| \bm{\mu_h}^{(l)}(\bm{x}) \right| \ge 0 \right \}.
\end{eqnarray}
\begin{remark}
It is possible to express $\Omega_h^{(l)}$ using only equality constraints.
In fact, for a given $\bm{x} \in \Omega$ and an equality constraint $h_i$, one can set $\alpha_{h_i}^{(l)}(\bm{x})$ such that whenever $\mu_{h_i}^{(l)}(\bm{x}) + \tau_{h_i}^{(l)}\sigma_{h_i}^{(l)}(\bm{x}) \leq 0$,  $\alpha_{h_i}^{(l)}(\bm{x})$ is set to $\mu_{h_i}^{(l)}(\bm{x}) + \tau_{h_i}^{(l)}\sigma_{h_i}^{(l)}(\bm{x})$, if $\mu_{h_i}^{(l)}(\bm{x}) - \tau_{h_i}^{(l)}\sigma_{h_i}^{(l)}(\bm{x}) \geq 0$, then $\alpha_{h_i}^{(l)}(\bm{x})=\mu_{h_i}^{(l)}(\bm{x}) - \tau_{h_i}^{(l)}\sigma_{h_i}^{(l)}(\bm{x})$, and otherwise $\alpha_{h_i}^{(l)}(\bm{x})$ is set to $0$. I.e.,
{\small $$
        \Omega_h^{(l)}=  \left \{ \bm{x} \in \Omega ~;~     \max \left[ [\bm{\mu}^{(l)}_{\bm{h}}(\bm{x}) + \bm{\tau}^{(l)}_{\bm{h}}\change{\odot}\bm{\sigma}^{(l)}_{\bm{h}}(\bm{x})]^- ,\right. 
    \left. \bm{\mu}^{(l)}_{\bm{h}}(\bm{x}) - \bm{\tau}^{(l)}_{\bm{h}}\change{\odot}\bm{\sigma}^{(l)}_{\bm{h}}(\bm{x}) \right]=  0 \right \},
$$
}
where we use $[s]^-$ to denote the element-wise operation $\min \left(0,s\right)$ and the $\max$ for the element-wise maximum operator.
In our preliminary tests, the obtained results with this choice of $ \Omega_h^{(l)}$ turn to be less competitive compared to the use of the definition given by (\ref{eq:UTB_h}).
\end{remark}

\begin{remark}
Using the UTB criterion, the obtained maximization sub-problem can be seen as a generalization of the original SEGO sub-problem formulation.
In fact, by setting $\bm{\tau_g}^{(l)}$ and $\bm{\tau_h}^{(l)}$ to zero, one gets exactly the SEGO maximization sub-problem \eqref{eq:inner_loop_segre}. 
\end{remark}

The use of the UTB feasibility criterion is meant to enlarge the regarded feasible domain during the first stages of the optimization process (where the size of the DoE is still small and the uncertainties are large).
In what comes next, we will use SEGO-UTB to denote the SEGO framework when the UTB feasibility criterion is used in the constraints formulation of the maximization sub-problem. \change{The full description of the SEGO-UTB framework is given in Algorithm \ref{alg:SEGO_UTB}}. 
\begin{algorithm}[ht!]
    \change{
     \begin{algorithmic}[1]
        \INPUT{: Objective and constraints functions, initial DoEs for objective and constraints, a maximum number of iterations max\_nb\_it and an evolution strategy for the constraints learning rate.\;}
        \FOR{$l = 0$ \TO \mbox{max\_nb\_it} - 1}
            \STATE {Build the surrogate models using GPs.\;}
            \STATE {Set $$\bm{x}^{(l+1)}= \arg \max_{\bm{x} \in \Omega_h^{(l)} \cap \Omega_g^{(l)}} \alpha_f^{(l)}(x),$$
            where the expression of $\alpha_f^{(l)}$ is given in \ref{eq:wb2s}, $\Omega_g^{(l)}$ and $\Omega_h^{(l)}$ are given by \eqref{eq:UTB_g} and \eqref{eq:UTB_h}, respectively. \;}
            \STATE {Evaluate the objective and constraints functions at $\bm{x}^{(l+1)}$.\;}
            \STATE {Update the DoE.\;}
        \ENDFOR
        \OUTPUT{: The best point found in the DoE\;}
    \end{algorithmic}}
    \caption{The SEGO-UTB framework.}
    \label{alg:SEGO_UTB}
\end{algorithm}

\begin{figure}[htb!]
    \centering
    \subfloat[SEGO ($1^{st}$ iter.) \label{fig:mc-utb:sego1}]{\includegraphics[width=0.33\textwidth]{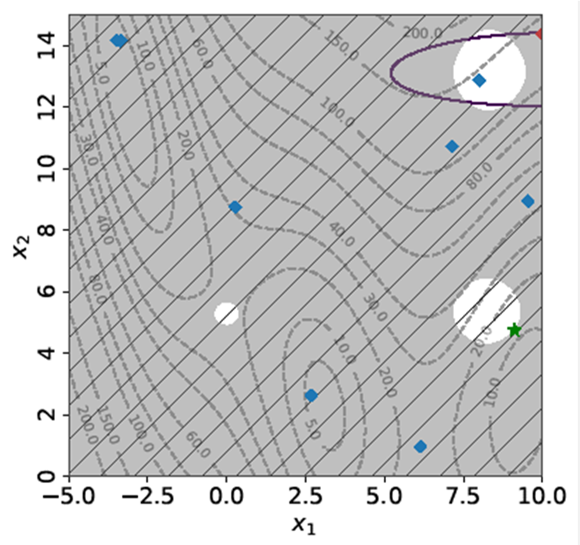}}
    \subfloat[SEGO ($2^{nd}$ iter.) \label{fig:mc-utb:sego2}]{\includegraphics[width=0.33\textwidth]{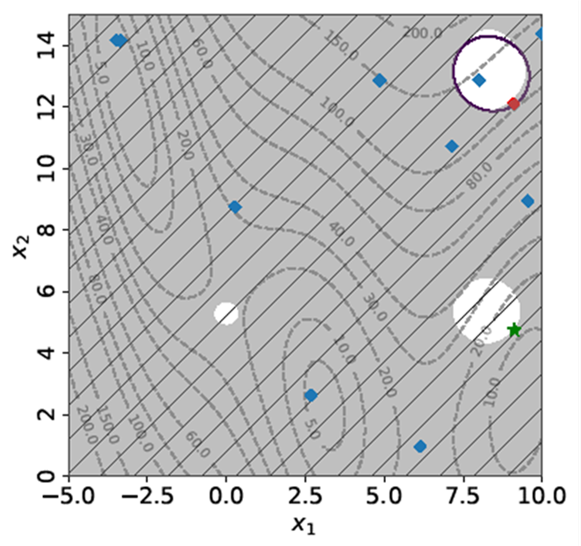}}
    \subfloat[SEGO (last iter.) \label{fig:mc-utb:segol}]{\includegraphics[width=0.33\textwidth]{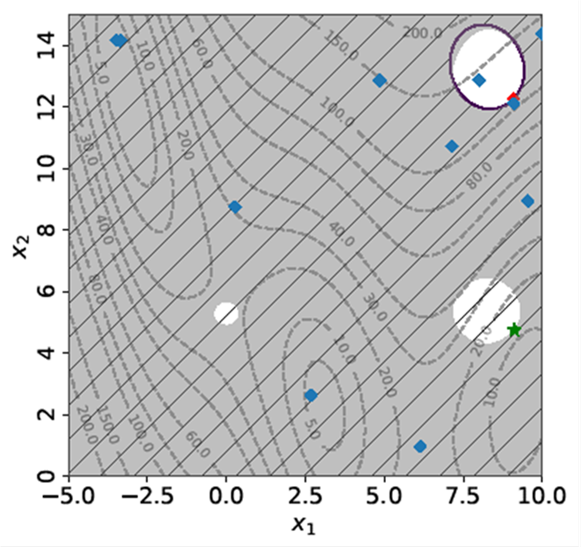}} \\
    \subfloat[SEGO-UTB ($1^{st}$ iter.)]{\includegraphics[width=0.33\textwidth]{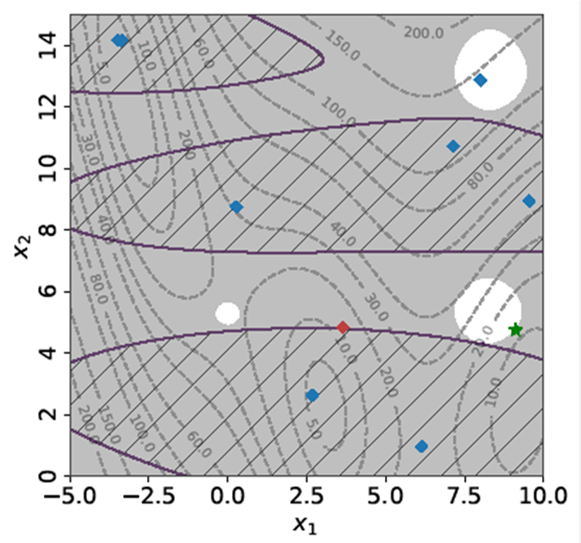}}
    \subfloat[SEGO-UTB ($2^{nd}$ iter.)]{\includegraphics[width=0.33\textwidth]{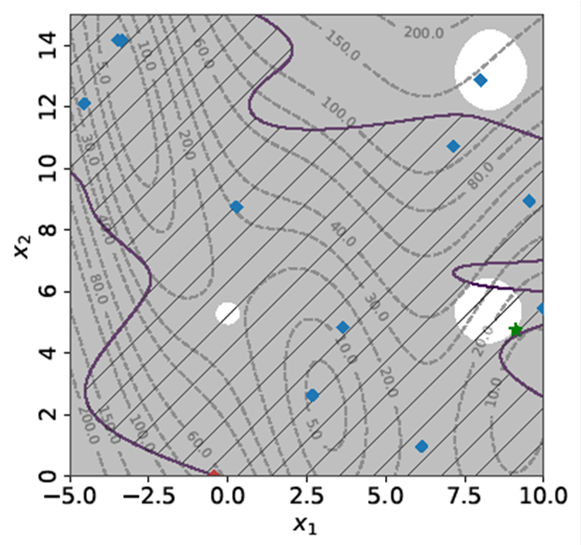}}
    \subfloat[SEGO-UTB (last iter.)]{\includegraphics[width=0.33\textwidth]{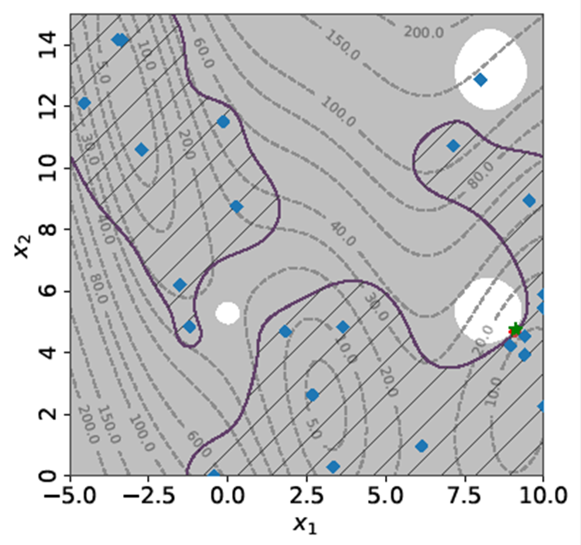}}
    \caption{An \change{illustration} of three iterations of SEGO and SEGO-UTB (with a $\tau_g = 3$) on the modified Branin problem. The hatched area is the SEGO/SEGO-UTB unfeasible domain, the grey area shows the true unfeasible domain and the dashed curves are the contour plots of the objective function. The blue squares are the current DoE whereas the green star indicates the global minimum of the problem. The red square is the new point to add in the DoE.}
    \label{fig:mc-utb}
\end{figure}

Figure~\ref{fig:mc-utb} illustrates the iterative process  of SEGO and SEGO-UTB for three chosen iterations (first, second and final iterations) on the modified Branin problem. The feasible domain evolves with the GPs associated to the constraints and the proposed learning rates. Clearly, by including the UTB, SEGO-UTB is able to explore more the feasible domain. It converges to the global minimum while SEGO could not explore the whole feasible domain. Therefore, although SEGO is showing a fast convergence (as it requires only $3$ iterations), it reaches only a local minimum.

For a given iteration $l$, the update strategy of the constraints learning rates $\bm{\tau_g}^{(l)}$ and $\bm{\tau_h}^{(l)}$, turns to be \change{an efficient tool to control the trade-off between exploration of the design space and the minimization of the objective function in the SEGO-UTB framework.} \remove{crucial for the dynamic of the SEGO-UTB framework to ensure a better trade-off between exploration of the \remove{the} design space and the minimization of the objective function.}
In the next subsection, different strategies for updating $\bm{\tau_g}^{(l)}$ and $\bm{\tau_h}^{(l)}$ are introduced.

\subsection{On the update of the constraints learning rate}
\label{ssec:utb_evol}

For simplicity reasons, the overall constraints function is denoted, for a given $\bm{x} \in \Omega$, by \hbox{$\bm{c}(\bm{x}) = [\bm{g}(\bm{x})^{\top},\bm{h}(\bm{x})^{\top}]^{\top}\in \mathbb{R}^{m+p}$}.
For a given iteration $l$, let \hbox{$\bm{\mu}^{(l)}_{\bm{c}}: \mathbb{R}^{d} \to \mathbb{R}^{m+p}$} and  \hbox{$\bm{\sigma}^{(l)}_{\bm{c}}: \mathbb{R}^{d} \to \mathbb{R}^{m+p} $} denote the mean and the standard deviation functions defining the GPs of the constraints $\bm{c}$, and let {$\bm{\tau}^{(l)}_{\bm{c}} = [[\bm{\tau}^{(l)}_{\bm{g}}]^{\top},[\bm{\tau}^{(l)}_{\bm{h}}]^{\top}]^{\top} \in \mathbb{R}^{m+p}$} be the associate constraints learning rate as given by the UTB feasibility criterion.

In general, the acquisition function $\alpha_f$ ensures that the exploration and exploitation trade-off is respected during the enrichment procedure.
Hence one can assume that the sample points are somehow well distributed in the design space during the optimization process.
In this context, one can use different strategies for updating the constraints learning rate $\tau^{(l)}_{c_i}$ (associated to the constraint $c_i$).
A first trivial strategy, is ensured by making the value of $\tau^{(l)}_{c_i}$ constant all over the iterations.
This choice is motivated by the fact that $\tau^{(l)}_{c_i}$ is used to scale $\sigma^{(l)}_{c_i}$, and the latter function decreases systematically whenever the model is getting accurate.
This updating strategy is noted (Cst.).
In this case, a natural constant choice for $\tau^{(l)}_{c_i}$ is $3$ for all \hbox{$i=1,\ldots, m+p$} and iteration index $l$.
With this value, the trust interval over all the reliable GPs is expressed with $99\%$ trust \changeb{for a single point $\bm{x}$ sampled in $\Omega$}. We note that, in case of only inequality constrained optimization problems, working with a fixed $\tau^{(l)}_{c_i}$ was also proposed in \cite{lam2015multifidelity,priem2019use}. 

A second possible strategy can be as follows.
In fact, as the quality of the GP approximation will most likely depend on the size of DoE, we expect that the larger is the size of the DoE, the better is the GP approximation.
This suggests naturally to be more confident on the GP prediction when the number of sample points in the DoE increases.
Hence, for a given $i=1,, \ldots, m+p$, the constraints learning rate $\tau^{(l)}_{c_i}$ (associated to the constraint $c_i$) should be reduced as far as the number of points of the DoE increases (i.e., $l$ getting larger).
Assuming that the maximum number of iterations (see \mbox{max\_nb\_it} in Algorithm \ref{alg:BO}) is large enough, so that the GP approximations can be considered very accurate, the constraints learning rate can be decreased systematically (from a given initial value, typically $\tau^{(0)}_{c_i}=3$) to reach zero at the end of the optimization process (i.e., $\tau^{(\mbox{max\_nb\_it})}_{c_i}=0$).

\begin{figure*}[htb!]
    \centering
    \subfloat[Decreasing strategy \label{fig:cst_de}]{\includegraphics[width=0.8\textwidth]{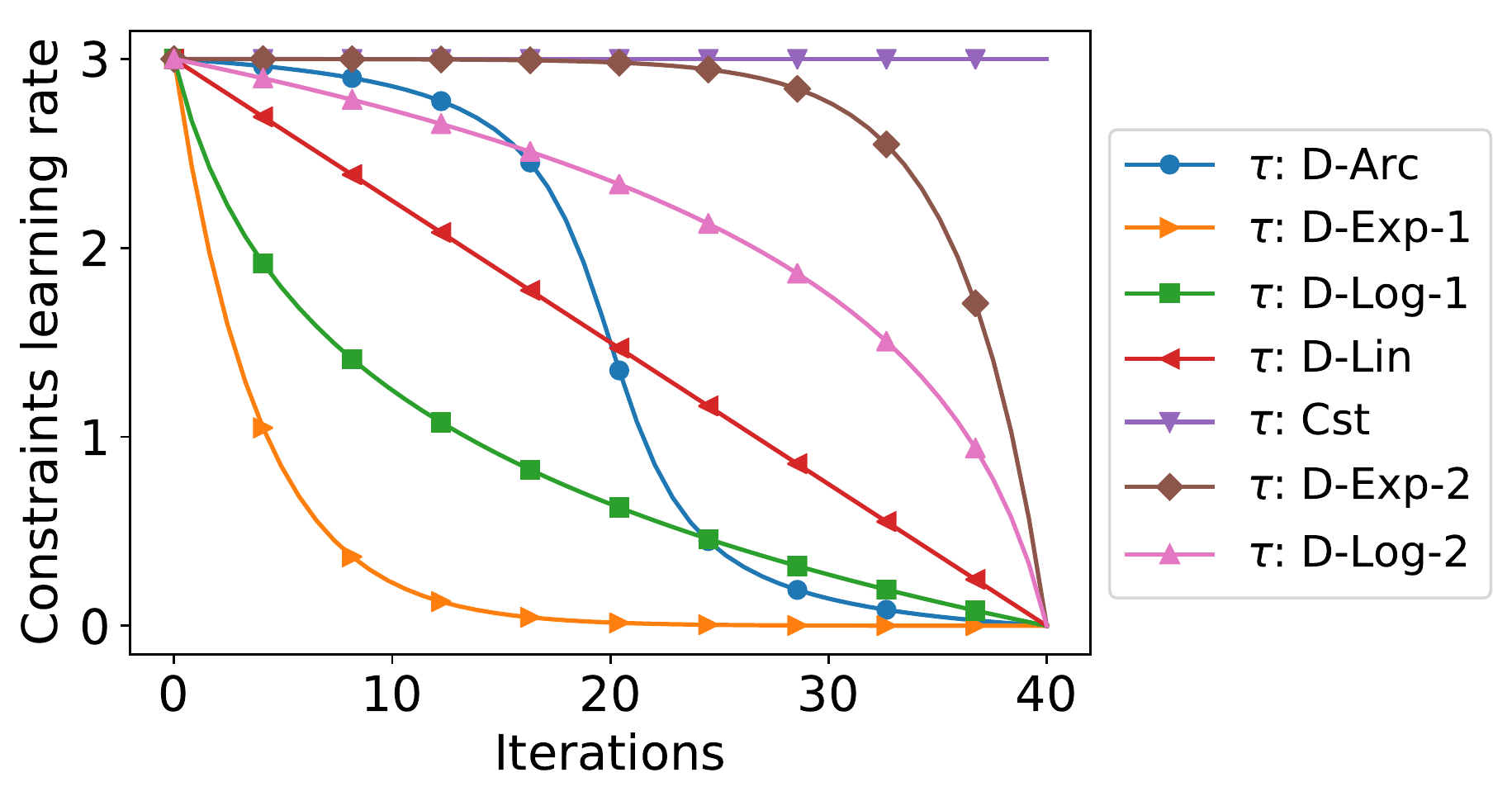}} \\
    \subfloat[Non-decreasing strategy \label{fig:cst_in}]{\includegraphics[width=0.8\textwidth]{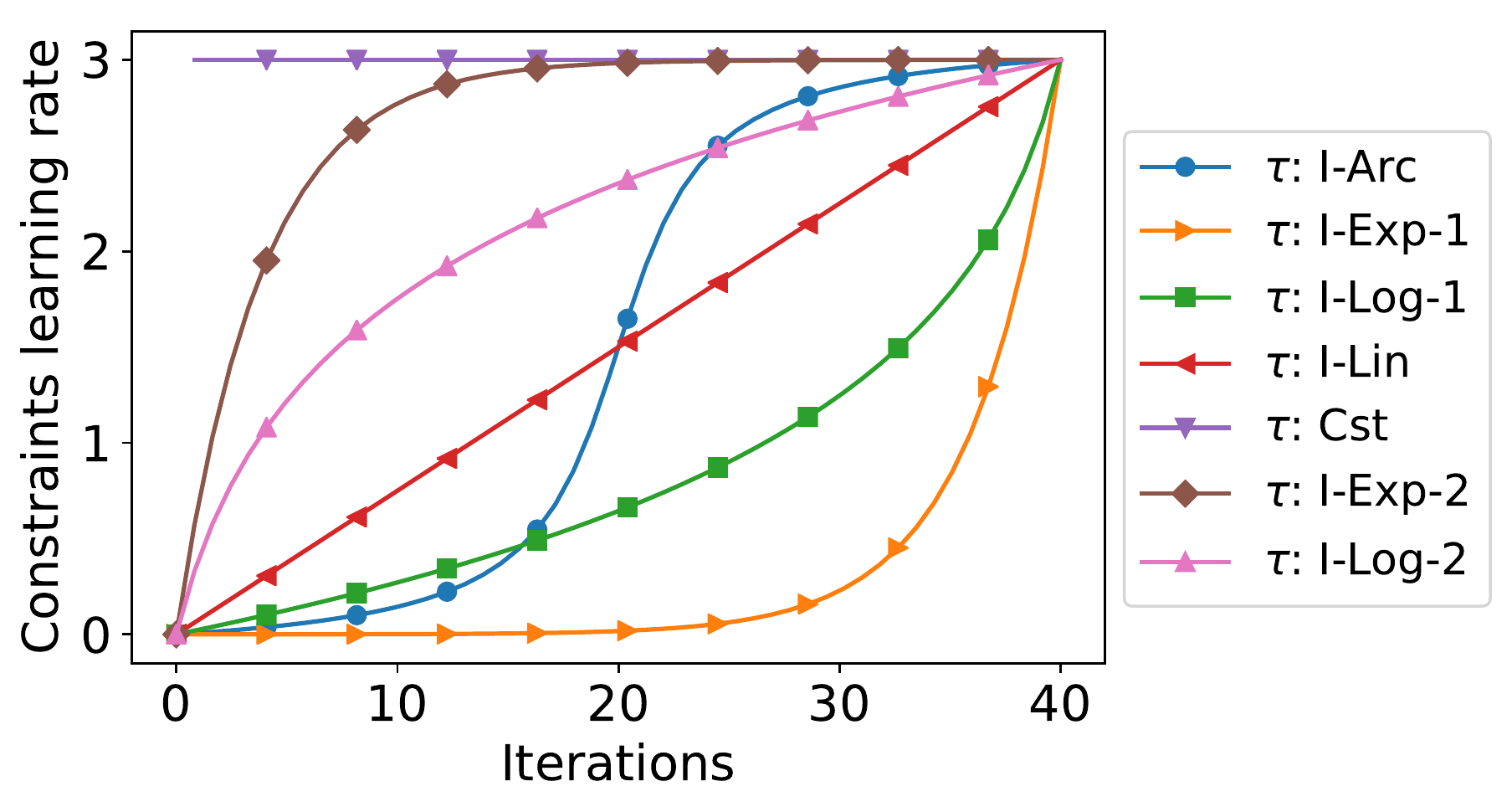}}
    \caption{Evolution of the constraints learning rate over iterations (with $\mbox{max\_nb\_it}=40$).}
    \label{fig:utb_strat}
\end{figure*}

Figure \ref{fig:cst_de} illustrates different decreasing monotonic profiles for each component of the constraints learning rate: arc-tangent (Arc), linear (Lin), two logarithmic (Log), and two exponential (Exp) profiles for $\tau^{(l)}_{c_i}$. 
Within such updating strategies, for each component, the constraints learning rate is reduced regardless the quality of the GP approximation of the constraint during the optimization process.

In our numerical tests, we noticed that SEGO performs well during the early stages of the optimization; this remark motivates the following updating strategy of the constraints learning rate.
In fact, we will try to mimic SEGO during the first iterations and, then, incorporate gradually the uncertainties on the GPs of the constraints in order to encourage a better exploration of the feasible domain.
We obtain, for each constraint, a non-decreasing learning rate with respect to the outer iterations of the regarded algorithm.
Namely, at the first iteration, the constraints learning rate related to each constraint is first set to zero and then systematically increased to reach a maximum value (typically, at the end of the optimization, one would have \change{$\tau^{(\mbox{max\_nb\_it})}_{c_i}=3$)}.
Different non-decreasing  strategies can be used; see Figure \ref{fig:cst_in} for the non-decreasing trends tested in the context of this paper.

In the next section, different decreasing and non-decreasing strategies for the constraints learning rate are tested on well-know mixed-constrained test problems.
A comparison with the state-of-the-art solvers is also included.

\section{Numerical tests}
\label{sec:test}

In this section, the potential of the proposed algorithm is evaluated using an extensive test set of 29 problems. 

\subsection{Implementation details}
\label{ssec:impl}

Our implementation choices for SEGO-UTB and SEGO are as follows. For the GPs, we choose to work with linear regression trend and \remove{an exponential} \change{a Gaussian} correlation function, all based on the open-source Python \textit{surrogate modeling toolbox} (SMT) \cite{BouhlelPython2019}.
The choice of all the hyper-parameters is handled by the default settings of the toolbox. \change{As acquisition function, we used the explicit \textit{scaled Watson Barnes} (WB2S) \cite{Bartoliadaptivemodeling2019}} \remove{The explicit WB2S \cite{Bartoliadaptivemodeling2019} is used to $\alpha_f$}. \change{For a given iteration $l$, the WB2S infill criterion is given by
\begin{equation}
    \alpha_f^{(l)}(\bm{x}) = s^{(l)} \ \text{EI}^{(l)}(\bm{x}) - \mu_f^{(l)}(\bm{x}) ,
    \label{eq:wb2s}
\end{equation}
where $\text{EI}^{(l)}(\bm{x})$ is the expected improvement function \cite{JonesEfficientglobaloptimization1998} at $\bm{x}$  and $\mu_f^{(l)}(\bm{x})$  is the mean function of the GP model of $f$ at the iteration $l$. To define the scale factor $s^{(l)}$, we first compute $\bm{x}_{\text{EI}_{\max}} = \arg \max_{\bm{x}} \text{EI}(\bm{x})$, then we set $s^{(l)}$ to $100 \frac{\left|\mu_f^{(l)}\left(\bm{x}_{\text{EI}_{\max}}\right)\right|}{\text{EI}\left(\bm{x}_{\text{EI}_{\max}}\right)}$ if $\text{EI}\left(\bm{x}_{\text{EI}_{\max}}\right) \neq 0$, and to $1$ otherwise. In our case, the point $\bm{x}_{\text{EI}_{\max}}$ is chosen as the point that maximizes the $\text{EI}$ function among $100d$ points from the design space (generated using Latin Hyper-cube Sampling strategy).

During the optimization process, the points in the DOE are not necessary feasible. In fact, the infill criterion WB2S can be evaluated even if all the points in the DOE are infeasible \cite{Bartoliadaptivemodeling2019}. However, at the end of the optimization, if all the points in the DOE are infeasible, SEGO-UTB and SEGO will return the point with the minimal constraints violation.}
The initial (resp. final) value associated with the decreasing (resp. non-decreasing) strategies for all components of the constraint learning rates \remove{$\tau_C$} \change{$\tau_c$} is set to 3.
The optimization sub-problem (\ref{eq:inner_loop_segre}) is solved in two steps. The first consists in finding a warm starting point by solving \eqref{eq:inner_loop_segre} with the \textit{Improved Stochastic Ranking Evolution Strategy} (ISRES) \cite{runarsson2005search} of the \change{Python} NLopt package \cite{johnson2014nlopt}.
The solution point is then obtained by solving \eqref{eq:inner_loop_segre} using SNOPT \cite{Gillsnopt2005}, from the PyOptSparse toolbox \cite{Perezpyopt2012}, where the starting point is set as the solution returned by ISRES.

\subsection{Solvers in the comparison}

In this paper, SEGO-UTB is compared to some of the state-of-the-art BO solvers (in addition to SEGO):
\begin{itemize}
    \item ALBO: a BO solver using an augmented Lagrangian approach \cite{PichenyBayesianoptimizationmixed2016}, 
    \item SUR: a BO solver using a stepwise uncertainty reduction \cite{Pichenystepwiseuncertaintyreduction2014}, 
     \item PESC: a BO solver based on the predictive entropy search \cite{Hernandez-Lobatogeneralframeworkconstrained2016},
     \item EFI: a BO solver based on the expected feasibility improvement \cite{SchonlauGloballocalsearch1998}.
\end{itemize}
\change{The solvers ALBO, EFI and SUR were taken from the DiceOptim R package \cite{PichenyDiceOptim2016} while the
PESC solver was taken from the Spearmint Python toolbox\footnote{\label{Spearmint}https://github.com/HIPS/Spearmint}. All the parameters of those BO solvers were kept unchanged except the the correlation function which is set to be Gaussian. In fact, the proposed default choice for the correlation function (a Matérn correlation function) did not perform well in our numerical tests.}

For completeness, two well-known derivative free solvers are also included in the comparison:
\begin{itemize}
    \item NOMAD: a mesh adaptive direct search solver \cite{audetmesh2006,lealgorithm2011}. The default parameters are kept unchanged.
    \item COBYLA: a trust-region solver based on linear approximations \cite{powelldirect1998}.
    We worked with the Scipy \change{Python} toolbox implementation of COBYLA  \cite{Jonesscipy2001}.
\end{itemize}

We note that SUR, PESC, EFI, NOMAD and  COBYLA, handle only inequality constraints.
To manage equality constraints, each constraint of the form $\bm{h}(\bm{x})=0$ are changed into two inequality constraints of the form $\bm{h}(\bm{x}) \geq 0$ and $\bm{h}(\bm{x}) \leq 0$. 

We stress also that, unlike all the tested BO solvers where one starts with an initial DoE, NOMAD and COBYLA require only a single initial point to start the optimization procedure.
To not penalize the latter two solvers, we choose the best valid point in the initial DoE as a first guess for NOMAD and COBYLA.
If there is no valid point, the best point (i.e., with the smallest violation of the constraints) in the initial DoE is chosen.

\change{Last, for all the tested solvers including SEGO and SEGO-UTB, two tolerances on the violation for each of constraints are considered, namely, $\epsilon_c=10^{-2}$ and $\epsilon_c=10^{-4}$.}

\subsection{Comparison results using convergence plots}
\label{subsec:cv_plots}

In this subsection, we will analyse the performance of all the tested solvers using convergence plots related to a set of four known problems.

\subsubsection{Test problems}

Among the four tested problems, three are taken from \citet{PichenyBayesianoptimizationmixed2016} for which ALBO is in particular very competitive.
One of the problems, is the \textit{Linear-Hartman-Ackley} (LAH) test case, which has four design variables, a linear objective function, one equality constraint (given by the Hartman function) and one inequality constraint (given by the Ackley function).
The second mixed optimization problem has two design variables with a re-scaled version of the "Goldstein-Price" function as objective function.
The problem is constrained with one inequality constraint (given by a sinusoidal function) and two equality constraints (using a centered "Branin" function and a function taken from \citet{ParrInfillsamplingcriteria2012}); henceforth, this problem is named GBSP.
The third problem involved two design variables, a linear objective function, a sinusoidal and a quadratic inequality constraints; henceforth named LSQ.
The additional fourth problem is the \textit{Modified Branin} (MB) test case  \cite{ParrReviewefficientsurrogate2010}.
This problem has two design variables, a non-linear objective function and one inequality constraint.
The full expressions of these problems can be found in \ref{app:repre_pb}.

\subsubsection{Convergence plots}
\label{ssec:plan_CV}

Similarly to \citet{PichenyBayesianoptimizationmixed2016}, we will build four convergence plots \change{by problem} to assess the good performance of our solver.
For each problem, the convergence plots are built on the following way.
First, we perform $100$ independent runs for each solver using $100$ different initial DoEs using the Latin Hypercube Sampling method. For each run, all the solvers are initiated with the same DoE.
The size of the initial DoEs and the maximum number of iterations are respectively set to $n_{start} = \max(d+1,5)$ and $\textit{max\_nb\_it} =  40d-n_{start}$ where $d$ is the dimension of the regarded problem (meaning a total budget of $40d$ function evaluations).

All the solvers are then compared by displaying the average and the standard deviation, up to a scaling factor, of the best values over the $100$ runs for increasing number of evaluations.
The best value is defined as the best valid value if there is, at least, one valid point in the DoE, otherwise, a penalization replaces the obtained invalid value.
The penalization is set to $3$ for LAH and GBSP, $2$ for LSQ and $150$ for the MB problem.

\subsubsection{Results}
In what follows, we stress that, concerning SEGO-UTB, different evolution strategies for updating the constraints learning rate are tested (as given in Figure \ref{fig:utb_strat}).
For clarity reasons, only the best \change{compromises} among the non-decreasing and decreasing strategies are considered \change{meaning SEGO-UTB ($\tau$: D-Exp-2) and SEGO-UTB ($\tau$: I-Exp-2).}
The constant constraints learning rate evolution (which corresponds to the strategy proposed in \cite{lam2015multifidelity} when only inequality constraints are present) is also included in the comparison. For completeness, the obtained results using all the constraints learning rate strategies can be found in \ref{app:res}.

\change{The obtained results considering two tolerances on the violation of the constraints} are presented on the following way. First, we confront the SEGO-like solvers (namely, SEGO, \change{SEGO-UTB ($\tau$: Cst), SEGO-UTB ($\tau$: D-Exp-2) and SEGO-UTB ($\tau$: I-Exp-2)).}
Then, the best SEGO-like solver is compared to the other solvers.

\begin{figure}[p]
    \centering
    \subfloat[$\epsilon_c = 10^{-2}$. \label{fig:mb_sego:2}]{\includegraphics[width=0.5\textwidth]{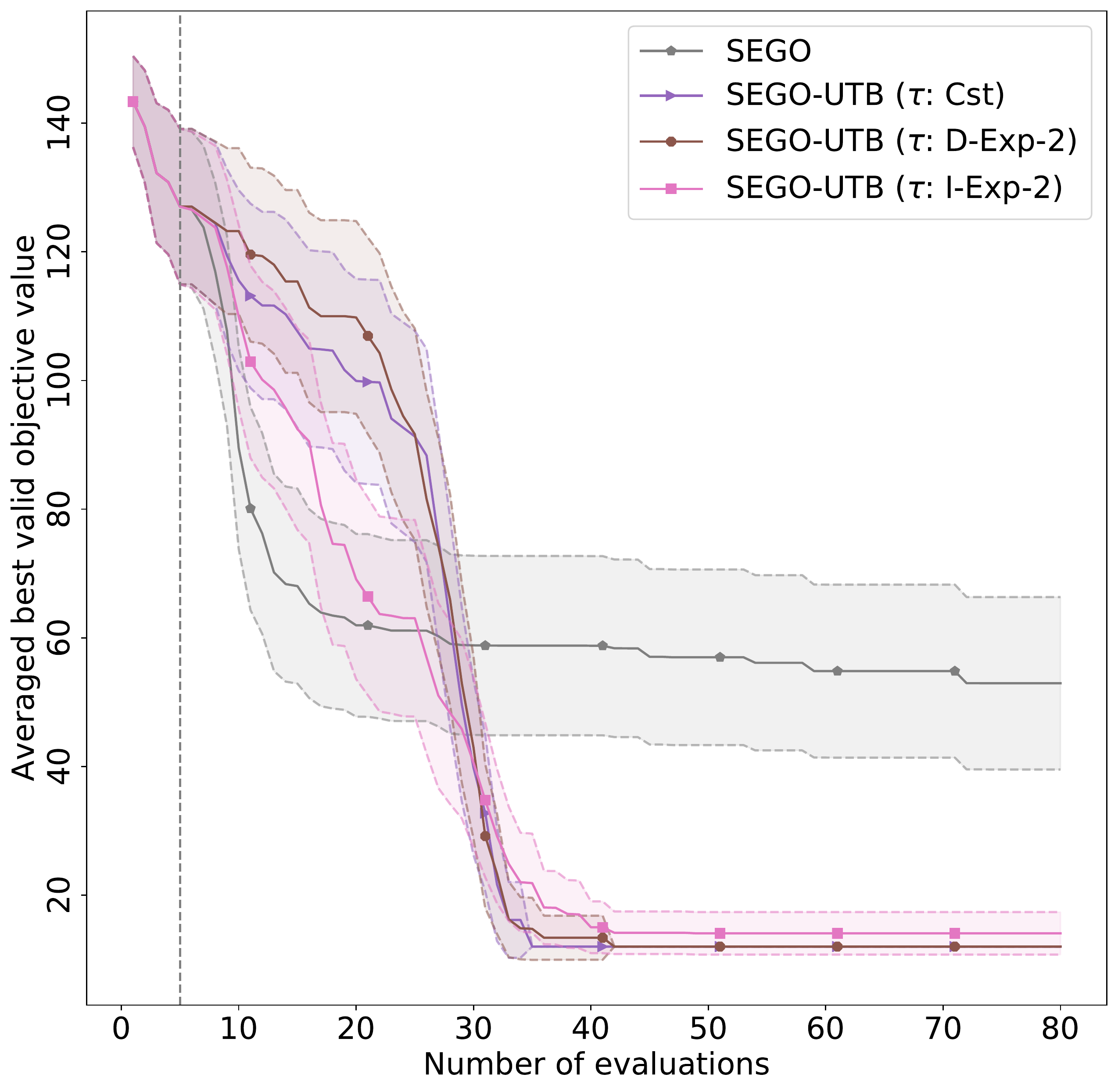}}
    \subfloat[$\epsilon_c = 10^{-4}$. \label{fig:mb_sego:4}]{\includegraphics[width=0.5\textwidth]{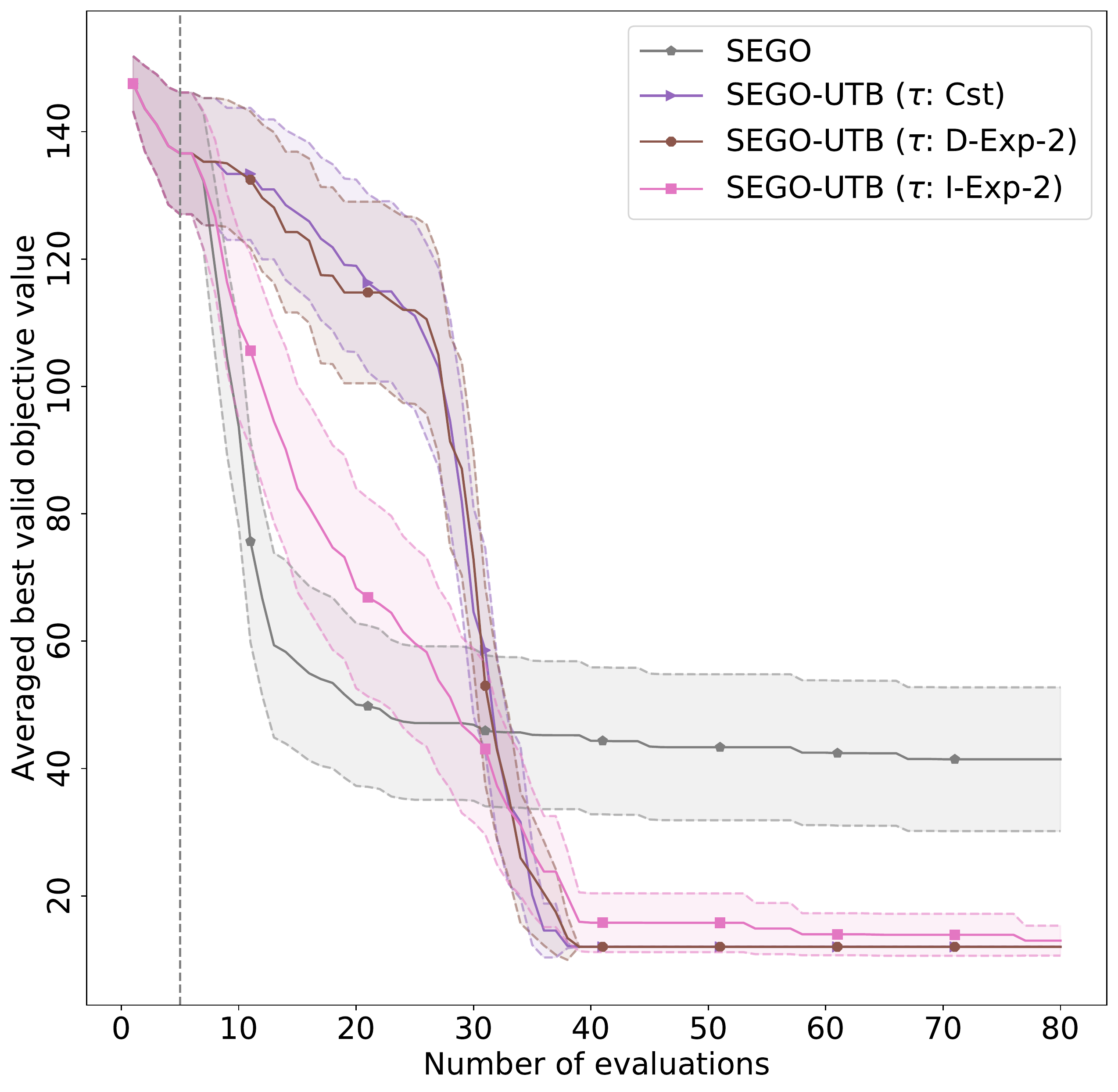}}
    \caption{Convergence plots for the MB problem on the SEGO-like solvers, considering the two levels of constraints violation $10^{-4}$ and $10^{-2}$. The vertical grey-dashed line outlines the number of points in the initial DoEs.}
    \label{fig:mb_sego}
\end{figure}

\begin{figure}[p]
    \centering
    \subfloat[$\epsilon_c = 10^{-2}$. \label{fig:mb:2}]{\includegraphics[width=0.5\textwidth]{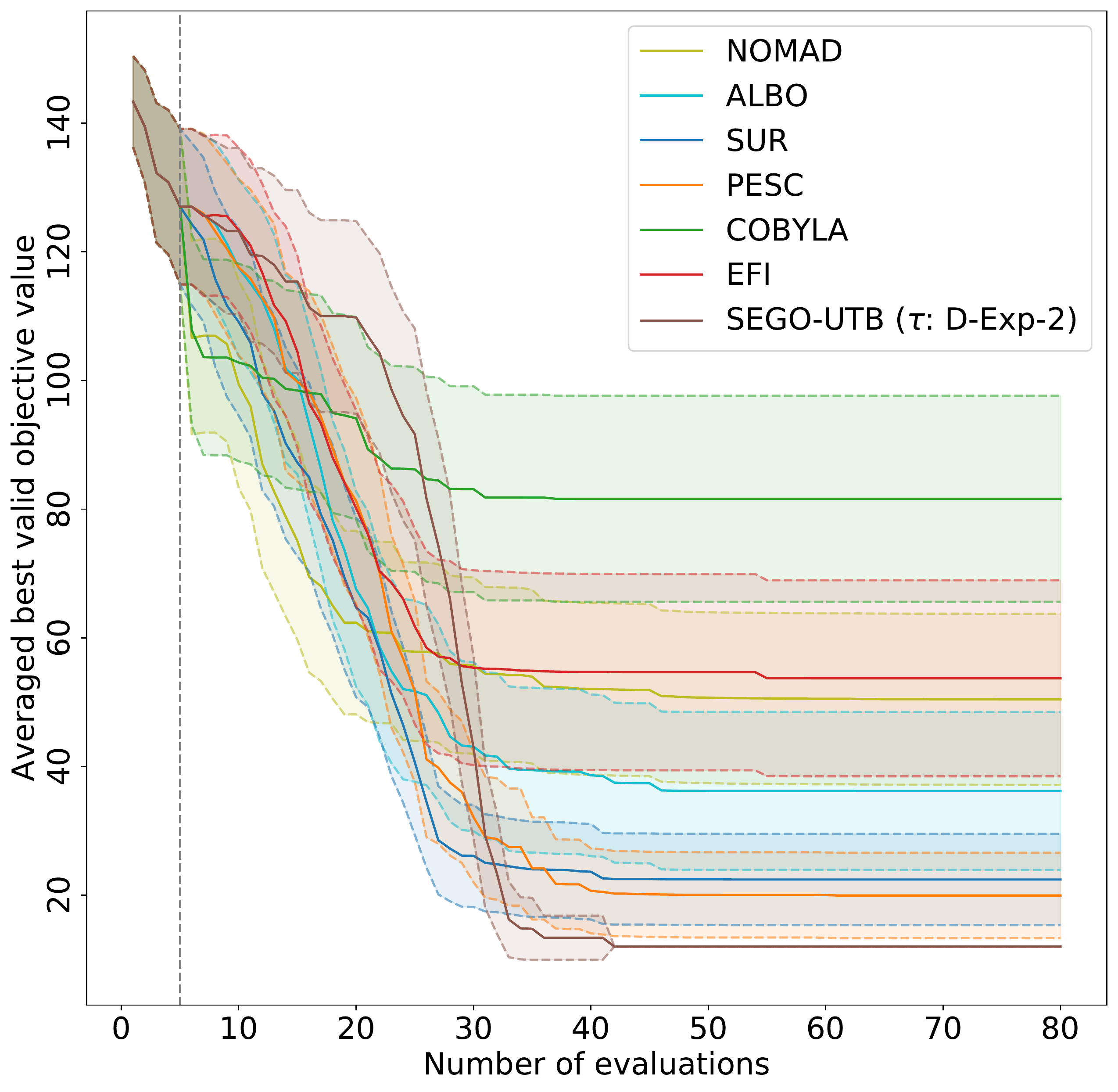}}
    \subfloat[$\epsilon_c = 10^{-4}$. \label{fig:mb:4}]{\includegraphics[width=0.5\textwidth]{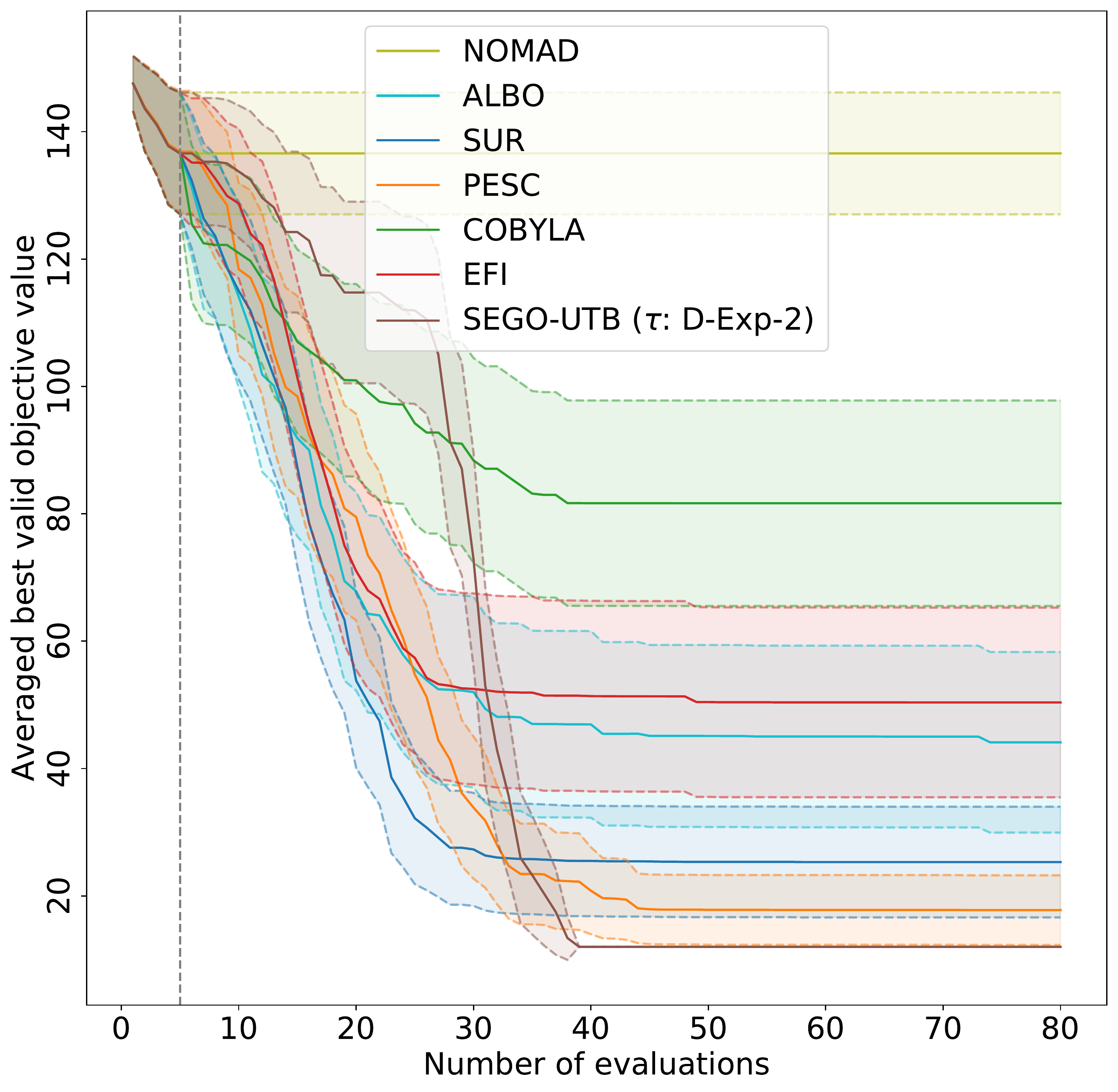}}
    \caption{Convergence plots for the MB problem NOMAD, COBYLA and BO solvers, considering the two levels of constraints violation $10^{-4}$ and $10^{-2}$. The vertical grey-dashed line outlines the number of points in the initial DoEs.}
    \label{fig:mb}
\end{figure}

Figure \ref{fig:mb_sego} depicts a comparison between SEGO and SEGO-UTB on the MB optimization problem.
Clearly, all the three SEGO-UTB variants are outperforming SEGO \change{for both levels of accuracy on the constraints violation.}
In fact, the averaged best valid value of SEGO  does not converge to the same value as SEGO-UTB.
Note also that SEGO displays a high standard deviation which means that it is not targeting all the time the same solution.
On the contrary, the SEGO-UTB variants are all targeting the global minimum zone.
Including uncertainties in the constraints models is leading to a better exploration of the feasible domain.
Also, one can see that SEGO-UTB \change{($\tau$: D-Exp-2)} displays the best performance.
For that reason, Figure \ref{fig:mb} shows a comparison of SEGO-UTB \change{($\tau$: D-Exp-2)} and the other tested solvers\change{ using the two considered constraints violations.}
Clearly, NOMAD  and COBYLA  are performing the worst among all the tested solvers\change{, in particular, when a strict tolerance on the constraints violation is used (see Figure \ref{fig:mb:4}). The high   associated standard deviation indicates that some of the runs converged to different values.}
The other tested BO solvers show a better performance although PESC, ALBO, SUR and EFI did not reach the global minimum.
In terms of the convergence speed, SEGO-UTB \change{($\tau$: D-Exp-2)} is being the fastest\change{ independently of the constraints violation tolerance.}

\begin{figure}[p]
    \centering
    \subfloat[$\epsilon_c = 10^{-2}$. \label{fig:lsq_sego:2}]{\includegraphics[width=0.5\textwidth]{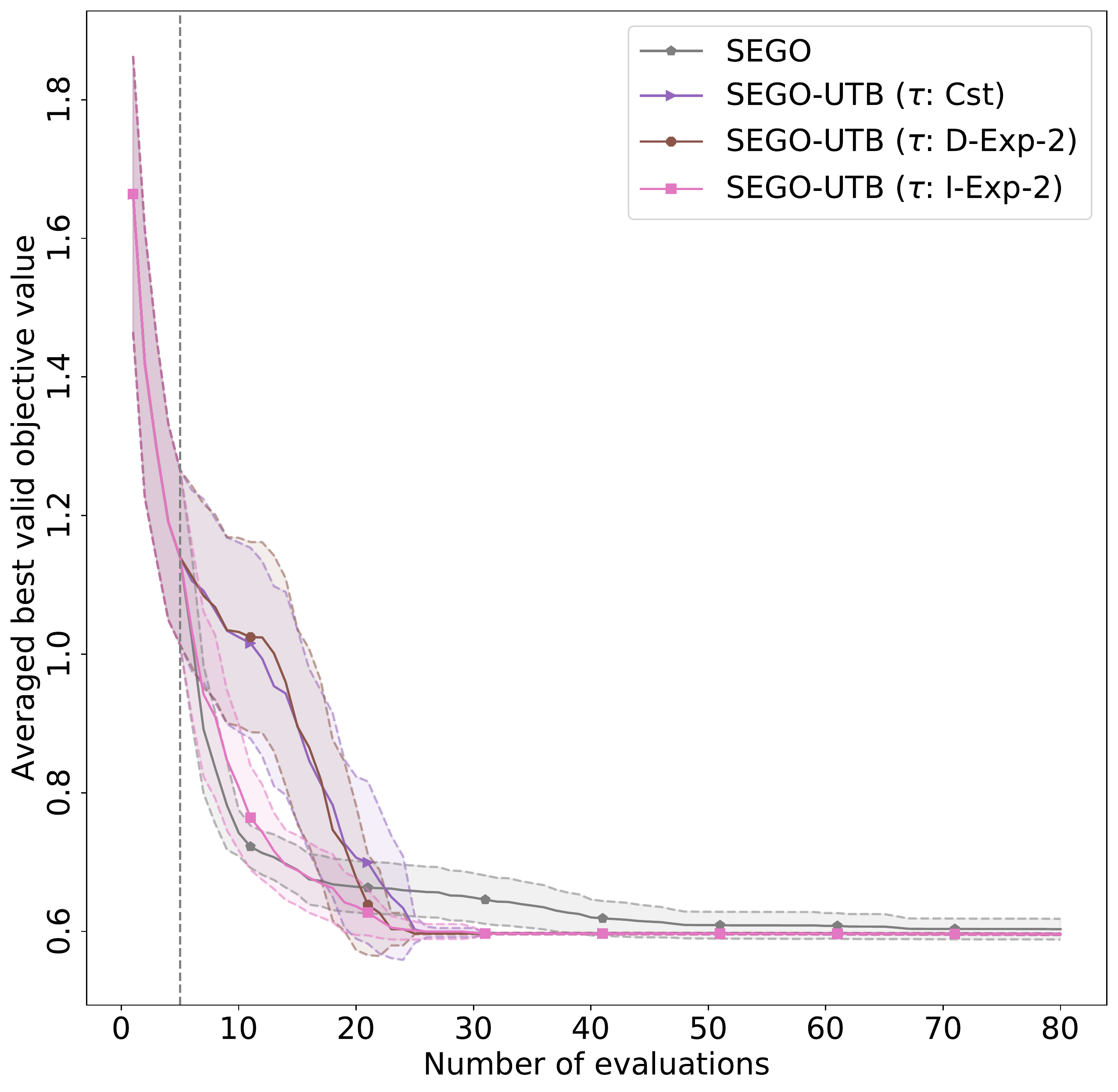}}
    \subfloat[$\epsilon_c = 10^{-4}$. \label{fig:lsq_sego:4}]{\includegraphics[width=0.5\textwidth]{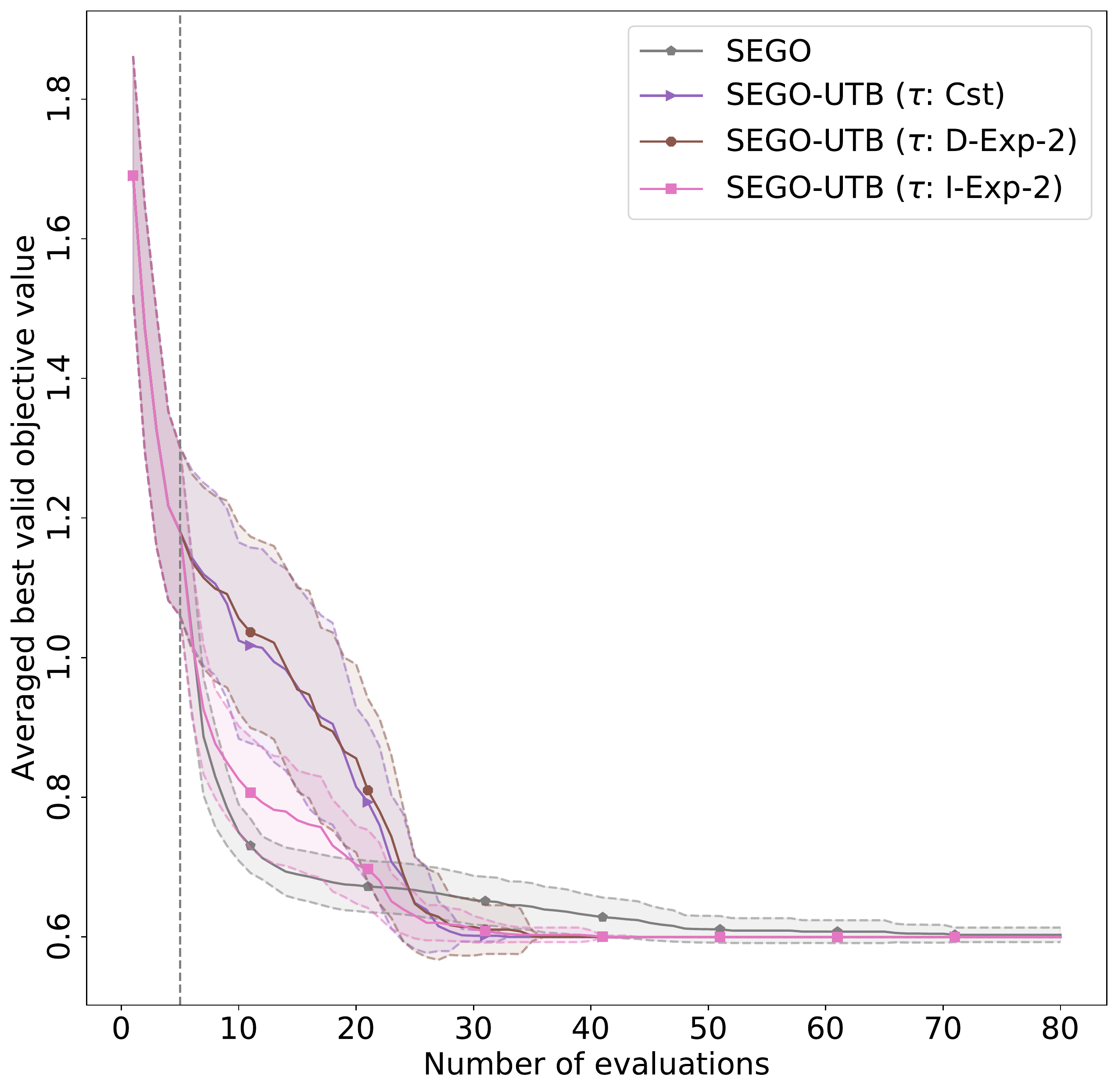}}
    \caption{Convergence plots for the LSQ problem on the SEGO-like solvers, considering the two levels of constraints violation $10^{-4}$ and $10^{-2}$. The vertical grey-dashed line outlines the number of points in the initial DoEs.}
    \label{fig:lsq_sego}
\end{figure}

\begin{figure}[p]
    \centering
    \subfloat[$\epsilon_c = 10^{-2}$. \label{fig:lsq:2}]{\includegraphics[width=0.5\textwidth]{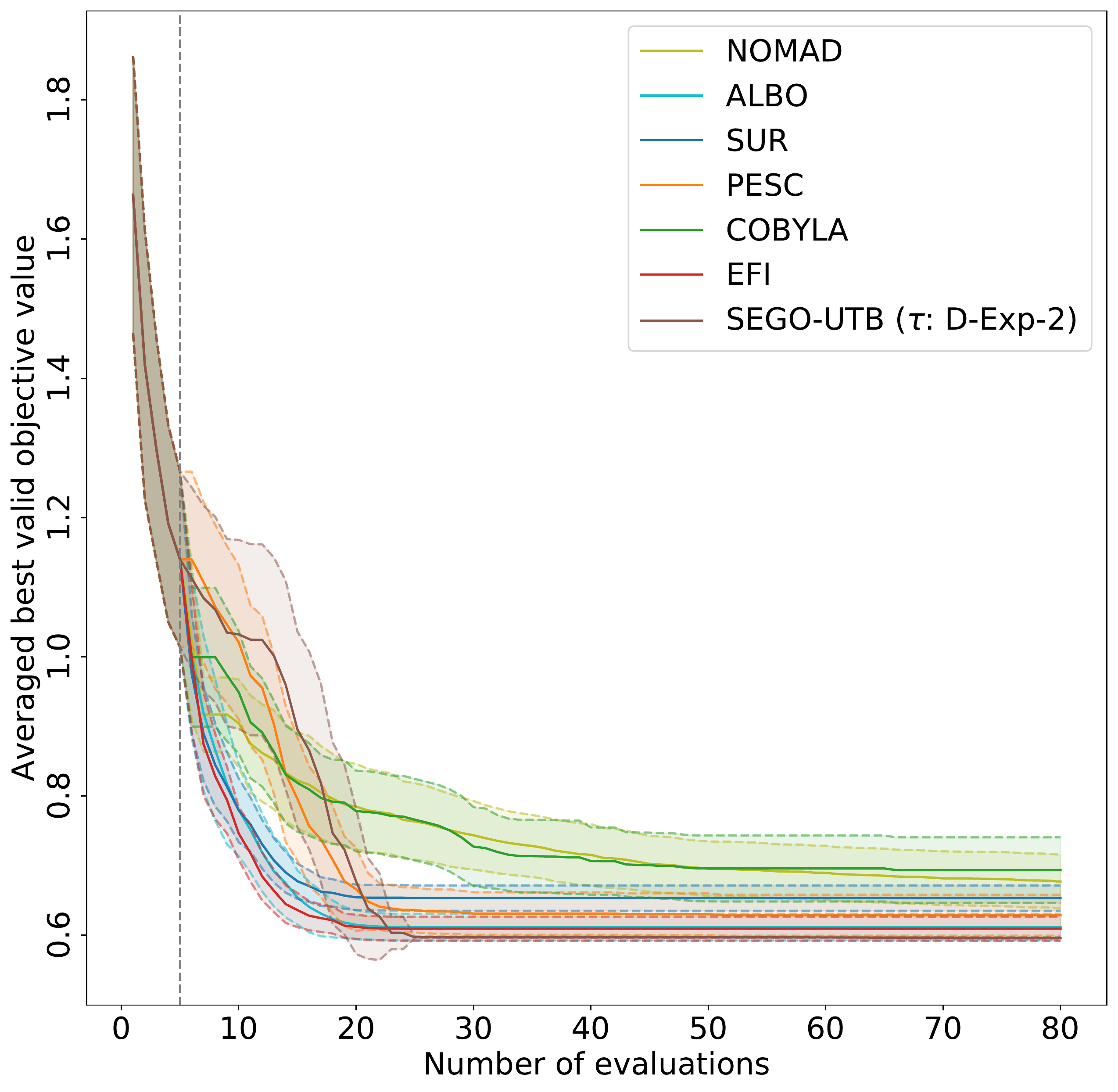}}
    \subfloat[$\epsilon_c = 10^{-4}$. \label{fig:lsq:4}]{\includegraphics[width=0.5\textwidth]{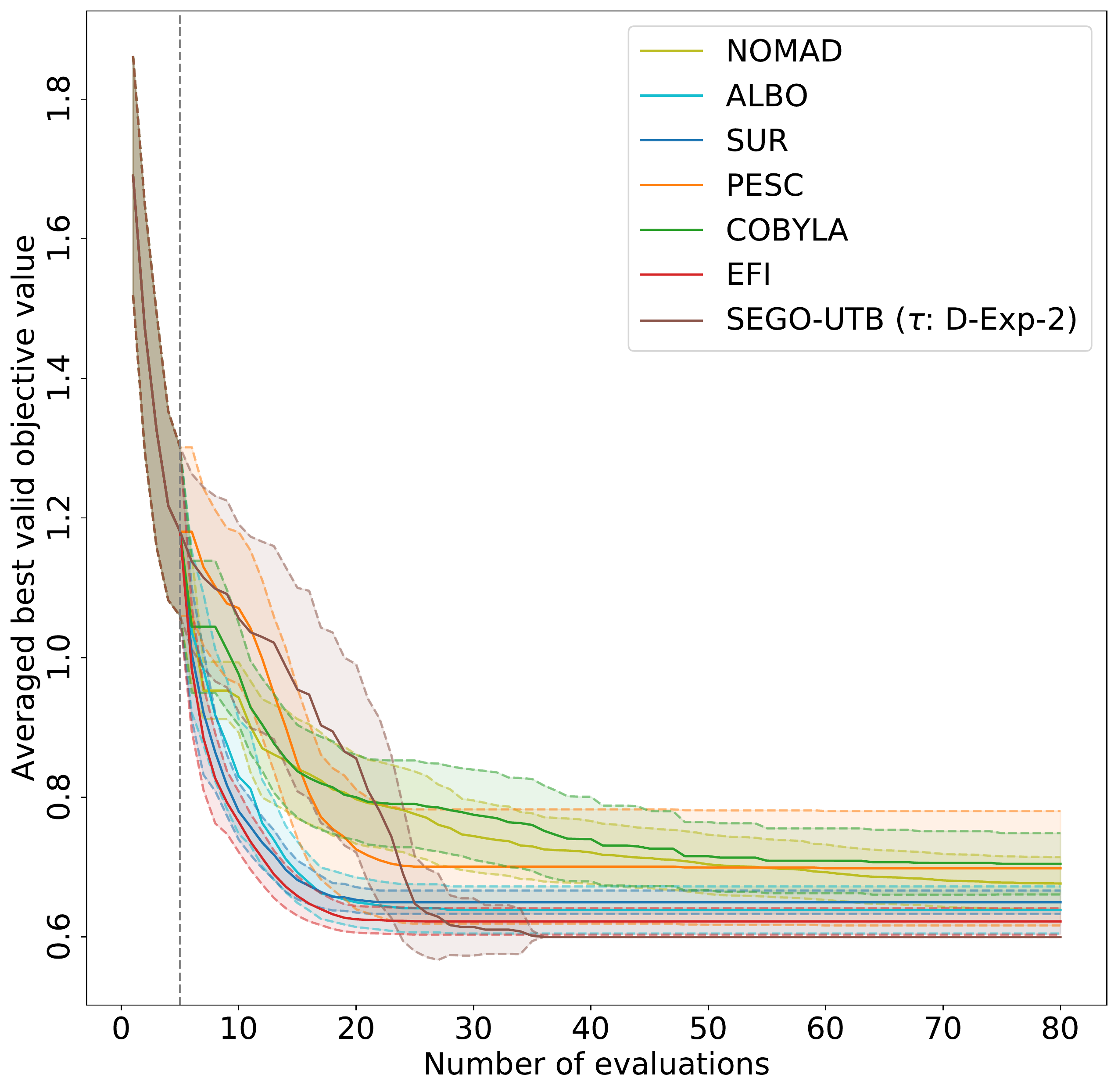}}
    \caption{Convergence plots for the LSQ problem NOMAD, COBYLA and BO solvers, considering the two levels of constraints violation $10^{-4}$ and $10^{-2}$. The vertical grey-dashed line outlines the number of points in the initial DoEs.}
    \label{fig:lsq}
\end{figure}

In Figures \ref{fig:lsq_sego} and \ref{fig:lsq}, we present the obtained results on the LSQ problem.
One can see that all the SEGO-like solvers are converging to the global minimum of this problem (see Figure~\ref{fig:lsq_sego}).
SEGO-UTB \change{($\tau$: I-Exp-2)} is exhibiting a slightly better performance in term of convergence speed.
In the comparison with the other solvers, see Figure~\ref{fig:lsq}, COBYLA and NOMAD are displaying the worst performance.
Figure \ref{fig:lsq} shows that PESC has better convergence properties with a high tolerance on the constraints violation.
Similarly to SEGO-UTB \change{($\tau$: D-Exp-2)}, the solvers ALBO, SUR and EFI are all converging to the global minimum in average except some runs; since their standard deviation is not converging to zero at the end of the optimization.

\begin{figure}[p]
    \centering
    \subfloat[$\epsilon_c = 10^{-2}$. \label{fig:gbsp_sego:2}]{\includegraphics[width=0.5\textwidth]{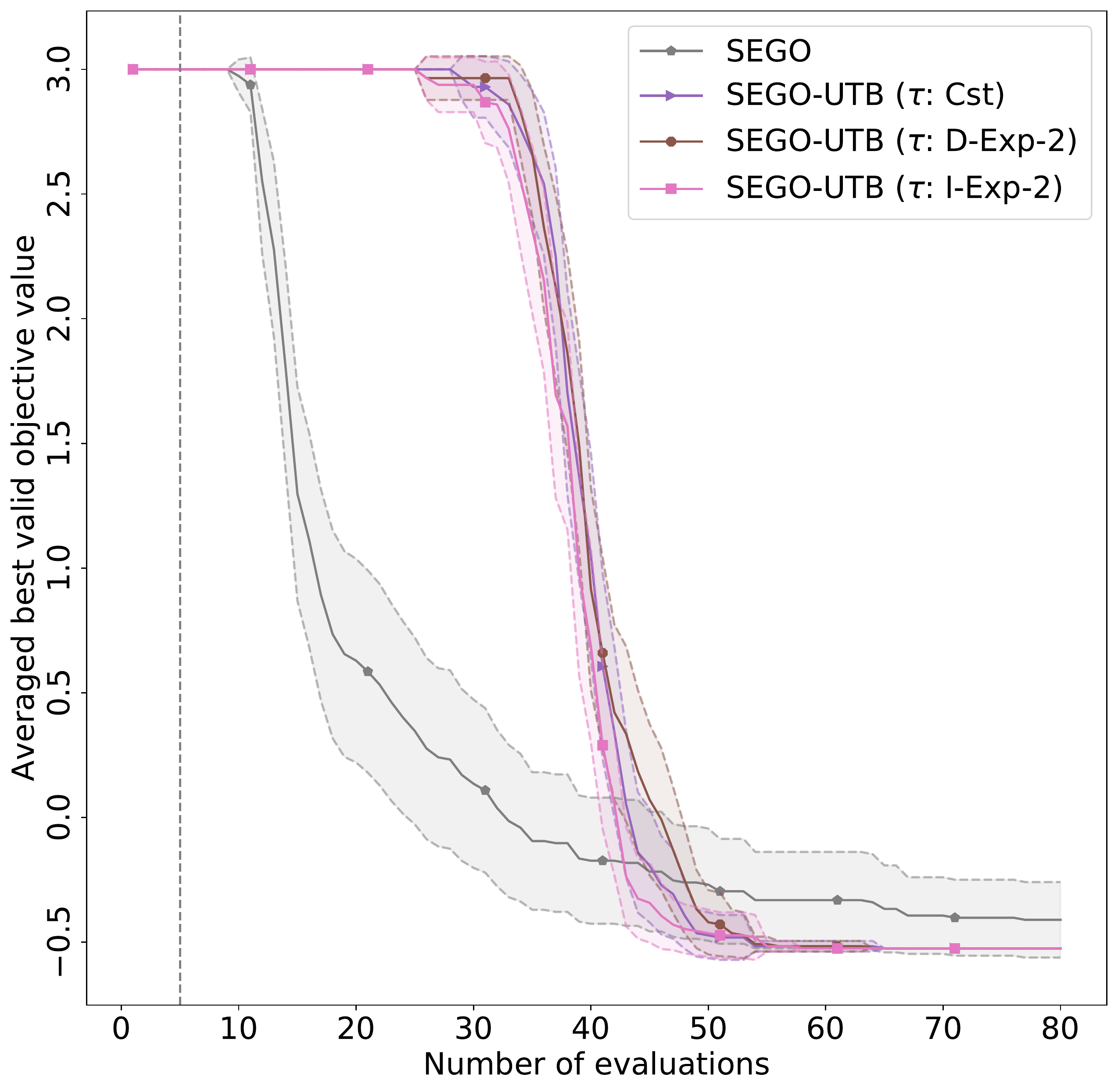}}
    \subfloat[$\epsilon_c = 10^{-4}$. \label{fig:gbsp_sego:4}]{\includegraphics[width=0.5\textwidth]{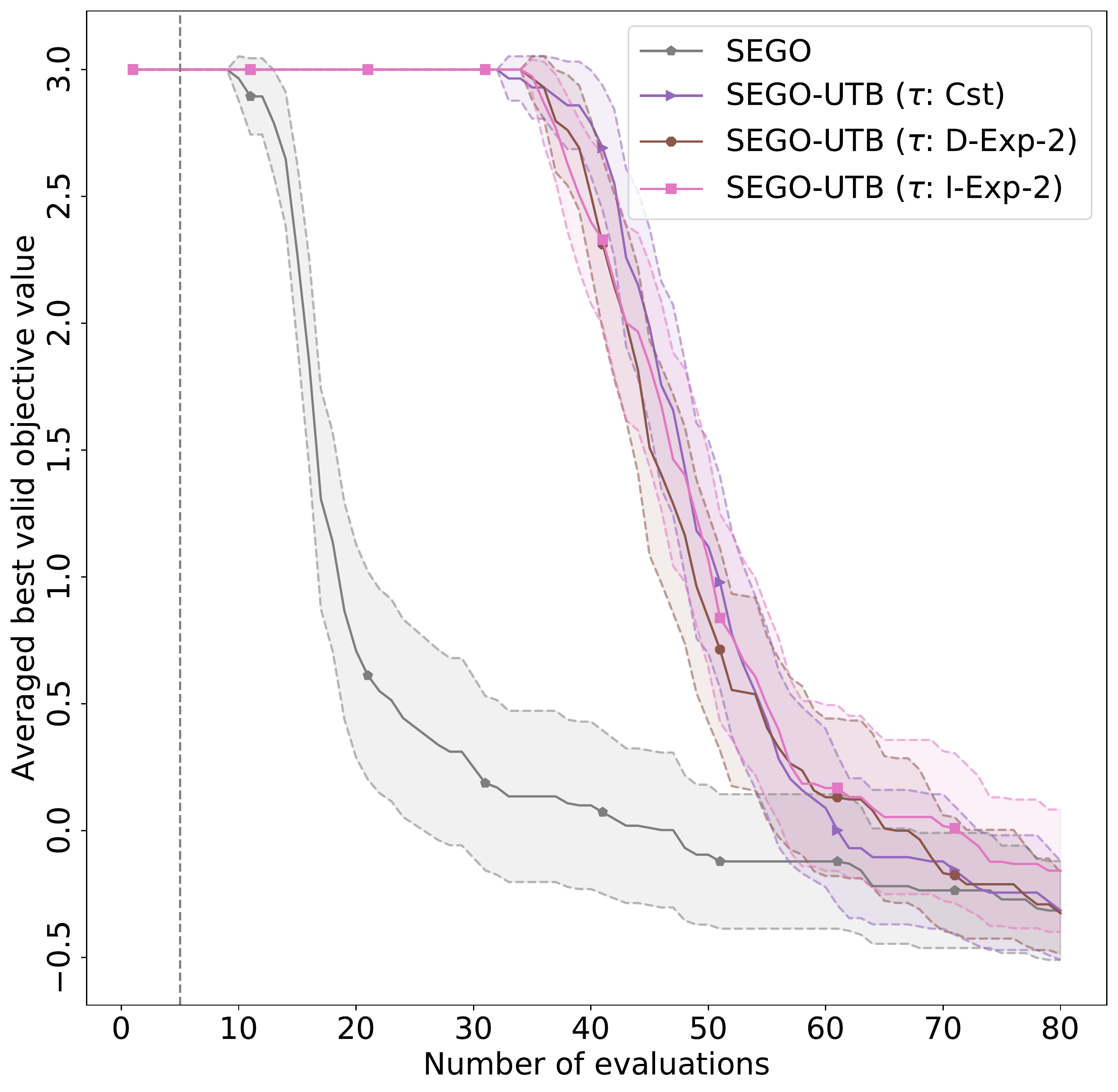}}
    \caption{Convergence plots for the GBSP problem on the SEGO-like solvers, considering the two levels of constraints violation $10^{-4}$ and $10^{-2}$. The vertical grey-dashed line outlines the number of points in the initial DoEs.}
    \label{fig:gbsp_sego}
\end{figure}

\begin{figure}[p]
    \centering
    \subfloat[$\epsilon_c = 10^{-2}$. \label{fig:gbsp:2}]{\includegraphics[width=0.5\textwidth]{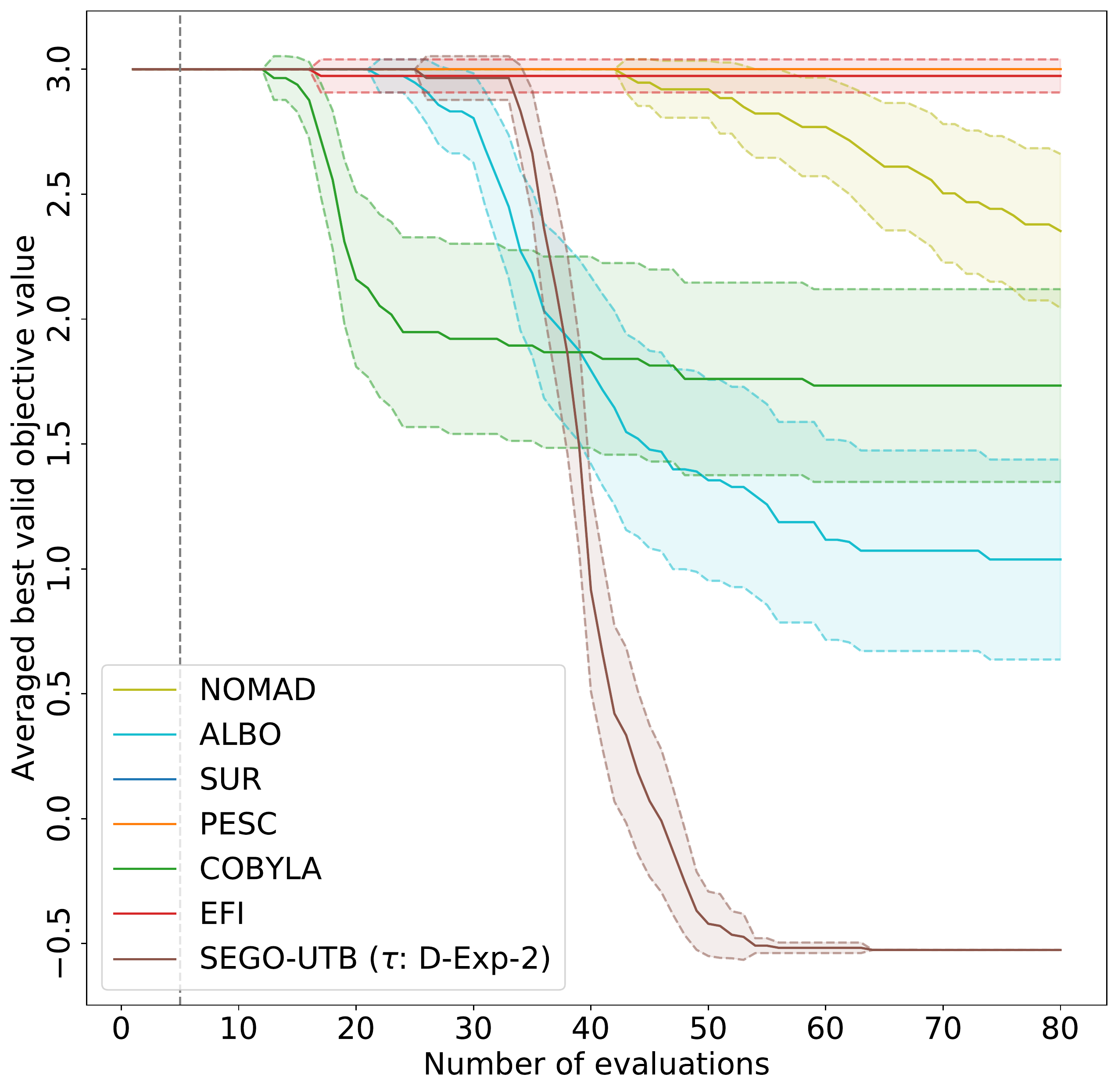}}
    \subfloat[$\epsilon_c = 10^{-4}$. \label{fig:gbsp:4}]{\includegraphics[width=0.5\textwidth]{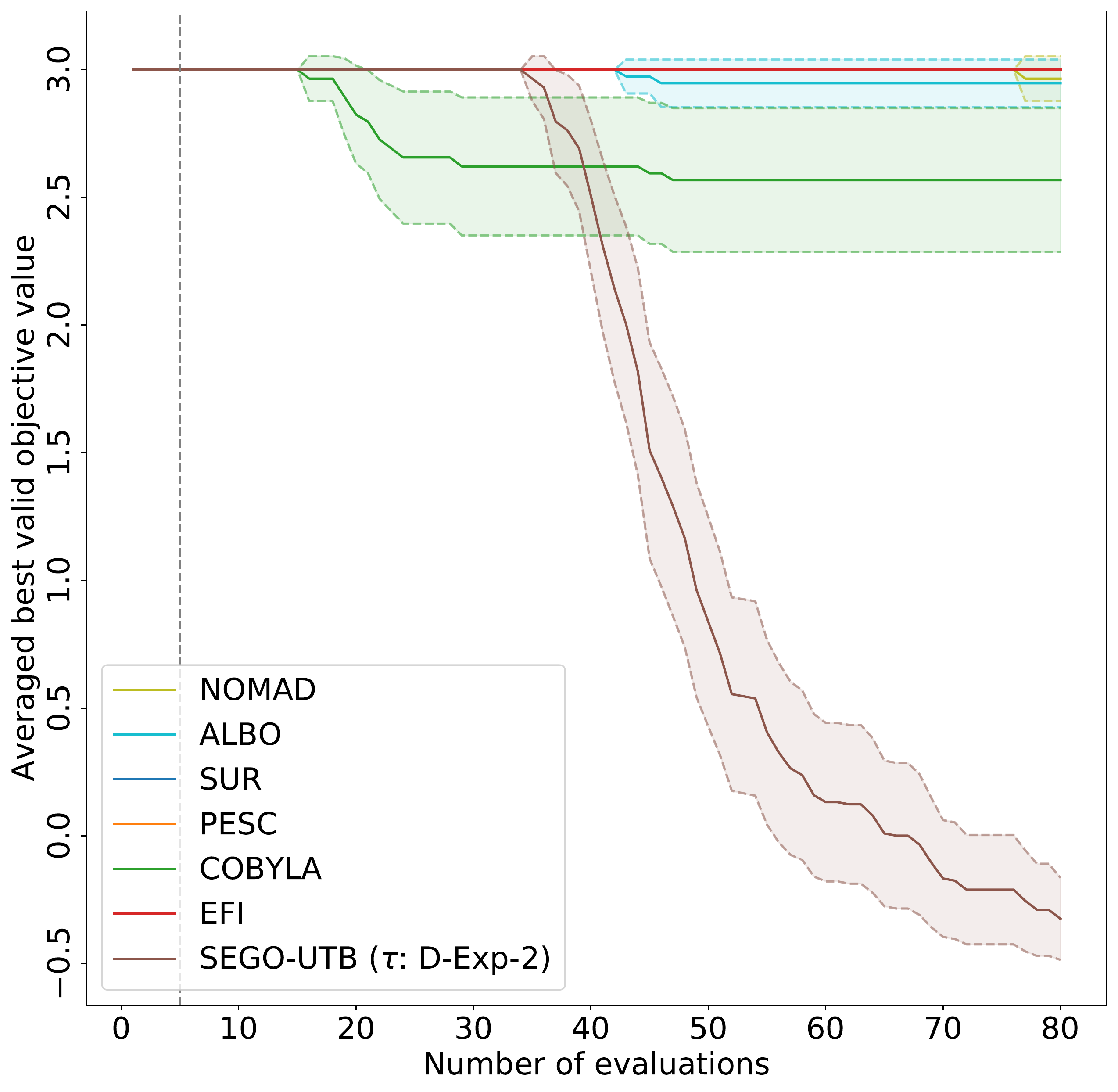}}
    \caption{Convergence plots for the GBSP problem NOMAD, COBYLA and BO solvers, considering the two levels of constraints violation $10^{-4}$ and $10^{-2}$. The vertical grey-dashed line outlines the number of points in the initial DoEs.}
    \label{fig:gbsp}
\end{figure}
 
\begin{figure}[p]
    \centering
    \subfloat[$\epsilon_c = 10^{-2}$. \label{fig:lah:2}]{\includegraphics[width=0.5\textwidth]{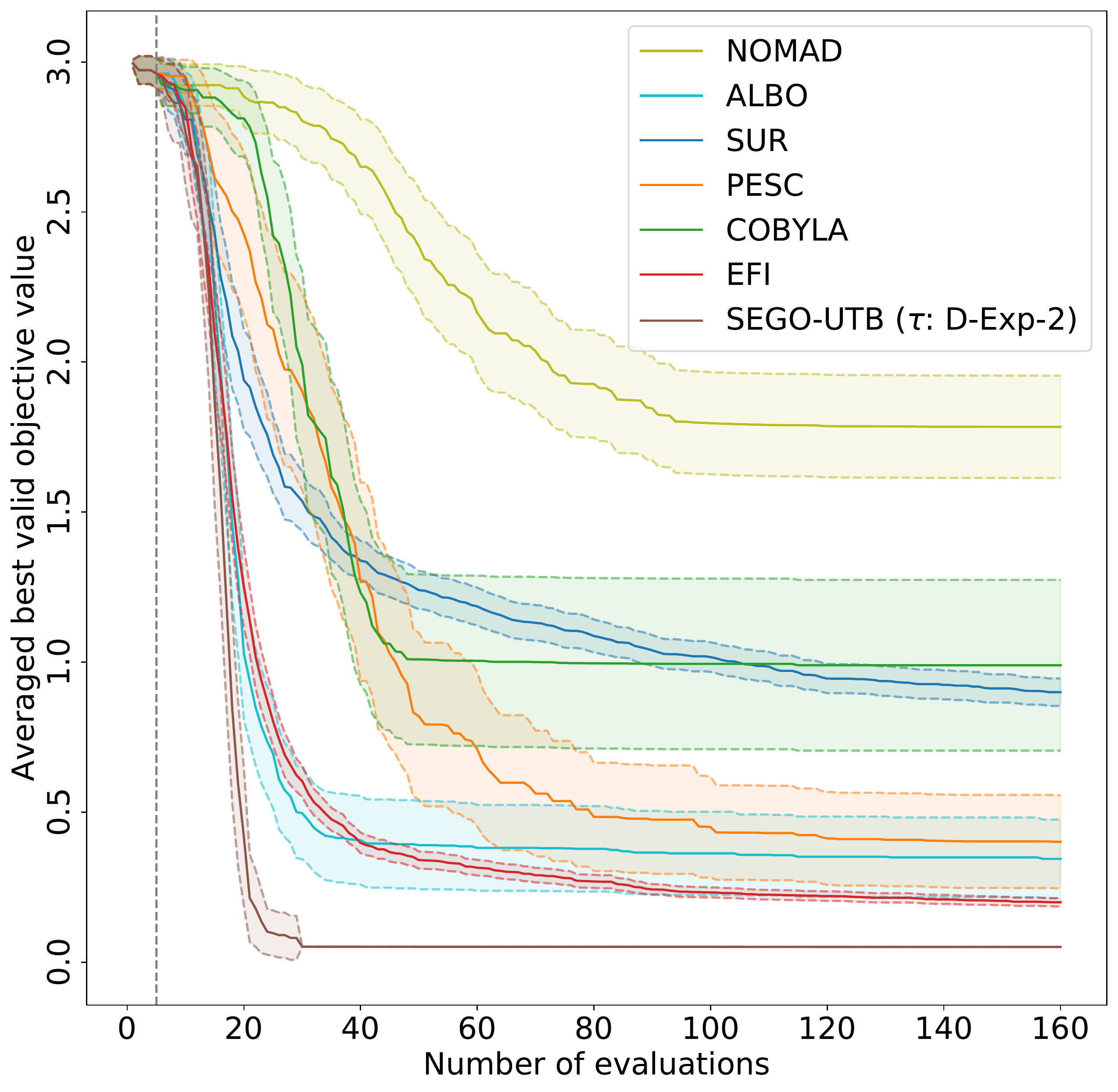}}
    \subfloat[$\epsilon_c = 10^{-4}$. \label{fig:lah:4}]{\includegraphics[width=0.5\textwidth]{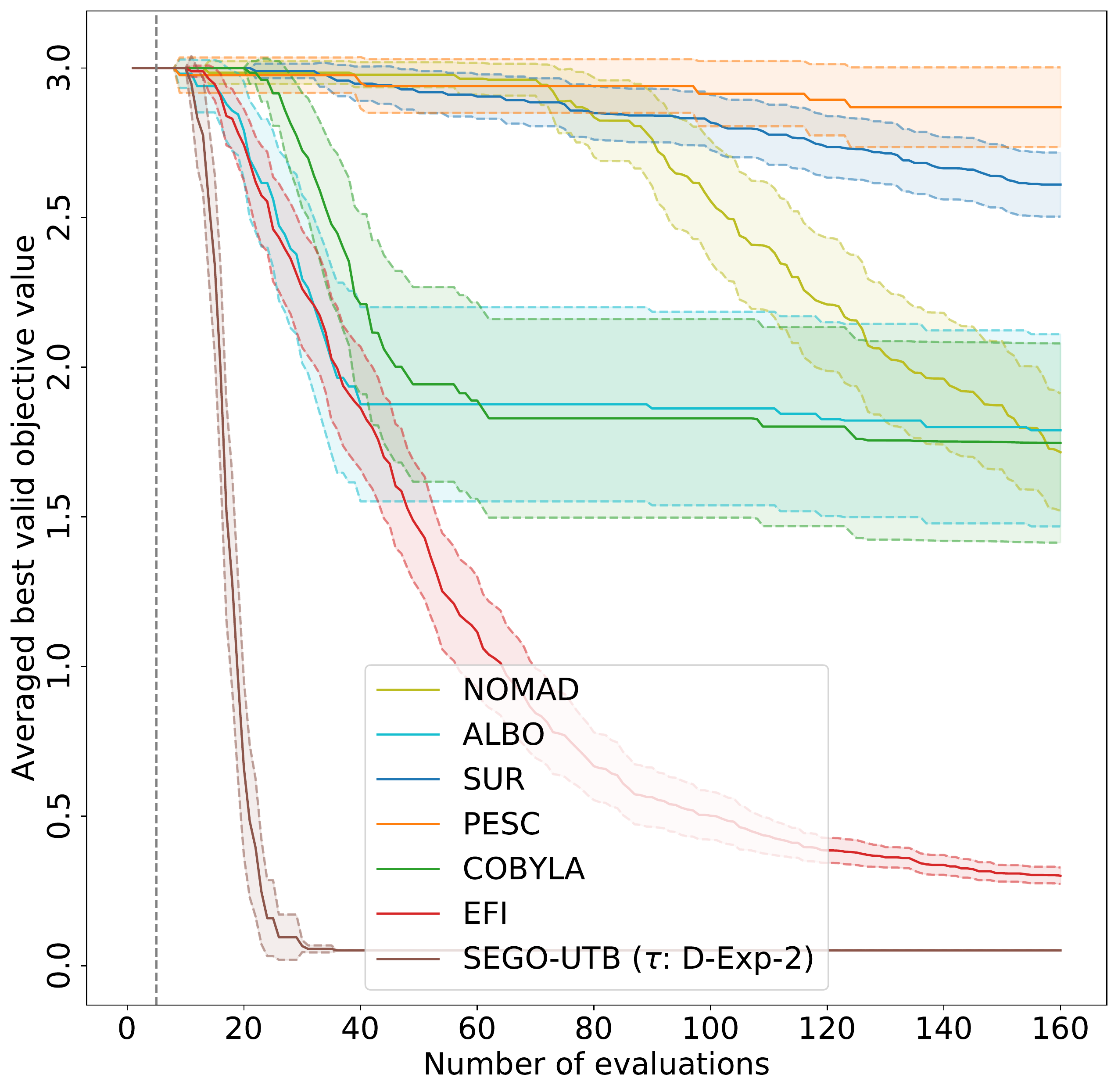}}
    \caption{Convergence plots for the LAH problem NOMAD, COBYLA and BO solvers, considering the two levels of constraints violation $10^{-4}$ and $10^{-2}$. The vertical grey-dashed line outlines the number of points in the initial DoEs.}
    \label{fig:lah}
\end{figure}

For the GBSP problem (which has mixed constraints), the obtained results are given in \change{Figures \ref{fig:gbsp_sego} and \ref{fig:gbsp}.}
SEGO displays good performance during the early stages of the optimization but does not reach the global minimum of the GBSP problem \change{for both tolerances on the constraints violations}.
The SEGO-UTB variants are able to explore better the feasible domain (with a slower convergence rate compared to SEGO) and converge to the global minimum for all runs (see Figure \ref{fig:gbsp_sego:2}) considering the constraints violation of $10^{-2}$.
\change{Figure \ref{fig:gbsp_sego:4} shows that the use of a stricter  tolerance on the constraints violation deteriorates the convergence of the SEGO-UTB solvers although they still converge close to the SEGO solution.}
In comparison with the other solvers, see Figure \ref{fig:gbsp}, the SEGO-like solvers are outperforming (by far) all the tested solvers.\change{
We note also that the performances of the solvers ALBO, NOMAD and COBYLA turn to be very sensitive to the regarded value of the constraints violation tolerance. In fact, ALBO, NOMAD \changeb{and COBYLA} display better performance when using a large tolerance, such performances get worst when the tolerance on the violation of the constraints is stricter.
\removeb{On the contrary, the solver COBYLA is able to improve its performances (but not as good as SEGO-UTB) when a high tolerance on the constraints is used.}
Overall, the obtained results,} in particular, confirm the efficiency of SEGO-UTB when handling problems with both equality and inequality constraints.
Again, see Figure \ref{fig:lah}, the superior performance of SEGO-UTB is confirmed on the LAH problem (which is mixed constrained) with the two levels of constraints violations.

\begin{figure}[p]
    \centering
    \subfloat[$\epsilon_c = 10^{-2}$. \label{fig:lah_sego:2}]{\includegraphics[width=0.5\textwidth]{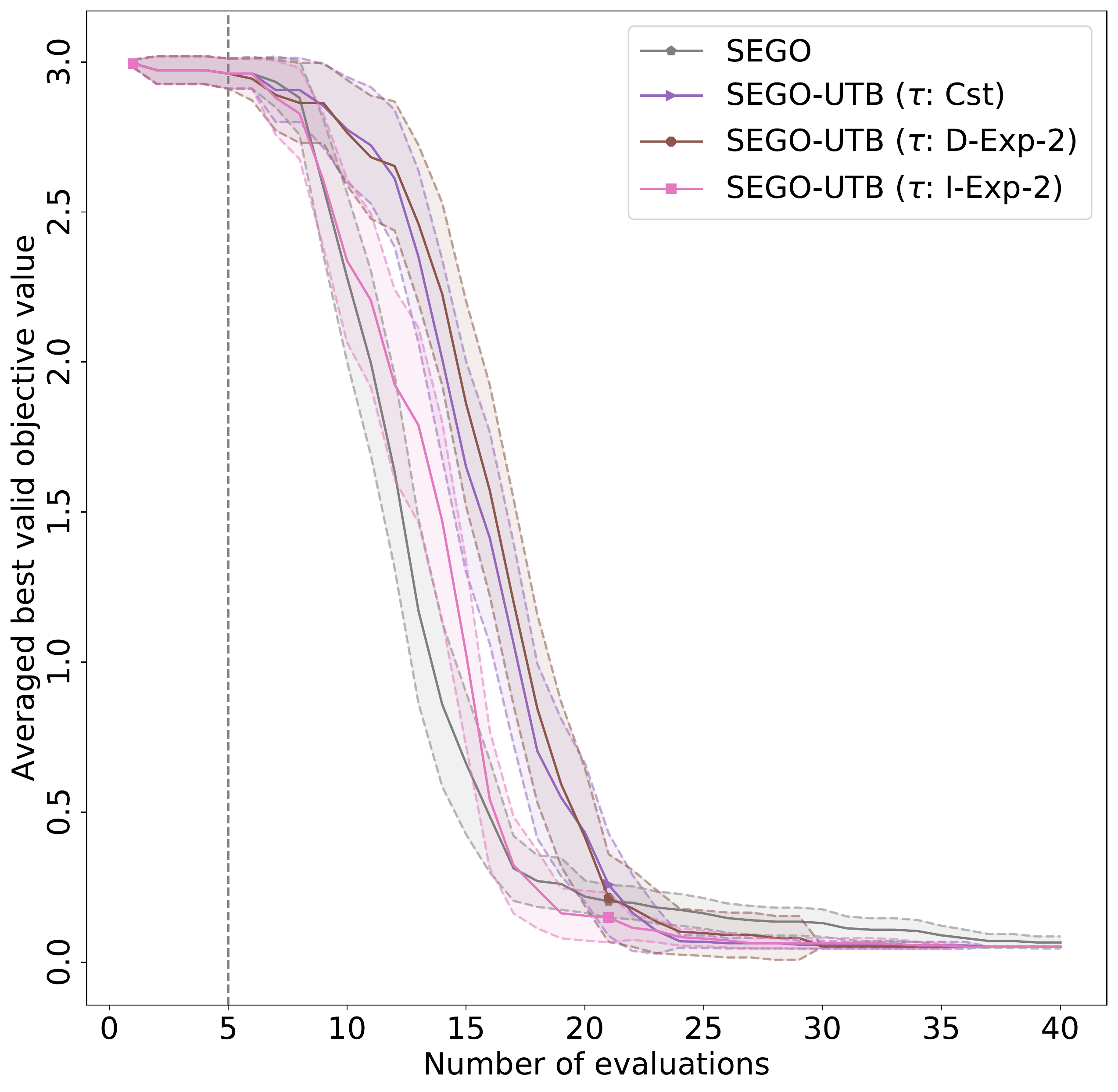}}
    \subfloat[$\epsilon_c = 10^{-4}$. \label{fig:lah_sego:4}]{\includegraphics[width=0.5\textwidth]{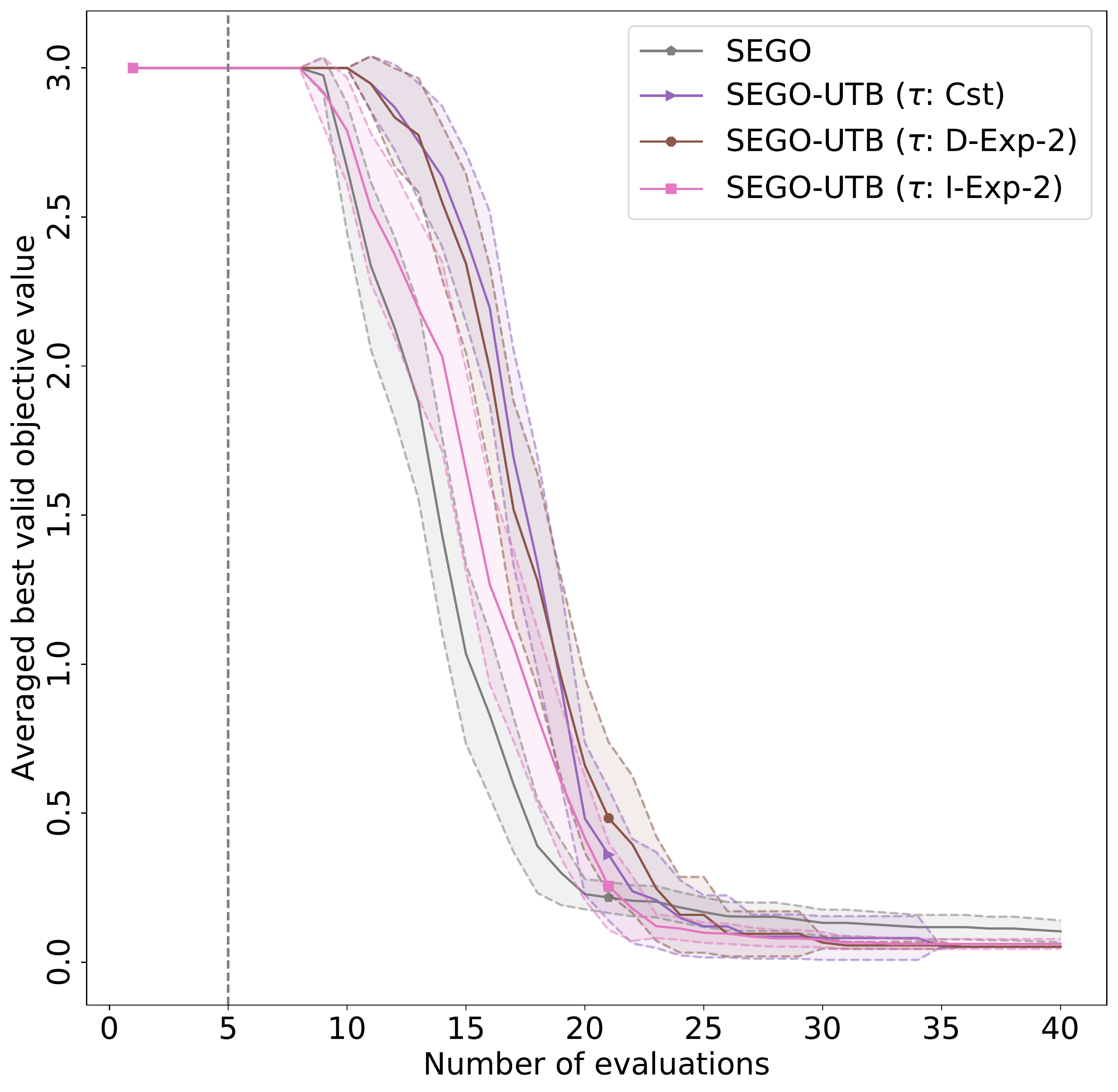}}
    \caption{Convergence plots for the LAH problem on the SEGO-like solvers, considering the two levels of constraints violation $10^{-4}$ and $10^{-2}$. The vertical grey-dashed line outlines the number of points in the initial DoEs.}
    \label{fig:lah_sego}
\end{figure}

In this subsection, the obtained results showed the potential of including the uncertainty while modeling the constraints during the optimization procedure.
In particular, one was able, using convergence plots, to show the good performance of the proposed method compare to the state-of-the-art BO solvers on four known test problems. 
In the next subsection, we will use data profiles (adapted to our constrained setting) to confirm the efficiency of the proposed method, all using a larger test pool formed by 29 test problems.

\subsection{Comparison results using data profiles}

\subsubsection{Problem instances}
Our benchmark set is composed of 29 optimization problems, from \cite{Mezura-MontesEmpiricalanalysismodified2012,PichenyBayesianoptimizationmixed2016,RegisEvolutionaryprogramminghighdimensional2014,ParrReviewefficientsurrogate2010}, which are mixed constrained (up to $38$ equality and inequality constraints) of $2$ to $10$ design variables.
A detailed description of the test problems is given in \ref{app:over}.
Our test set can be divided into two classes.
The first one, referred as the \textit{weakly non-linear constrained} (WNLC) set, is composed of 16 constrained optimization problems where the constraints are linear or quadratic.
The second class is composed of the rest of  the problems in our test pool, this class is referred to the \textit{highly non-linear constrained} (HNLC) problems.

Due to the stochastic nature of the BO solvers, we create instances for the tested problems.
In fact, the obtained results for a given problem may depend on the choice of the chosen initial DoEs.
Thus, an instance of the problem is related to the choice of an initial DoE.
In our case, we generate 10 different initial DoEs for each problem which leads to the creation of 290 problem instances. 

\subsubsection{Data profiles}
\label{sssec:plan_DP}

Data profiles \cite{more2009benchmarking} are designed for derivative-free optimization, to show how well a solver performs, given some computational budget, when asked to reach a specific reduction in the objective function value, measured in our case by 
\begin{equation}
    \label{eq:cv_test}
    \tilde f(x^{(l)}) - f_{\mbox{opt}}  \leq \epsilon ( | f_{\mbox{opt}}| + 1 ),
\end{equation}
where $\epsilon \in [0,1]$ is the required level of accuracy, $f_{\mbox{opt}}$ represents the best objective value
found (within the feasible domain) by all solvers tested for a specific problem and within a given maximal computational budget.
$\tilde f(x^{(l)})$ is set to the value of the objective function at the iteration $l$ if $x^{(l)}$ satisfies the constraints, and set to $+\infty$ otherwise. We note that we had to adapt the data profiles \cite{more2009benchmarking} to our constrained setting; for that reason, we are using the same convergence test as proposed in \cite{YDiouane_SGratton_LNVicente_2015}.

Data profiles plot the percentage of problems solved by the solvers under consideration for different values of the computational budget.
Let $\mathcal{S}$ be the set of the tested solvers and $\mathcal{P}$ the set of the problem instances.
A data profile is computed, for each solver $s \in \mathcal{S}$, as the percentage of the problem instances that can be solved within $\kappa$ objective function evaluations:
\begin{equation*}
   \frac{1}{|\mathcal{P}|}~\mbox{size}\left \{ p \in \mathcal{P}~~:~~ \frac{t_{p,s}}{d_p} \le \kappa  \right\},
\end{equation*}
where $d_p$ is the dimension of the problem instance $p \in \mathcal{P}$,  $t_{p,s}$ is the number of function evaluations required by solver $s \in \mathcal{S}$ on problem instance $p$ to satisfy the convergence test \eqref{eq:cv_test} for a given tolerance $\epsilon$.
If the convergence test is not satisfied after the maximum budget of function evaluations, $t_{p,s}$ is set  to $+\infty$.
The units budget are expressed with $d_p$ to allow the combination of problems of different dimensions in the same profile.

We used in our experiments a maximal computational budget consisting of $40d_p$ function evaluations, as we are primarily interested in the behavior of the algorithms for problems where the evaluation of the objective function is expensive.
For the level of accuracy \change{used in the convergence test \eqref{eq:cv_test}, we set} $\epsilon=10^{-3}$.

\subsubsection{Results}

We note that, due to the dimension of the tested problems and the number of constraints, all the tested BO optimizers (except the SEGO-like solvers) did not give good results and were computationally very expensive.
\change{Table \ref{tab:time_4pb} shows the CPU-time average using 10 runs for the BO solvers on five problems using a budget of $40d$ maximum function evaluations. Clearly, one can see that the solvers ALBO, SUR, PESC and EFI run on the G07 problem are very consuming in terms of CPU-time. Thus, those solvers cannot be tested on the all 29 problems in a reasonable time. We note also that the solver SUR cannot handle more than four constraints (which is the case of many problems in our test bed) in the DiceOptim implementation.}
For \change{all these reasons}, we consider that the BO solvers, ALBO, SUR, PESC and EFI are not adapted for our test problems, the presented results include only the solvers: NOMAD, COBYLA, SEGO and SEGO-UTB. 

\begin{table}[!htb]
    \bigcentering
    \caption{CPU-time average (in seconds) using 10 runs for the BO solvers on five problems using a budget of $40d$ maximum function evaluations.}
    \change{
    \begin{tabular}{ c || c | c | c | c | c | c}
        \toprule
        Problem & SEGO & SEGO-UTB & ALBO & SUR & EFI & PESC \\ 
        \midrule
        LAH & 203.64 & 2 708.99 & 5 131.97 & 2 592.32 & 1 212.04 & 2 729.64 \\
        GBSP & 164.84 & 266.76 & 608.33 & 1 510.61 & 497.04 & 777.06 \\ 
        LSQ & 157.43 & 177.04 & 833.83 & 417.97 & 207.82 & 592.70 \\ 
        MB & 79.10 & 93.52 & 328.41 & 1 059.24 & 535.18 & 615.98 \\ 
        G07 & 2 098.86 & 4 245.55 & 611 025.72 & $-$ &  1 195 346.89 & 55 755.87 \\
        \bottomrule
    \end{tabular}}
    \label{tab:time_4pb} 
\end{table}

For clarity reasons (similarly to Section \ref{subsec:cv_plots}), we test different evolution \change{strategies} for the constraints learning rate within the SEGO-UTB solver, but only the best among the non-decreasing and decreasing strategies are kept.
The constant constraint learning rate evolution is also included in our tests.
The complete results \change{for decreasing and non-decreasing strategies} are given in \ref{app:res}. \change{Note that all these tests are performed for the two levels of constraints violations $10^{-2}$ and $10^{-4}$.}

\begin{figure}[!htb]
    \centering
    \subfloat[$\epsilon_c=10^{-2}$. \label{fig:DP_all:2}]{\includegraphics[width=0.5\textwidth]{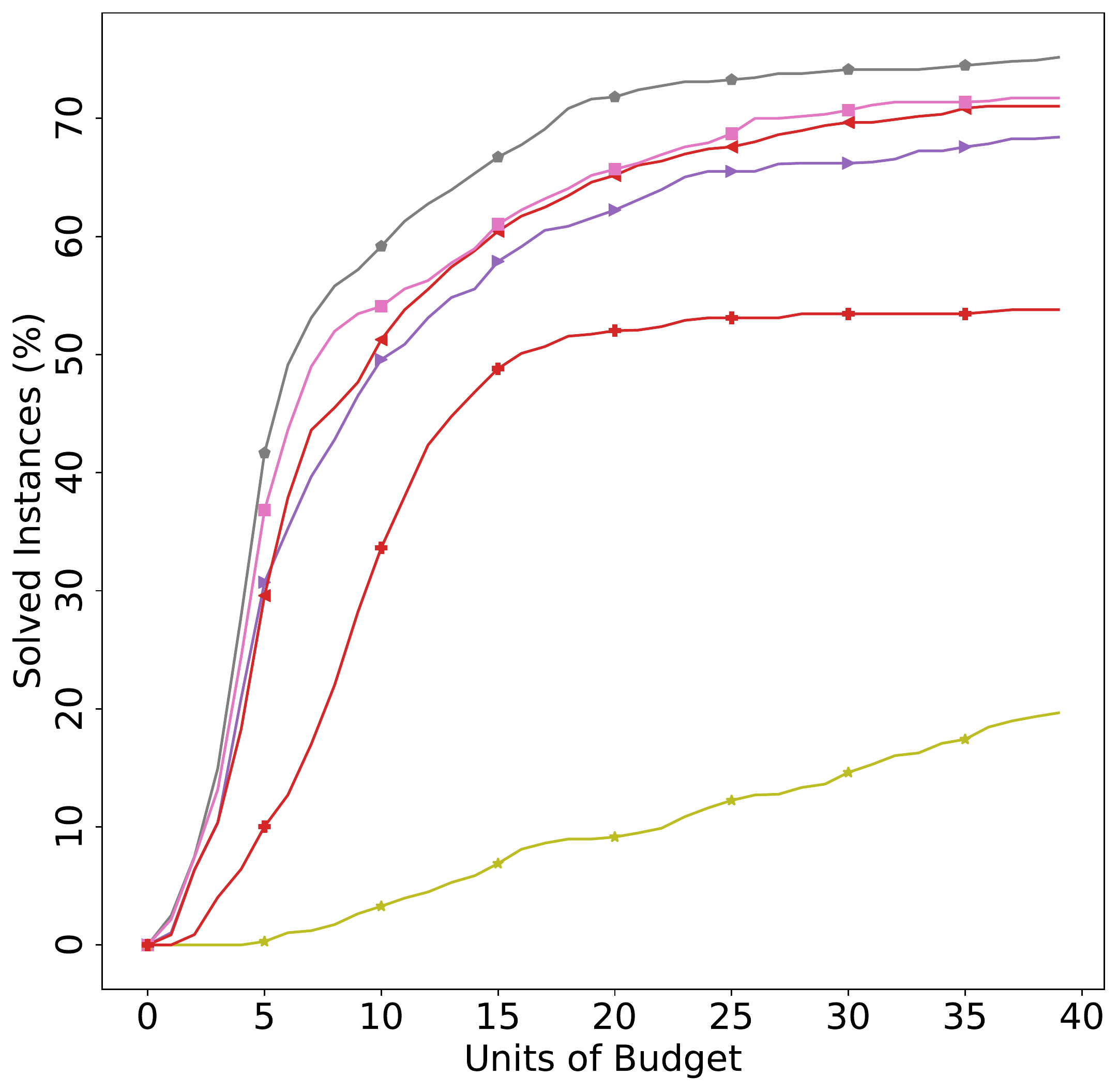}}
    \subfloat[$\epsilon_c=10^{-4}$. \label{fig:DP_all:4}]{\includegraphics[width=0.5\textwidth]{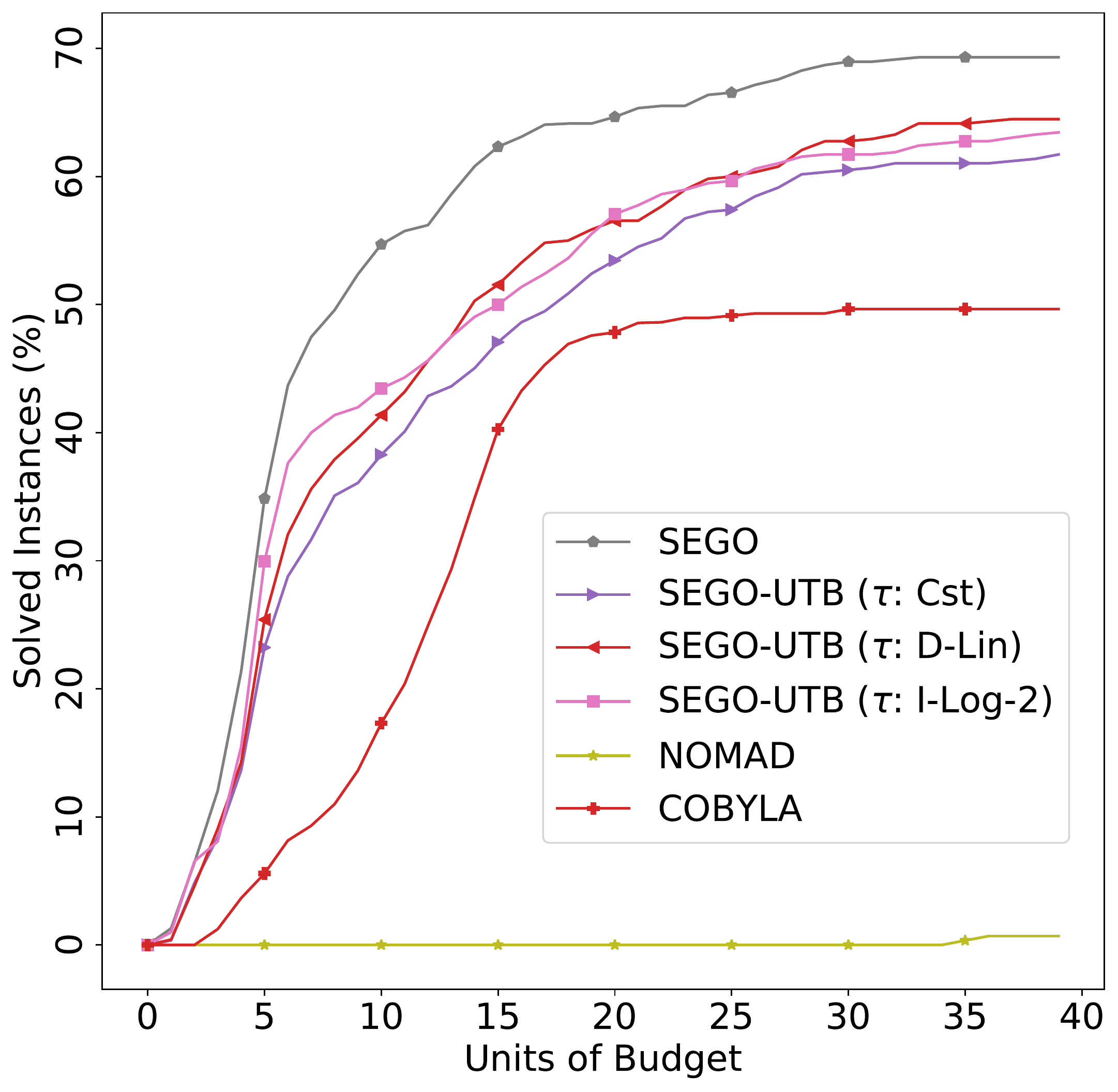}}
    \caption{Obtained data profiles of the 29 problems, considering the two levels of constraints violation $10^{-4}$ and $10^{-2}$.}
    \label{fig:DP_all}
\end{figure}

    We now comment on the data profiles obtained by the regarded solvers.
    Figure~\ref{fig:DP_all:2} depicts the obtained data profiles when considering all the tested problems considering the constraints violation $\epsilon_c=10^{-2}$.
    As in the convergence plots tests, the SEGO-like solvers appear as the best.
    In fact, using a maximal budget, the SEGO-like solvers are able to solve around \changeb{70\%}  the tested instances, COBYLA solves  \changeb{53\%} and NOMAD around 20\%.
    For smaller units of budget, the gap between the SEGO-like solvers \changeb{is similar}.
    \changeb{Note that, for large budgets, SEGO displays slightly better performance compared to the other SEGO-UTB solvers.
    Typically, at the end of the optimization, SEGO is able to solve 75\% of the instances while SEGO-UTB ($\tau:$ I-Log-2) is solving 72\%. The SEGO-like solvers outperform COBYLA and NOMAD even when a stricter tolerance on the constraints violation is used, see Figure \ref{fig:DP_all:4}.  
    For instance, with the maximal budget and using a tolerance $\epsilon_c=10^{-4}$ on the violation of the constraints, SEGO is outperforming all the tested solvers by solving 70\% of the tested instances.
    In all our tests, the use of stricter tolerance on the constraints violation reduced the percentage of the instances solved by each solver.
    We acknowledge also that NOMAD is not well adapted for equality constraints, this, in particular, explains the bad performance of NOMAD when a stricter tolerance on the violation of the constraints is used.}
    
\begin{figure}[p]
    \centering
    \subfloat[$\epsilon_c=10^{-2}$. \label{fig:dp_wnlc:2}]{\includegraphics[width=0.5\textwidth]{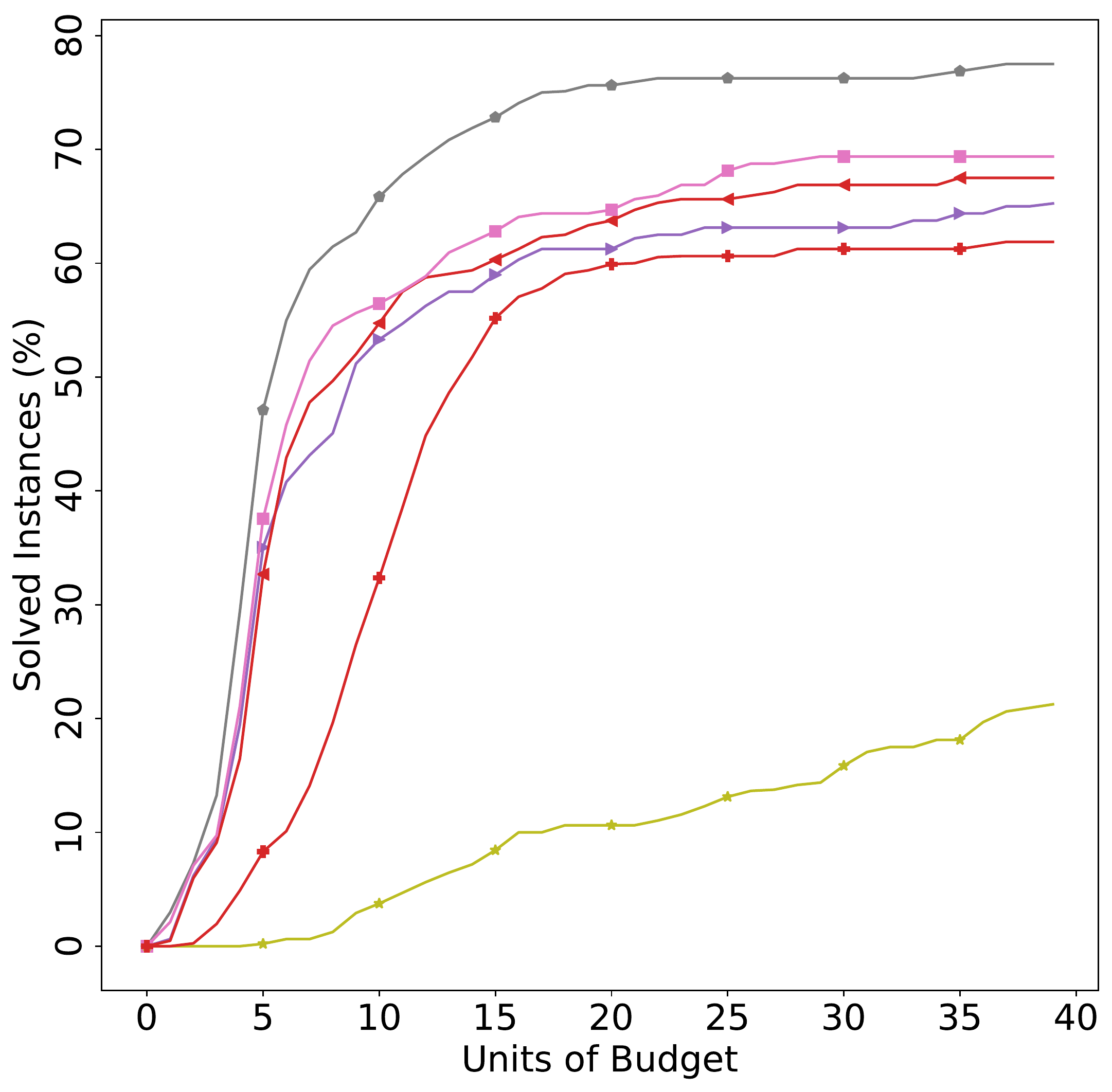}}
    \subfloat[$\epsilon_c=10^{-4}$. \label{fig:dp_wnlc:4}]{\includegraphics[width=0.5\textwidth]{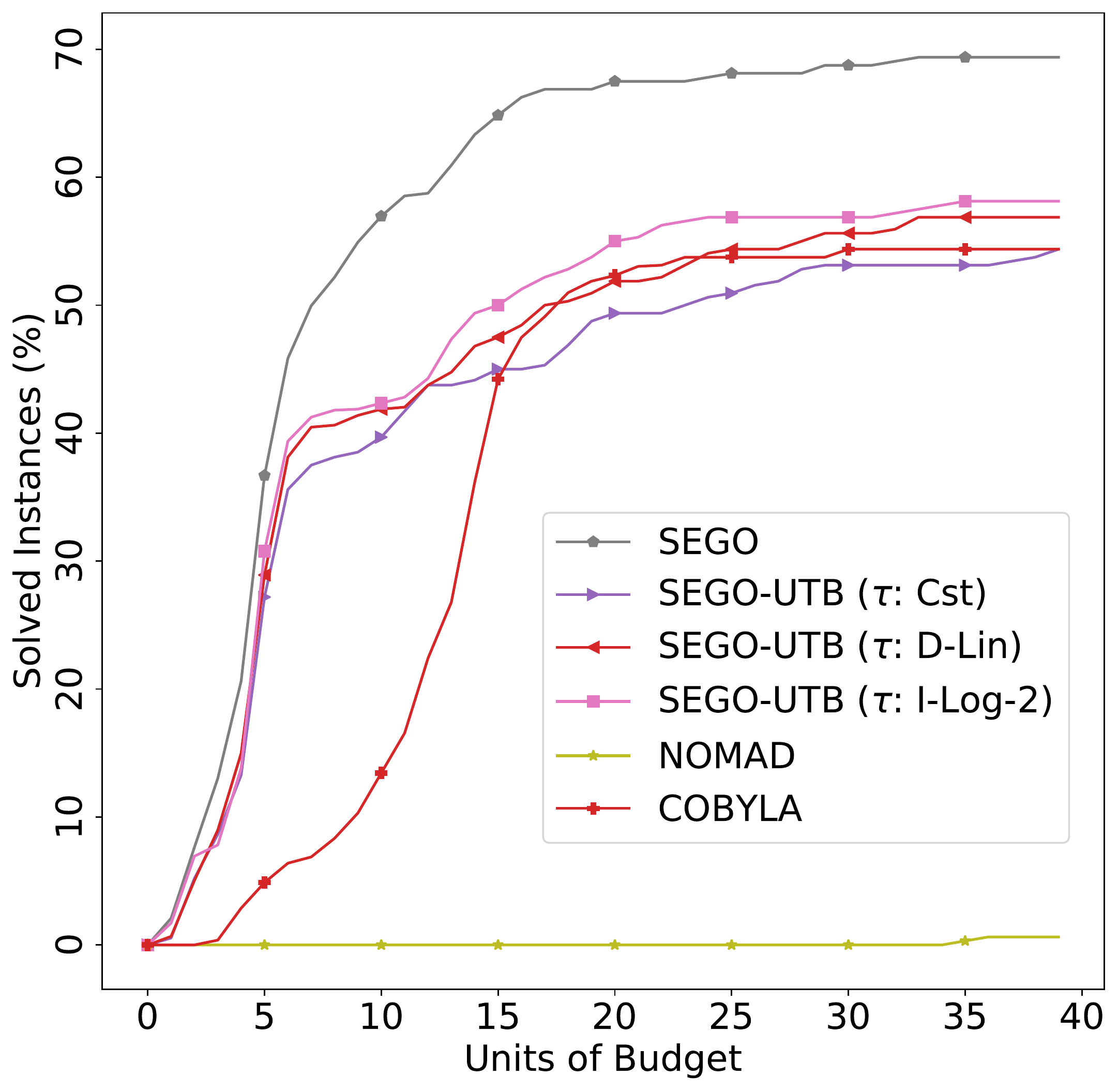}}
    \caption{Obtained data profiles of the WNLC problems, considering the two levels of constraints violation $10^{-4}$ and $10^{-2}$.}
    \label{fig:dp_wnlc}
\end{figure}

\begin{figure}[p]
    \centering
    \subfloat[$\epsilon_c=10^{-2}$. \label{fig:dp_hnlc:2}]{\includegraphics[width=0.5\textwidth]{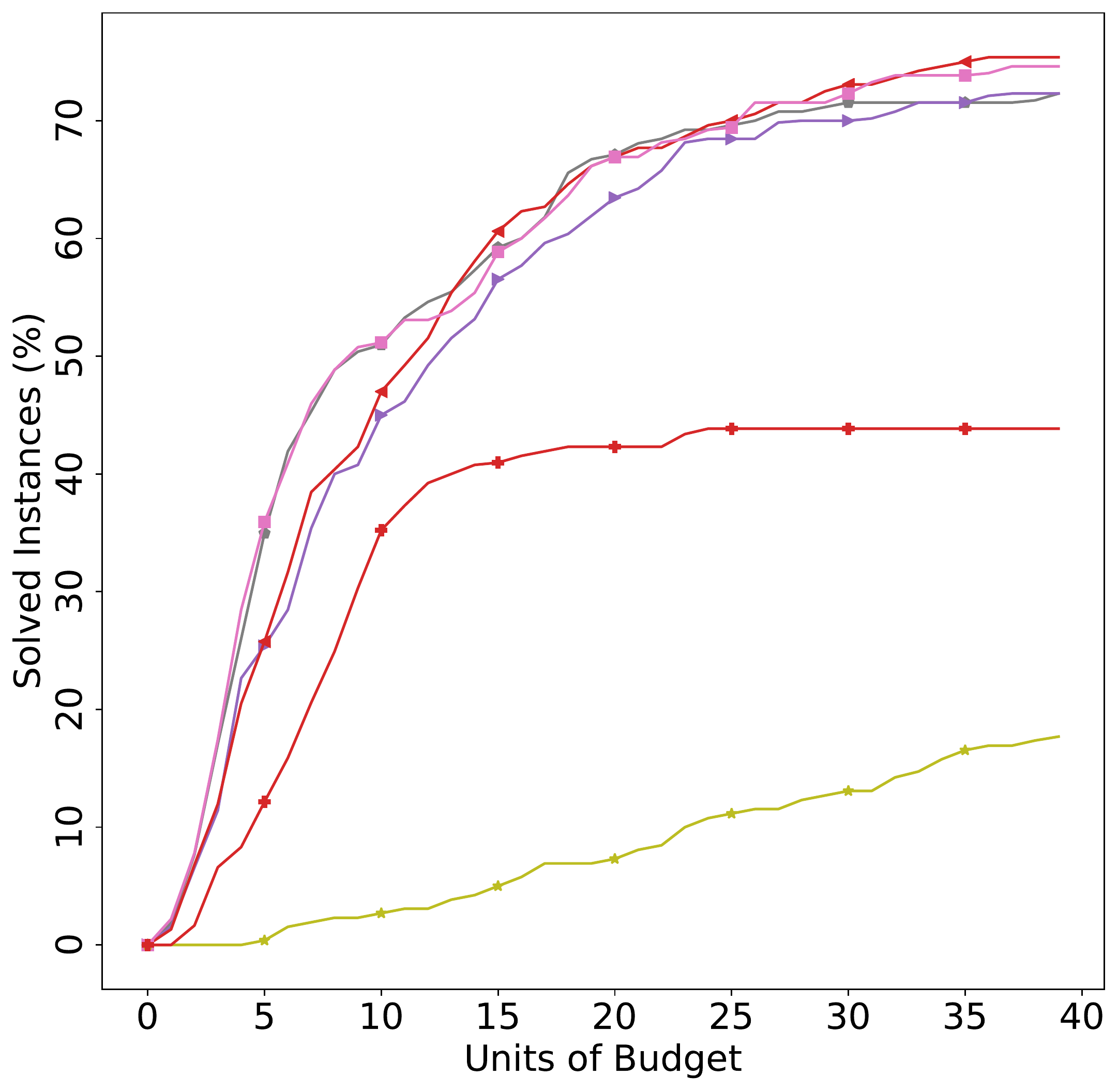}}
    \subfloat[$\epsilon_c=10^{-4}$. \label{fig:dp_hnlc:4}]{\includegraphics[width=0.5\textwidth]{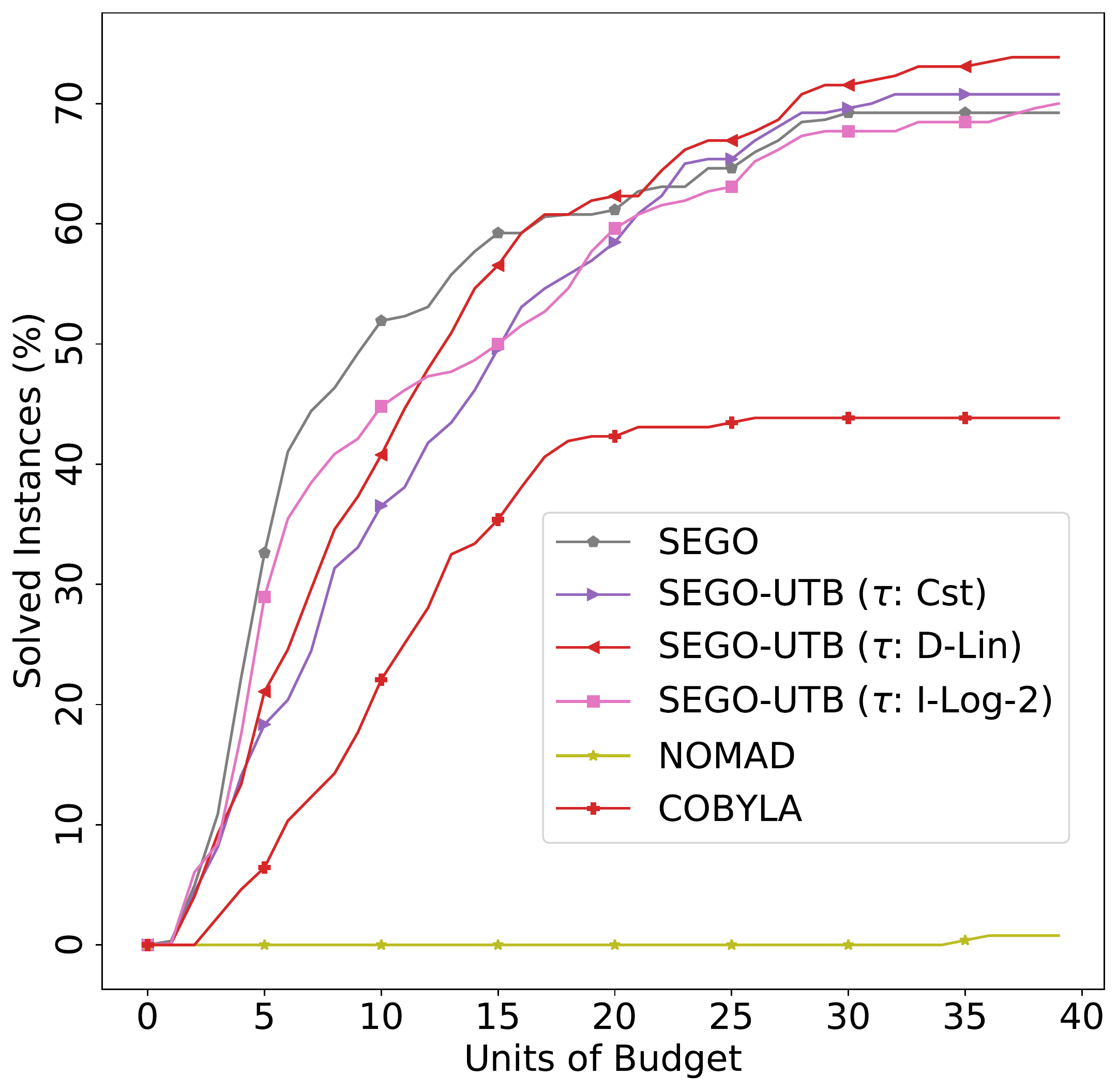}}
    \caption{Obtained data profiles of the HNLC problems, considering the two levels of constraints violation $10^{-4}$ and $10^{-2}$.}
    \label{fig:dp_hnlc}
\end{figure}

    For a deeper analysis, we plot two data profiles respectively using the two problem classes WNLC and HNLC
    The obtained profiles are depicted in Figures \ref{fig:dp_wnlc} and \ref{fig:dp_hnlc}.
    Clearly, one can see that SEGO is solving more instances than SEGO-UTB on WNLC problems with both constraints violations. 
    On the contrary, SEGO-UTB solvers are showing similar slightly better performances on the HNLC problems.

Overall, one concludes that including the uncertainties of the constraints within the SEGO framework turns to offer a better exploration of the feasible domain, in particular when the constraints present high non-linearities.

\section{An application to Aircraft Design}
\label{sec:FAST}

\begin{figure}[hbt]
	\centering
	\includegraphics[width=0.7\textwidth, keepaspectratio]{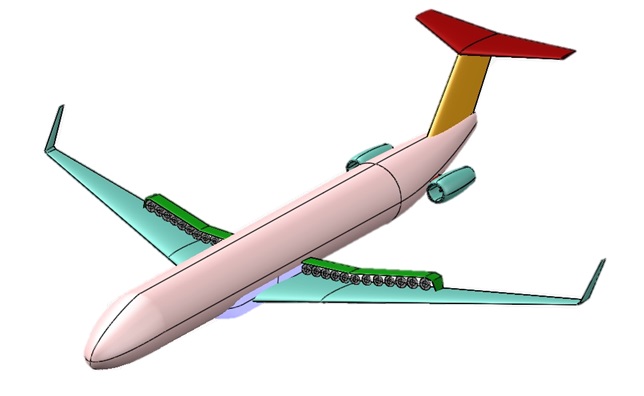}
	\caption{A Hybrid aircraft concept, featuring distributed propulsion ~\cite{sguegliafasthybrid2018}.}
	\label{fig:hybrid_aircraft}
\end{figure}

This section presents an aircraft design application where the SEGO solvers lead to a significant improvement.
We target to design a hybrid aircraft, featuring distributed electric propulsion \change{\cite{sguegliafasthybrid2018, sguegliaExplorationDimensionnementPriorites2019, sguegliafasthybrid2020}}, the related concept of such aircraft is shown in Figure~\ref{fig:hybrid_aircraft}.
Its main feature is the propulsive chain, which is made up of turbo-generators and batteries, that supply electric power to the set of distributed ducted fans, placed along the wing. The two gas turbines are evident at the rear of the aircraft. Meanwhile, batteries are not shown since they are placed within the cargo bay.
To reduce emissions, the aircraft is able to fly at least to 3000~ft in fully electric mode and it is designed to carry 150 passengers for a range of 1200~nmi.

The \textit{fixed-wing aircraft sizing tool} (FAST) \cite{fastmainpaper2017} is used to explore this aircraft concept and is fully coded in Python. 
It is based on engineering methods, to have reliable results with low computational cost~\cite{roskambookpartI}.
Some modules have been modified to consider the features introduced by the hybrid chain \cite{sguegliafasthybrid2018, sguegliafasthybrid2020}. We note also that the unconventionality of the concept, showing more interaction between the disciplines, (i.e. for the aerodynamics and the propulsion), makes the MDO a powerful tool to explore their sizing \cite{graybli2018, freemanhybrid2014}.
Thus it is a relevant case study for the BO solvers already introduced in this paper.

\subsection{The aircraft optimization problem}

The optimization problem consists in minimizing the total energy consumption (e.g. sum of the fuel and batteries energy) with respect to the geometry (e.g. geometrical parameters defining wing, horizontal and vertical tail). This objective will be denoted by TEC.

In particular, the wing is defined by surface $S_w$, aspect ratio $AR_w$, wing position $x_w$ and sweep angle, computed at 25\% of the chord, $\Lambda_{25,w}$.
The horizontal and vertical tail are defined by their surfaces $S_{HT}$ and $S_{VT}$, aspect ratios $AR_{HT}$ and $AR_{VT}$ and sweep angles $\Lambda_{25,HT}$ and $\Lambda_{25,VT}$, computed at 25\% of the chord.
To consider the propulsive aspects, battery volume $\tau_b$ is added as design variable too. 
Finally, the cruise altitude $h_{cruise}$ belongs to design variables vector, to ensure that the aircraft flies at the maximum aerodynamic efficiency. 
With this notation, it is possible to write the design vector as $x~=~\left[x_w, S_w, AR_w, \Lambda_{25,w}, S_{HT}, AR_{HT}, \Lambda_{25,HT}, S_{VT}, AR_{VT}, \Lambda_{25,VT}, \tau_b, h_{cruise}\right]^\top$ that contains sub-vectors of the geometrical parameters for the wing, the horizontal and vertical tails.

The design space exploration is reported in Table~\ref{tab:design_space_hybrid_aircraft} where the values are taken within common data for the type of aircraft considered \cite{roskambookpartI}.
\begin{table}[!ht]
	\centering
	\caption{Design optimization space definition.}
	\vskip 0.1in
	\begin{tabular}{l l r r | l l r r}
		\toprule
		\textbf{Variable} & & \textbf{Min.} & \textbf{Max.} & \textbf{Variable} & & \textbf{Min.} & \textbf{Max.}\\
		\midrule
		$h_{cruise}$ & [kft] & 27 & 35 & $S_{HT}$ & [\si{\square\meter}] & 20 & 80\\
		$\tau_b$ & [\si{\cubic\meter}] & 1.5 & 3.0 & $AR_{HT}$ & & 3 & 5 \\
		$x_w$ & [\si{\meter}] & 15 & 18 & $\Lambda_{25,HT}$ & [\si{\degree}] & 25 & 45 \\
		$S_w$ & [\si{\square\meter}] & 100 & 130 & $S_{VT}$ & [\si{\square\meter}] & 20 & 50 \\
		$AR_{w}$ & & 9 & 12 & $AR_{VT}$ & &1 & 2 \\
		$\Lambda_{25,w}$ & [\si{\degree}] & 20 & 45 & $\Lambda_{25,VT}$ & [\si{\degree}] & 30 & 45\\
		\bottomrule
	\end{tabular}
	\label{tab:design_space_hybrid_aircraft}
\end{table}

The problem is subject also to constraints.
To ensure the feasibility of the aircraft design: the wing has to store enough fuel for the whole mission ($MFW\geq m_f$, with $MFW$ maximum fuel weight and $m_f$ the mission fuel) and generate enough lift in approach condition for landing ($C_{L_{\max}}\geq C_{L_{app}}$, with $C_{L_{\max}}$ maximum and $C_{L_{app}}$ approach lift coefficient).
Horizontal tail is designed to ensure takeoff rotation, meaning to have positive pitching moment for every center of gravity position~\cite{RaymerAircraftDesignConceptual2018} ($\mathcal{M}_{takeoff}=0$).
Vertical tail is instead designed to counterbalance the fuselage yaw moment in cruise~\cite{RaymerAircraftDesignConceptual2018} ($\mathcal{N}_{cruise}=0$).
The batteries are subject to two constraints, related to power and energy requirements.
The first demands that they produce enough power at takeoff ($P_b\geq P_{takeoff}$), and the second ensures that, at the end of the flight, there is still a 20\% energy available. 
The parameter that controls the energy consumption is the state of charge SoC, defined as the ratio between the energy consumed and the total energy stored.
With the SoC definition, the second condition can be written as $\text{SoC}_{fin}\geq 0.20$\change{.}
The $\text{SoC}_{min}$ limit is the safety margin for most of batteries, to not damage the system~\cite{tremblaybatt2009}.
Another constraint related to stability is the static margin $SM$, which has to be included between $SM_{min}=0.05$ and $SM_{max}=0.10$, according to certification~\cite{RaymerAircraftDesignConceptual2018}. 

Regarding the restrictions coming from airport configuration,
for the type of aircraft considered, the takeoff field length TOFL must not exceed $\text{TOFL}_{max}=2.2$~\si{\kilo\meter} and the wing span $b_w$ is below $b_{w_{max}}=36$~\si{\meter}~\cite{debarros1997}.
Finally, the last constraint is given to ensure that the cruise altitude is chosen to maximize the efficiency, that is $C_{L_{cruise}}=C_{L_{opt}}$, where the left side is the cruise lift coefficient and the right side is the lift coefficient at which the maximum aerodynamics efficiency occurs.
To sum up, for the regarded aircraft design problem, the equality constraints are represented by $\bm{h}= [\mathcal{M}, \mathcal{N}, C_{L_{cruise}}-C_{L_{opt}}]^\top$ while the inequality constraints are depicted as follows: $\bm{g}=[b_{w_{max}}-b_w, MFW-m_f, C_{L_{\max}}-C_{L_{app}}, \text{SoC}-\text{SoC}_{min},P_b-P_{takeoff}, \text{TOFL}_{max}-\text{TOFL}, SM-SM_{min},~SM_{max}-SM]^\top$.
We thus obtain the following optimization problem:
\begin{equation*}
    \min\left\{\mbox{TEC}(\bm{x}) ~~\mbox{w.r.t.}~~ \bm{x} \in \mathbb{R}^{12} ~~\mbox{s.t.}~~ 0 \leq \bm{g}(\bm{x}) \in \mathbb{R}^{8} ~\mbox{and}~ 0 = \bm{h}(\bm{x}) \in \mathbb{R}^3 \right\}
   \end{equation*}

We note that in the numerical tests presented in Section~\ref{subsec:cv_plots} we tolerate a violation of the constraints up to $10^{-2}$. However, in the regarded aircraft optimization design problem, the tolerated violation on the constraints $\bm{h}$ and $\bm{g}$ is driven by the physical properties of the problem  \cite{roskambookpartI}, respectively, as follows $\epsilon_h=[10^{-2},10^{-2},~ 5000]^\top$ and $\epsilon_g=[10^{-2},100,10^{-2},10^{-2},1000,100, 10^{-2},10^{-2}]^\top$.
If a point, during the optimization process, exceeds the tolerated constraints violation, the objective function is penalized with the value $4.10^5$. All the other implementation choices are kept as explained in Section~\ref{sssec:plan_DP} for all the tested solvers. The presented convergence plot for each solver is built using $10$ runs.

\subsection{Results}

Similarly to the reported results in Section~\ref{sssec:plan_DP}, only the SEGO-like solvers, COBYLA and NOMAD will be included in the presented comparison. The other BO solvers, due to the dimension of the regarded problem, were very consuming in CPU and did not lead to any acceptable results in a reasonable time. 

\begin{figure}[hbt]
	\centering
	\includegraphics[width=0.7\textwidth, keepaspectratio]{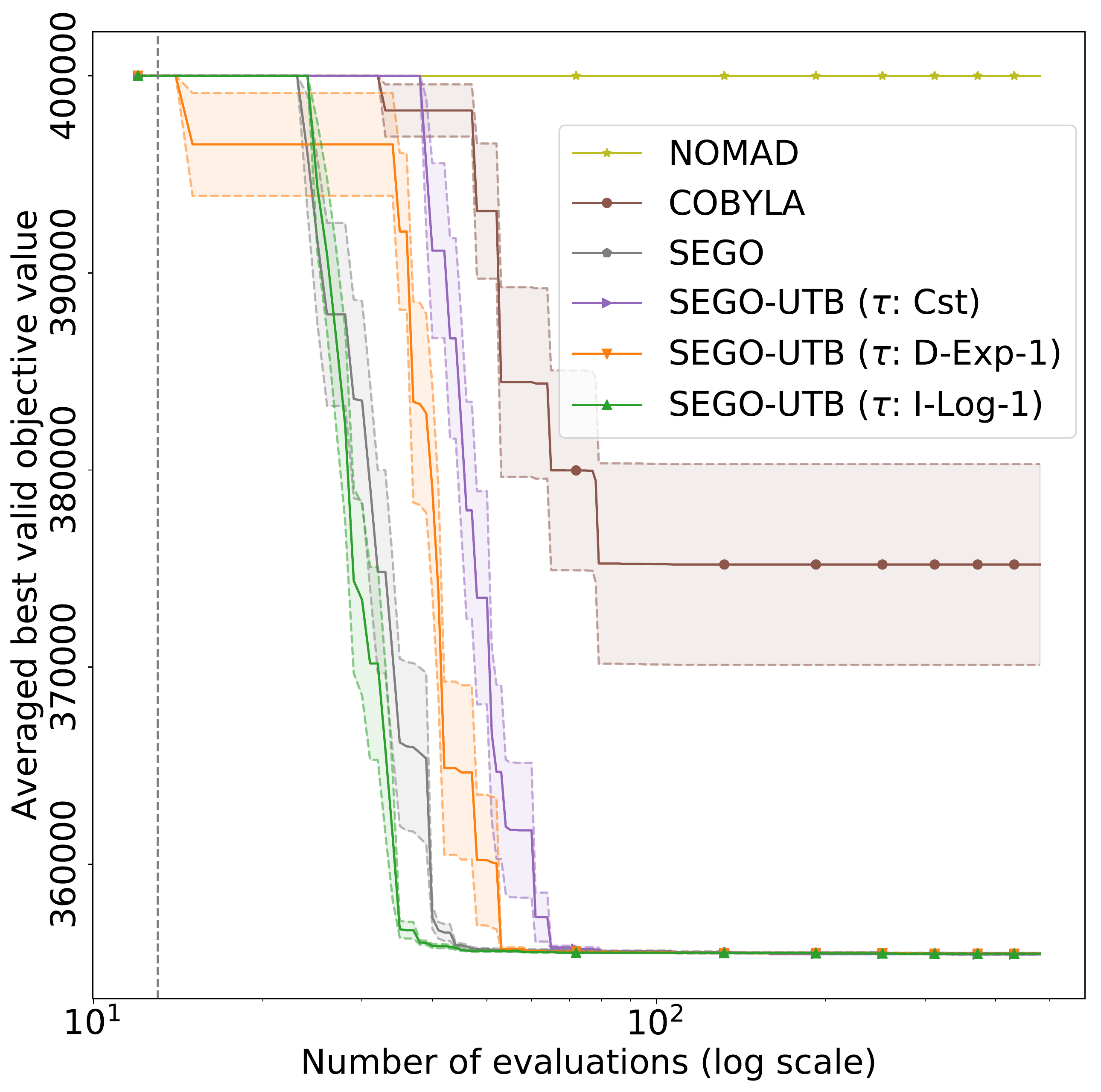}
	\caption{Convergence plots for the FAST problem. The vertical grey-dashed line outlines the number of points in the initial DoEs.}
	\label{fig:fast:cv}
\end{figure}

Figure~\ref{fig:fast:cv} depicts the obtained convergence plots while solving the FAST optimization problem. One can see that the SEGO-like solvers are outperforming COBYLA and NOMAD. In fact, NOMAD was never able to find any feasible point for the 10 optimization runs performed. 
The high standard deviation of COBYLA implies that it does not converge to the same optimum for each run. Among the SEGO-like solvers, SEGO-UTB ($\tau$: I-Log-1) appears to converge the fastest to the best solution. For sake of visualization, only the three best SEGO-UTB solvers are presented, the complete obtained results are given in Figure \ref{fig:fast_ad}.  

\begin{figure}[hbt]
	\centering
	\includegraphics[width=0.9\textwidth, keepaspectratio]{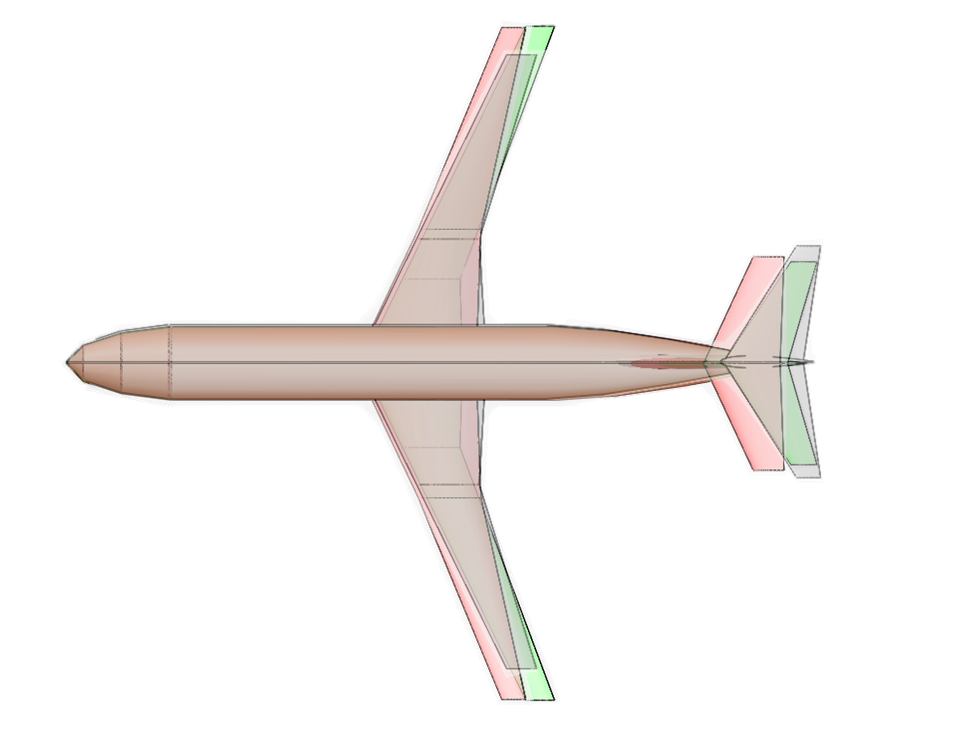}
	\caption{One of the obtained plan-forms using SEGO-like solvers (in red) and COBYLA (in green) as well as the baseline solution (in grey) of the FAST aircraft.}
	\label{fig:fast:plan}
\end{figure}

The obtained plan-forms (related to the FAST problem) by the SEGO-like solvers and COBYLA as well as the baseline solution are all displayed in Figure~\ref{fig:fast:plan}. The differences in the plan-forms are clearly observable. Physically, we note that both SEGO-like solvers and COBYLA solutions present a smaller sweep angle for the wing than the baseline one.
The value of sweep is suggested by common design books to reduce the compressibility effects~\cite{roskambookpartI}, but despite the transonic regime (we recall that the Mach number is 0.78), wave drag is not yet diverging, and thus the optimum goes towards a reduction of the sweep to improve aerodynamics. 
The minimum found by COBYLA corresponds to the best aerodynamic solution between the three; however the increased sweep leads to a greater wing mass which is penalizing.
The minimum found by SEGO represents a balance between aerodynamics and mass reduction.
As a consequence, the tails are reduced because of the reduction in mass and wing area (snowball effects). 

In what comes next, we will use the so-called ``parallel plots'' to illustrate the behaviour of the tested solvers with regard to the exploration of the design space. In a parallel plot, we display the values of specific targeted data during the optimization process for a given solver (e.g., the values of the explored design variables). In our case, we depict in the parallel plots (from bottom to top) the required number of iterations to converge, the 12 design variables, the objective function value and last the constraints violation. On the top of that, we will use the red color to refer to the reference design (i.e. the best feasible design found so far by all the tested optimizers, here SEGO-UTB ($\tau$: Cst)), in black the optimum design found by the tested solver, in green the feasible explored design, and the color blue to outline unfeasible designs.
Due to the stochastic nature of our tests, we build our parallel plots using a median run on the following way: for each run of the optimizer, we store the best valid objective value. If none of the runs converges to a feasible point, we collect the minimal violation explored by the optimizer.
The median run is then selected based on the stored values for all runs.

\begin{figure}[p]
    \bigcentering
    \subfloat[SEGO-UTB ($\tau$: I-Log-1) \label{fig:pargraph:IL}]{\includegraphics[height=0.85\textheight]{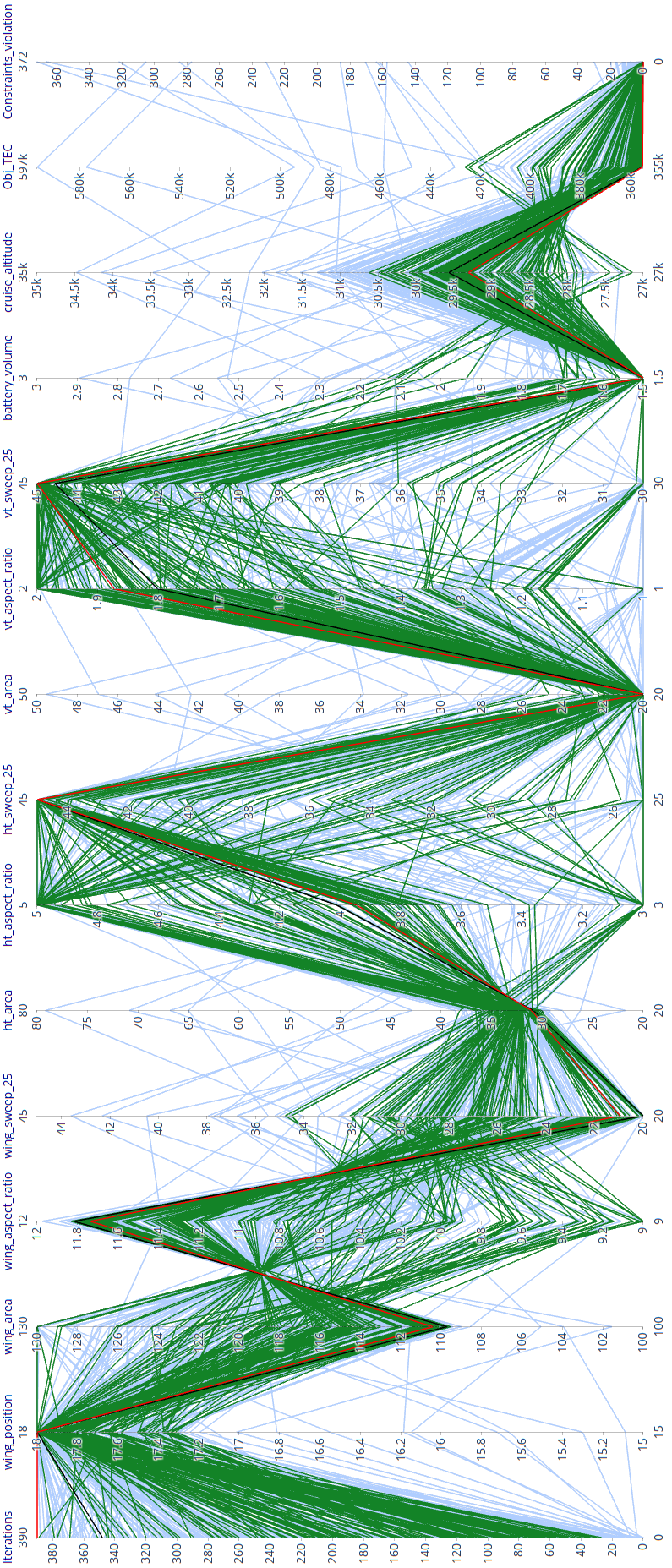}}
    \subfloat[COBYLA \label{fig:pargraph:Cobyla}]{\includegraphics[height=0.85\textheight]{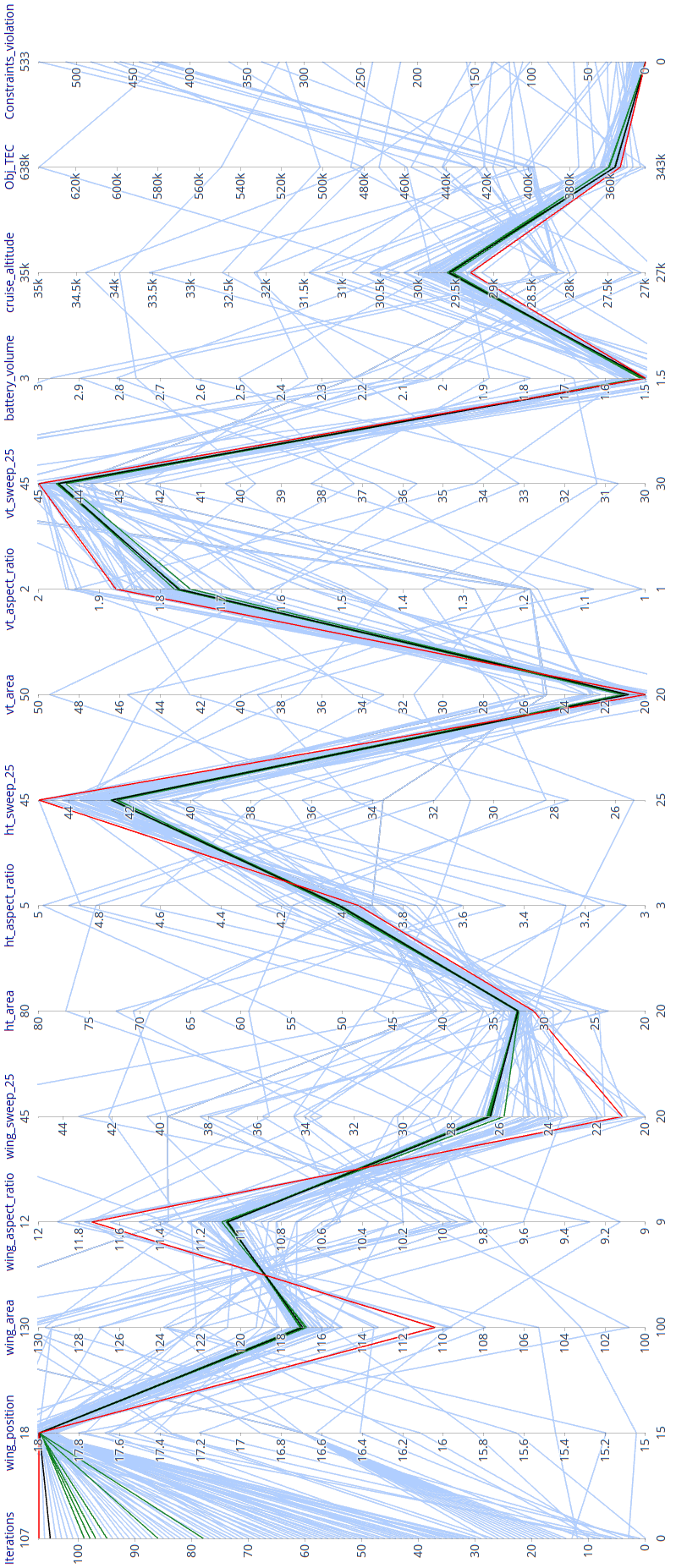}}
    \caption{Parallel plots using the median run for FAST problem. In grey: the designs outside of the design space; in blue: the unfeasible designs; in green: the feasible designs; in black: the optimum; in red: the reference design.}
    \label{fig:pargraph}
\end{figure}
Figure~\ref{fig:pargraph} represents the obtained parallel plots based on the median run of the two solvers SEGO-UTB ($\tau$: I-Log-1) and COBYLA.
The parallel plots for the other optimizers are displayed in \ref{app:pb}.
We note that in terms of the exploitation behaviour, represented by the convergence speed to a feasible design, SEGO-UTB ($\tau$: I-Log-1) is clearly outperforming COBYLA. In fact, unlike COBYLA, the majority of designs obtained by SEGO-UTB ($\tau$: I-Log-1) get feasible from the beginning of the optimization, see the iterations axis (on the bottom) of the parallel plots given by Figure~\ref{fig:pargraph:IL}.
The exploration behaviour is observed between iterations 250 and 390 with a majority of blue curves on the iterations axis.
Apart from the horizontal and vertical tails aspect ratio values (i.e. $7^{\text{th}}$ and $10^{\text{th}}$ axis), the SEGO-UTB ($\tau$: I-Log-1) converges to the reference design.
We notice also that COBYLA is not always evaluating designs that respect the bound constraints, as shown by the designs drawn in grey on Figure~\ref{fig:pargraph:Cobyla} between iterations 15 and 30.
Finally, the COBYLA best design has an optimal TEC value close to the reference one (around 5\%) even if some variables are different from the reference value as depicted in Figure~\ref{fig:pargraph:Cobyla}. However, these variables are of second importance according to the previous work of \citet{sguegliafasthybrid2018}. In fact, the main driver for hybrid aircraft design is the battery volume which matches in both cases.

\section{Conclusions}
\label{sec:clc}

The SEGO solver addresses the mixed constrained optimization problems in the Bayesian optimization scope.
However, it has difficulties to solve problems where the constraints are not well approximated by the GPs during the optimization process. 
In this paper, on the top of SEGO, we propose to use the upper trust bound while modeling the constraints to enhance the exploration of the design space. The proposed estimation combines the GP mean prediction and the associated uncertainty estimation.
The included upper trust bounds on the constraints were monitored using constraints learning rates. Three different evolutions for such rates were explored; constant, decreasing and non-decreasing.

\change{Using 29 constrained optimization problems, our proposed methods outperformed existing BO solvers (i.e., ALBO, EFI, SUR and PESC), COBYLA and NOMAD. In particular, the use of a non-decreasing logarithmic update strategy for the constraints learning rate, turns to be very useful for the SEGO framework. The good performances of the proposed solvers were confirmed on an aircraft design problem. That is why, we suggest to use the SEGO-UTB \changeb{($\tau$: I-Log-2)} \removeb{($\tau$: I-Log-1)} as default settings for a naive user even if the choice of the constraints learning is clearly problem dependent. Indeed, an exploratory constraints learning rate must be preferred to exploitative one on highly non linear problems.}

\section*{Acknowledgements}
This work is part of the activities of ONERA–ISAE–ENAC joint research group.

\section*{References}
\bibliography{biblio}
\bibliographystyle{elsarticle-num-names}

\newpage
\appendix
\renewcommand*{\thesection}{\appendixname~\Alph{section}}

\section{Academic Problems}
\label{app:pb}
\setcounter{equation}{0}
\setcounter{figure}{0}
\setcounter{table}{0}

In this appendix, we provide the academic problems.
A focus is first done on the representative problems and then an overview of the 29 problem benchmark information is given.

\subsection{The representative problems}
\label{app:repre_pb}

An outline of the presentation of the representative constrained problems is as follows.
As some of the problems share the same objective and constraints functions, we first introduce the objective functions.
Then, an overview of the constraints is given. Finally, The problems definitions are provided. 

\subsubsection{Objective functions}

The $f_1$ objective function is linear, $f_2$ is a centered and rescaled version of the \textit{Goldstein-Price} function \cite{PichenyBayesianoptimizationmixed2016} and $f_3$ is the modified Branin function \cite{ParrReviewefficientsurrogate2010}. 
\begin{equation*}
    \label{eq:obj}
    \begin{split}
        f_1(\bm{x}) = & ~ \sum\limits_{i=1}^{d}x_i \\ 
        f_2(\bm{x}) = & ~ \frac{\log \left[ \left(1 + a(4x_1 + 4x_2 - 3)^2\right) \left(30 + b(8x_1 - 12x_2 + 2)^2\right) \right] - 8.69}{2.43}\text{, with} \\
        a = & ~ 75 - 56(x_1 + x_2) + 3(4x_1-2)^2 + 6(4x_1 - 2)(4x_2 - 2) + 3(4x_2 - 2)^2 \\
        b = & ~ -14 -128x_1 +12(4x_1 -2)^2 +192x_2 - 36(4x_1 - 2)(4x_2 - 2) \\
            & ~ + 27(4x_2 - 2)^2 \\
        f_3(\bm{x}) = & ~ \left[ \left( x_2 - \frac{5.1x_1^2}{4\pi^2} + \frac{5x_1}{\pi} - 6 \right)^2 +\left( 10 - \frac{10}{8\pi} \right) \cos{(x_1) + 1} \right] + \frac{5x_1 + 25}{15} \\
    \end{split}
\end{equation*}

\subsubsection{Constraints functions}

The $c_1$ and $c_2$ constraints are the \textit{toy} problem ones, $c_3$ is the centered and rescaled Branin function, $c_4$ is taken from \citet{ParrInfillsamplingcriteria2012}, $c_5$ is the centered \textit{Ackley} function, $c_6$ is the \textit{Hartman} function centered and rescaled and $c_7$ is the constrained function of the modified Branin problem.
All these functions can be found in \citet{PichenyBayesianoptimizationmixed2016,ParrReviewefficientsurrogate2010}.
\begin{equation*}
    \label{eq:cons}
    \begin{split}
        c_1(\bm{x}) =   & ~ 0.5 \sin{\left(2\pi\left( x_1^2 -2x_2 \right)\right)} +x_1 +2x_2 - 1.5 \\ 
        c_2(\bm{x}) =   & ~ -x_1^2 -x_2^2 + 1.5 \\
        c_3(\bm{x}) =   & ~ 15 - \left(15x_2 - \frac{5}{4\pi^2}\left(15x_1-5\right)^2 + \frac{5}{\pi}\left(15x_1-5\right) - 6\right)^2 \\
                        & ~ - 10\left(1-\frac{1}{8\pi}\right)\cos{(15x_1-5)} \\
        c_4(\bm{x}) =   & ~ 4 - \left(4 - 2.1\left(2x_1-1\right)^2 + \frac{\left(2x_1-1\right)^4}{3}\right)\left(2x_1-1\right)^2 -(2x_1-1)(2x_2-1) \\
                        & ~ -16\left(x_2^2-x_2\right)(2x_2-1)^2 -3\sin{(12(1-x_1))}-3\sin{(12(1-x_2))} \\
        c_5(\bm{x}) =   & ~ 20 \exp{\left(-0.2\sqrt{\frac{1}{4}\sum\limits_{i=1}^{4}{\left(3x_i-1\right)^2}}\right)} + \exp{\left(\frac{1}{4}\sum\limits_{i=1}^{4}{\cos{(2\pi(3x_i-1))}}\right)} \\
                        & ~ - 17 - \exp{(1)} \\
        c_6(\bm{x}) =   & ~ \frac{1}{0.8387} \left[-1.1+\sum\limits_{i=1}^{4}{C_i \exp{\left(-\sum\limits_{j=1}^{4}{a_{ij}(x_j-p_{ij})^2}\right)}} \right] \\
        c_7(\bm{x}) =   & ~ 6 - \left(4-2.1\overline{x}_1^2 + \frac{\overline{x}_1^4}{3}\right)\overline{x}_1^2 - \overline{x}_1\overline{x}_2 - \left(4\overline{x}_2^2-4\right)\overline{x}_2^2 - 3\sin{\left(6(1-\overline{x}_1)\right)} \\
                        & ~ - 3\sin{\left(6(1-\overline{x}_2)\right)}
    \end{split}
\end{equation*}
with:
\begin{gather*}
    \bm{a} = 
        \begin{bmatrix}
            10.00 & 0.05 & 3.00 & 17.00 \\
            3.00 & 10.00 & 3.50 & 8.00 \\
            17.00 & 17.00 & 1.70 & 0.05 \\
            3.50 & 0.10 & 10.00 & 10.00 \\
        \end{bmatrix}, ~ 
    \bm{p} = 
        \begin{bmatrix}
            0.131 & 0.232 & 0.234 & 0.404 \\
            0.169 & 0.413 & 0.145 & 0.882 \\
            0.556 & 0.830 & 0.352 & 0.873 \\
            0.012 & 0.373 & 0.288 & 0.574 \\
        \end{bmatrix}, \\
    \bm{C} = 
        \begin{bmatrix}
            1.0 & 1.2 & 3.0 & 3.2 
        \end{bmatrix}, ~ \overline{x}_1 = \frac{x_1-2.5}{7.5}, ~ \overline{x}_2 = \frac{x_2 - 7.5}{7.5}
\end{gather*}

\subsubsection{Problems}

The four representative problems and the \textit{modified Branin problem with equality constraints} (MBE) are expressed as follows:
\begin{align}
    \text{(MB): } & \min_{\bm{x} \in \Omega_3}{f_3(\bm{x}) ~~ \text{s.t.} ~~ c_7(\bm{x}) \leq 0} \\
    \text{(LSQ): } & \min_{\bm{x} \in \Omega_1}{f_1(\bm{x}) ~~ \text{s.t.} ~~ c_1(\bm{x}) \geq 0 , ~ c_2(\bm{x})} \geq 0 \\
    \text{(GBSP): } & \min_{\bm{x} \in \Omega_1}{f_2(\bm{x}) ~~ \text{s.t.} ~~ c_1(\bm{x}) \geq 0, ~ c_2(\bm{x})} = 0, ~ c_3(\bm{x})=0 \\ 
    \text{(LAH): } & \min_{\bm{x} \in \Omega_2}{f_1(\bm{x}) ~~ \text{s.t.} ~~ c_5(\bm{x}) \leq 0 , ~ c_6(\bm{x})} = 0  \\
    \text{(MBE): } &  \min_{\bm{x} \in \Omega_3}{f_3(\bm{x}) ~~ \text{s.t.} ~~ c_7(\bm{x}) = 0}
\end{align}
with:
$\Omega_1 = [0,1]^2$, $\Omega_2 = [0,1]^4$ and $\Omega_3 = [-5,10] \times [0,15]$.

\begin{figure}[htb!]
    \centering
    \subfloat[MB \label{fig:pb:MB}]{\includegraphics[width=0.45\textwidth]{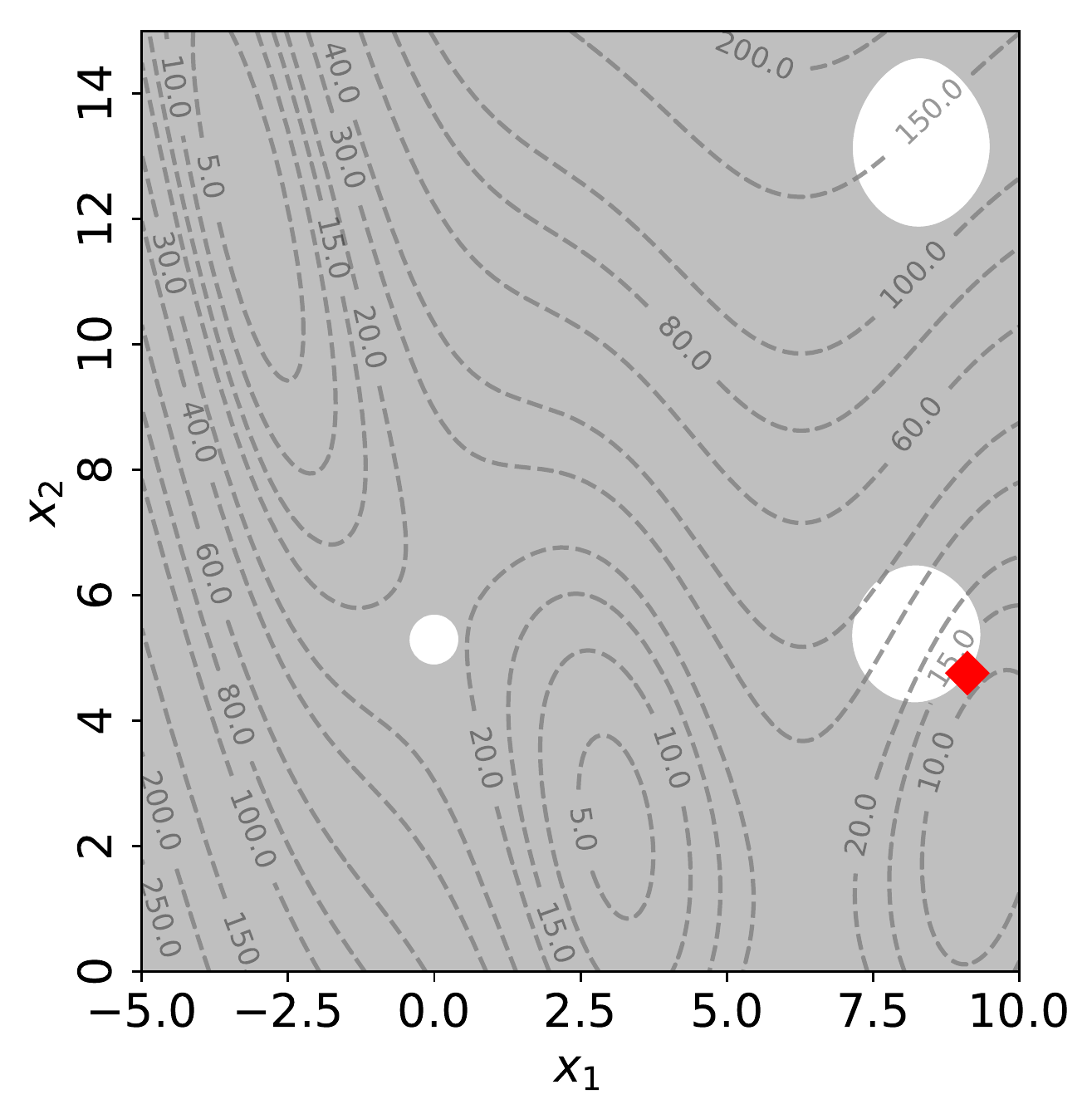}}
    \subfloat[LSQ \label{fig:pb:LSQ}]{\includegraphics[width=0.45\textwidth]{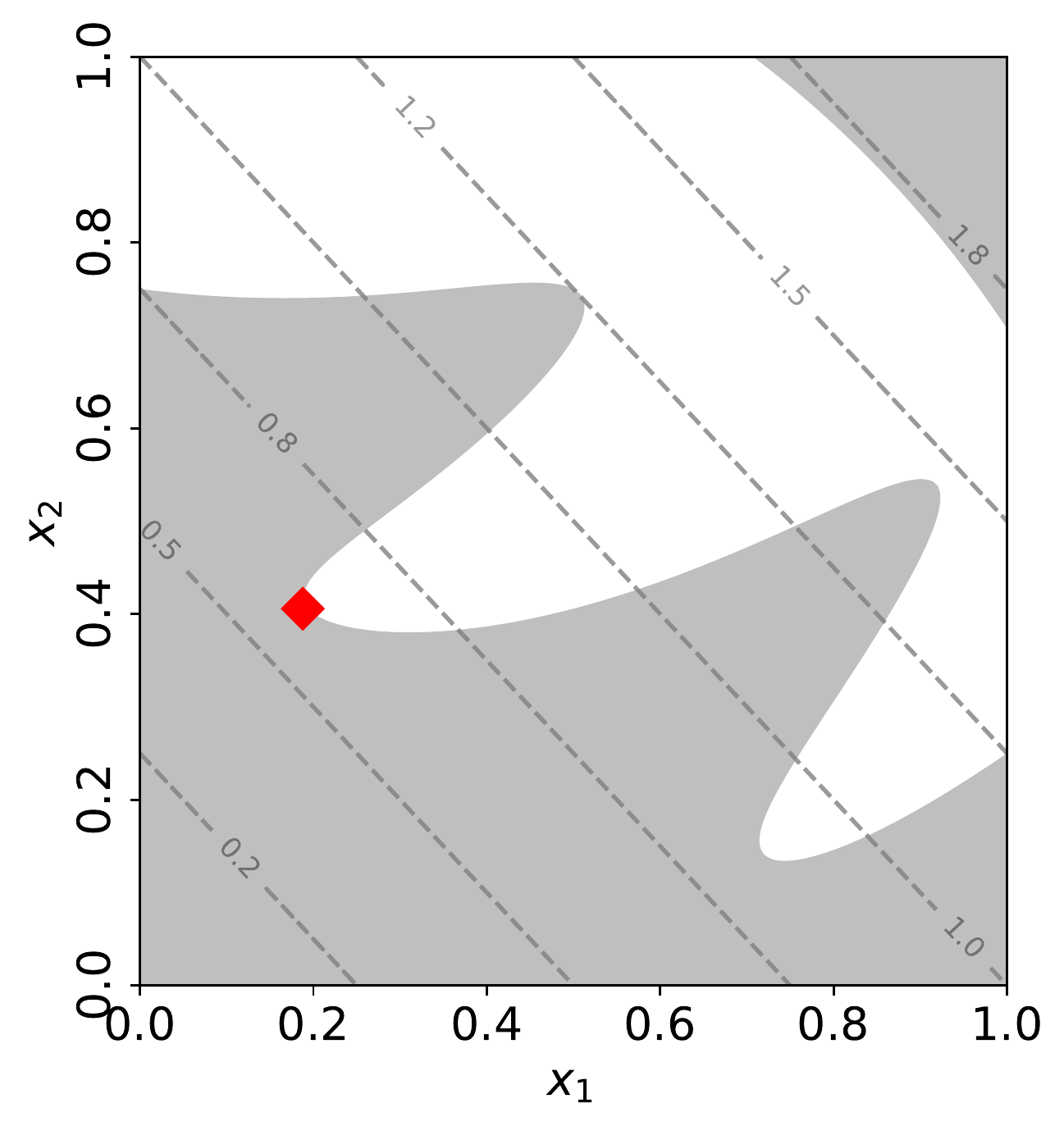}} \\
    \subfloat[MBE \label{fig:pb:MBE}]{\includegraphics[width=0.45\textwidth]{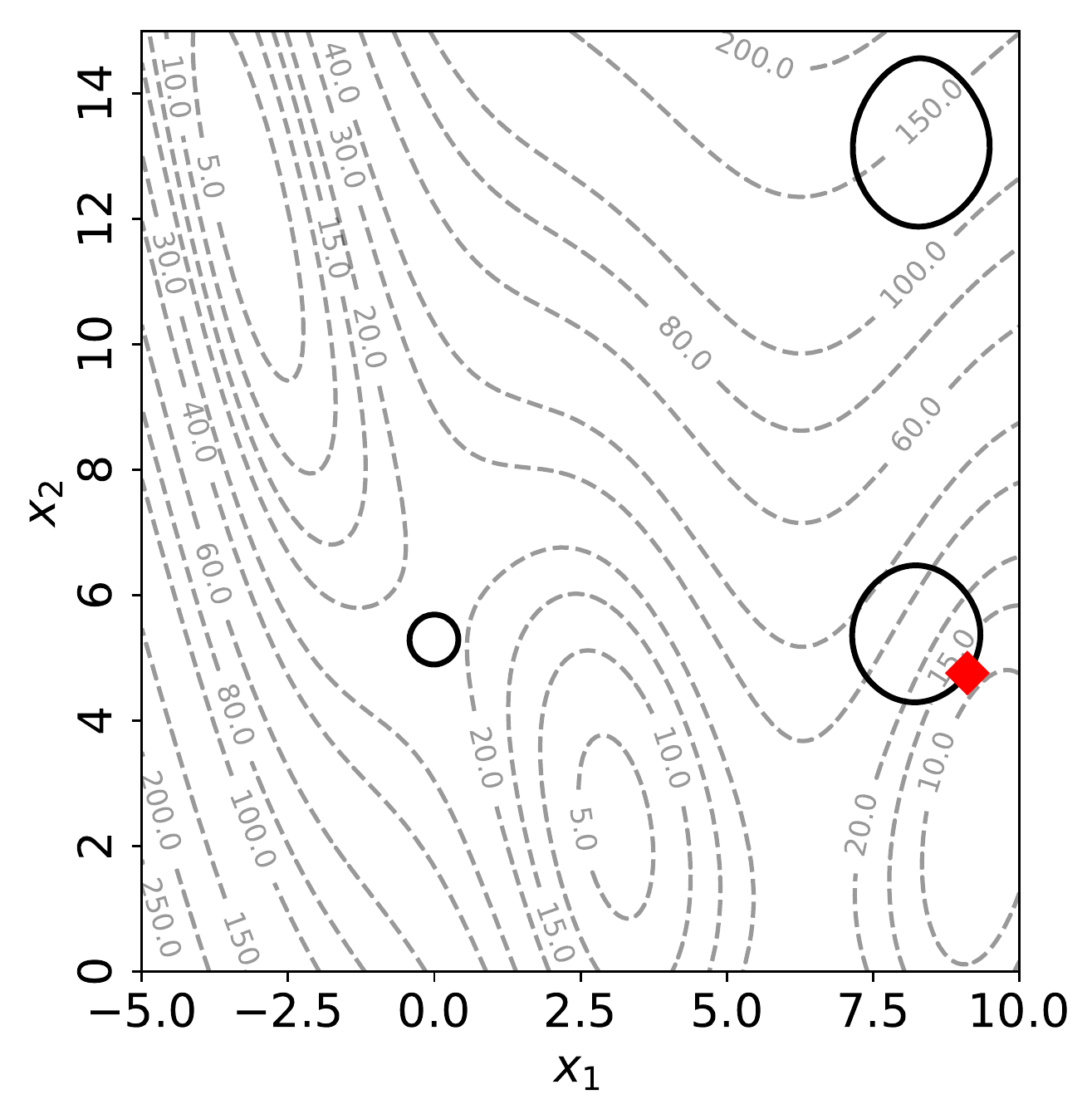}}
    \subfloat[GBSP \label{fig:pb:GBSP}]{\includegraphics[width=0.45\textwidth]{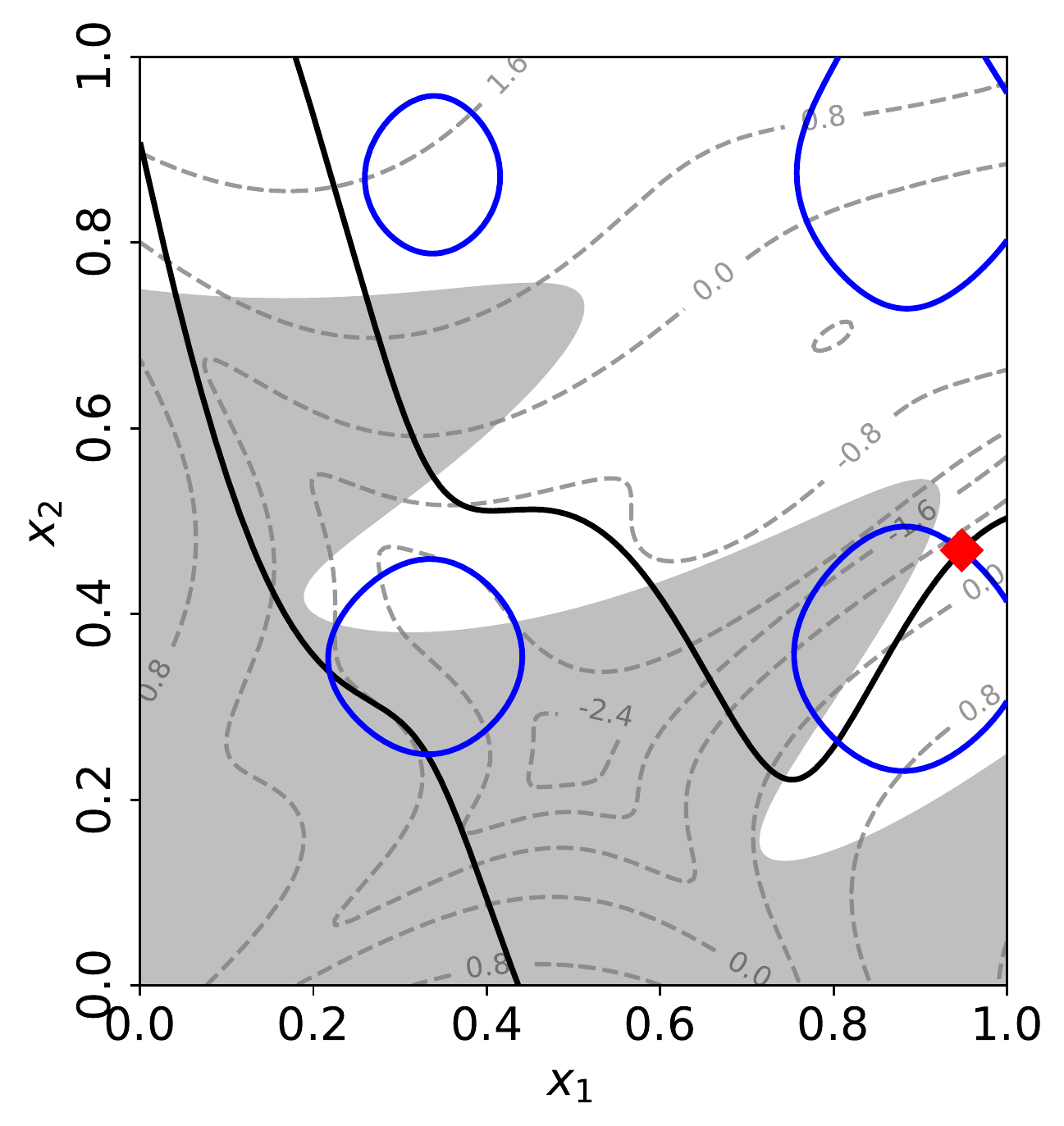}}
    \caption{Representation of the problems. Dashed curves: contour plot of the objective function ; grey area: unfeasible domain for inequality constraints ; filled curves: feasible domain for equality constraints ; red square: global optimum.}
    \label{fig:pb}
\end{figure}

The two dimensional problems are drawn in Figure~\ref{fig:pb} to a better understanding of the addressed challenges.
Figure~\ref{fig:pb:MB} shows that the MB problem has disjoint feasible domains and a multimodal objective function.
The LSQ problem is interesting because of the non convexity of the feasible domain (see Figure~\ref{fig:pb:LSQ}).
The MBE equality constrained problem has a multimodal objective function and three disjoint feasible domains have been drawn in Figure~\ref{fig:pb:MBE}.
Then, Figure~\ref{fig:pb:GBSP} displays the GBSP mixed constrained problem.
It is the most challenging one as it gathers a multimodal objective function, a non convex feasible domain concerning the inequality constraint and disjoint feasible domains for equality ones.
The resulting feasible domain is thus restricted to only two points.
Finally, the LAH problem cannot be displayed as it is a four dimensional problem.

\subsection{The 29 problem benchmark}
\label{app:over}

The 29 problem benchmark is composed of well known optimization problems from \citet{Mezura-MontesEmpiricalanalysismodified2012,ParrReviewefficientsurrogate2010,PichenyBayesianoptimizationmixed2016,RegisEvolutionaryprogramminghighdimensional2014}.
We picked the problems between $2$ to $10$ design variables with equality, inequality and mixed constraints.
These problems are also segregated into two categories:
\begin{itemize}
    \item WNLC, meaning that all their constraints are linear or quadratic.
    \item HNLC, meaning at least one of the constraints is not linear or quadratic.
\end{itemize}
All these information are detailed for each problem in Table~\ref{tab:pb_inf}.
\clearpage
\thispagestyle{empty}

\begin{sidewaystable}[htb!]
\bigcentering
\caption{\changeb{Some of the features of the tested constrained optimization problem.}}
\label{tab:pb_inf}
\begin{tabular}{c|c|c|c|c|c||c|c|c|c|c|c}
\toprule
\textbf{Problem} & \textbf{Nb. of} & \textbf{Nb. of} & \textbf{Nb. of} & \textbf{Linear} & \textbf{Ref.} & \textbf{Problem} & \textbf{Nb. of} & \textbf{Nb. of} & \textbf{Nb. of} & \textbf{Linear} & \textbf{Ref.}\\
\textbf{Name} & \textbf{Variables} & \textbf{Eq. Cst.} & \textbf{Ieq. Cst.} & \textbf{Type} & \textbf{Value} & \textbf{Name} & \textbf{Variables} & \textbf{Eq. Cst.} & \textbf{Ieq. Cst.} & \textbf{Type} & \textbf{Value} \\ \midrule
G03 & 10 & 1 & 0 & WNLC & $-1.000$ & G04 & 5 & 0 & 6 & WNLC & $-3.067 \cdot 10^{4}$ \\
G05 & 4 & 3 & 2 & HNLC & $5126$ & G06 & 2 & 0 & 2 & WNLC & $-6962$ \\
G07 & 10 & 0 & 8 & WNLC & $24.23$ & G08 & 2 & 0 & 2 & WNLC & $-9.583 \cdot 10^{-2}$ \\
G09 & 7 & 0 & 4 & WNLC & $680.6$ & G10 & 8 & 0 & 6 & WNLC & $7049$ \\
G11 & 2 & 1 & 0 & WNLC & $0.750$ & G12 & 3 & 0 & 1 & WNLC & $-1.000$ \\
G13 & 5 & 3 & 0 & HNLC & $2.201 \cdot 10^{-3}$ & G14 & 10 & 3 & 0 & WNLC & $-47.71$ \\
G15 & 3 & 2 & 0 & WNLC & $961.7$ & G16 & 5 & 0 & 38 & HNLC & $-1.918$ \\
G17 & 6 & 4 & 0 & WNLC & $8864$ & G18 & 9 & 0 & 13 & WNLC & $-0.8661$\\
G21 & 7 & 5 & 1 & WNLC & $193.8$ & G23 & 9 & 4 & 2 & WNLC & $-400.1$\\
G24 & 2 & 0 & 2 & HNLC & $-6.031$ & WB4 & 4 & 0 & 6 & HNLC & $0.4734$ \\
GBSP & 2 & 2 & 1 & HNLC & $-0.5252$ & GTCD & 4 & 0 & 1 & HNLC & $2.965 \cdot 10^{6}$ \\
Hesse & 6 & 0 & 6 & WNLC & $-310.0$ & LAH & 4 & 1 & 1 & HNLC & $5.176 \cdot 10^{-2}$\\
LSQ & 2 & 0 & 2 & HNLC & $0.600$ & SR7 & 7 & 0 & 11 & HNLC & $2994$ \\
MB & 2 & 0 & 1 & HNLC & $12.00$ & MBE & 2 & 1 & 0 & HNLC & $12.00$\\
PVD4 & 4 & 0 & 3 & HNLC & $5809$ & & & & \\
\bottomrule
\end{tabular}
\end{sidewaystable}

\clearpage
\section{Additional results}
\label{app:res}
\setcounter{equation}{0}
\setcounter{figure}{0}
\setcounter{table}{0}

In this appendix, we comment the results on the four test problems from \citet{PichenyBayesianoptimizationmixed2016,ParrReviewefficientsurrogate2010}, the 29 problem benchmark and the FAST problem for the non-decreasing and decreasing constraints learning rate strategies.
For more information on the tests plan and methodology, see Sections~\ref{sec:test} and \ref{sec:FAST}.

\subsection{The representative problems}

\begin{figure}[p]
    \centering
    
    \subfloat[Decreasing constraints learning rate strategies using $\epsilon_c=10^{-2}$. \label{fig:gbsp_ad:dec2}]{\includegraphics[width=0.45\textwidth]{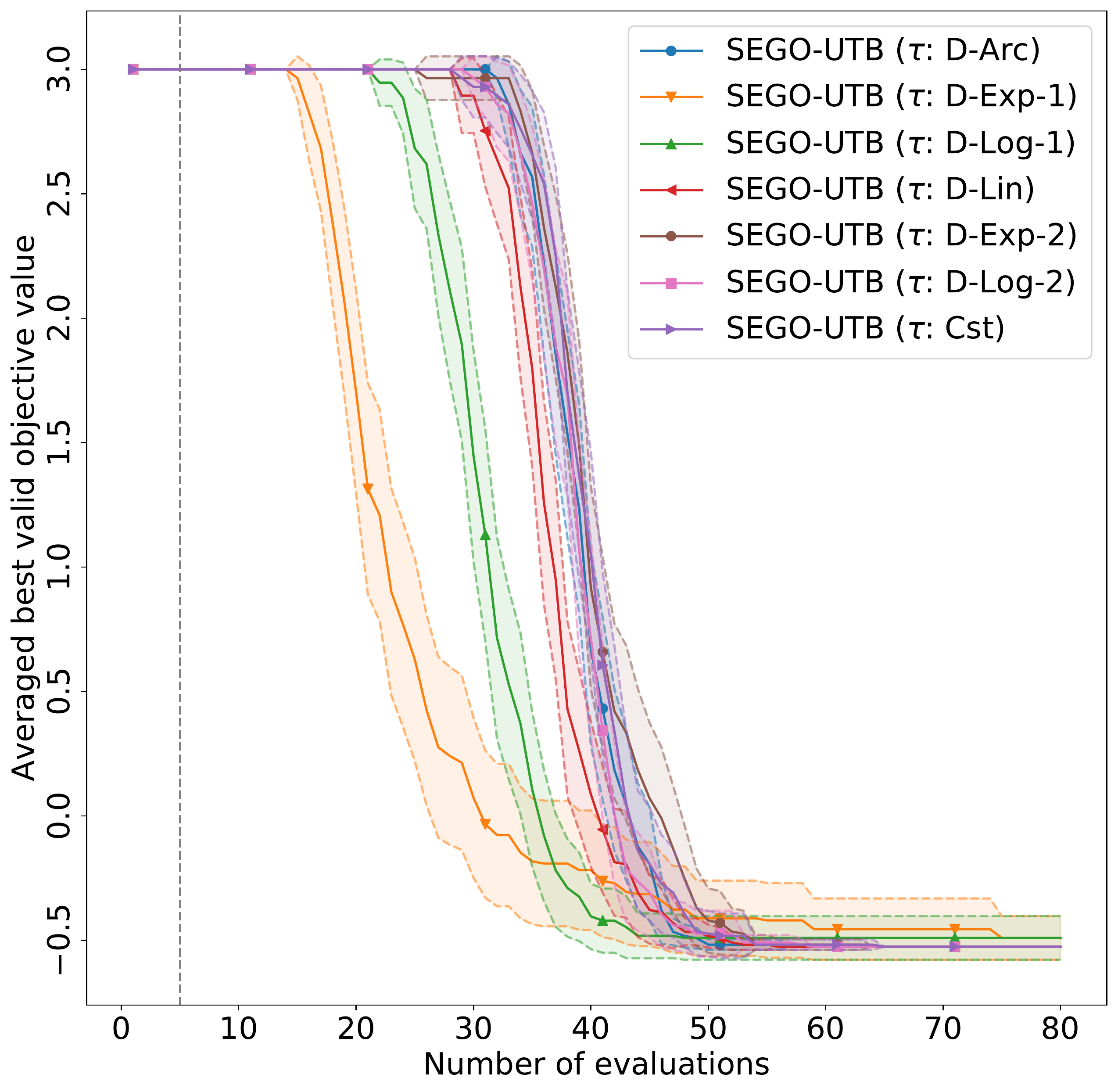}}
    \subfloat[Non-decreasing constraints learning rate strategies using $\epsilon_c=10^{-2}$. \label{fig:gbsp_ad:inc2}]{\includegraphics[width=0.45\textwidth]{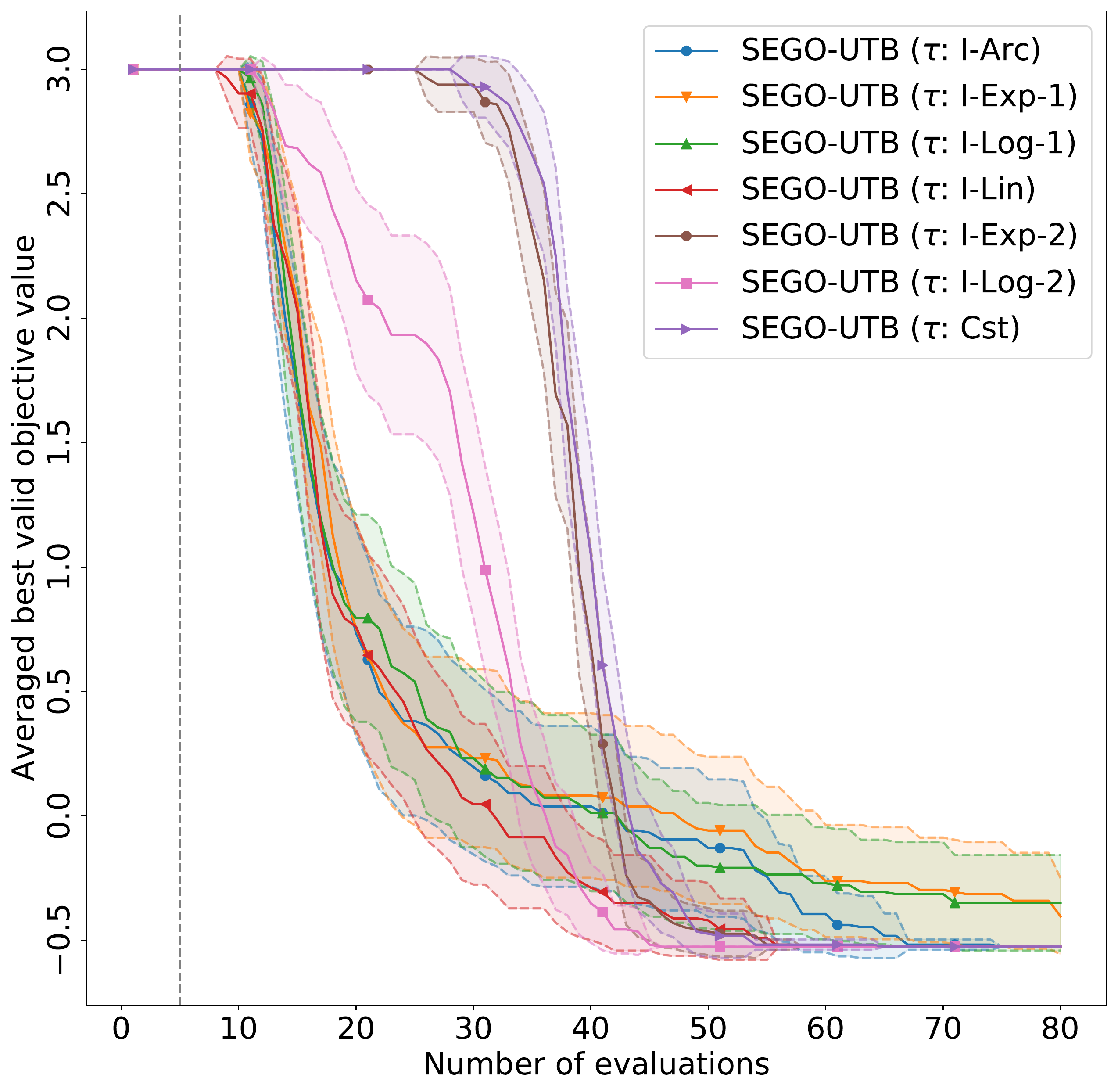}} \\
    
    \subfloat[Decreasing constraints learning rate strategies using $\epsilon_c=10^{-4}$. \label{fig:gbsp_ad:dec4}]{\includegraphics[width=0.45\textwidth]{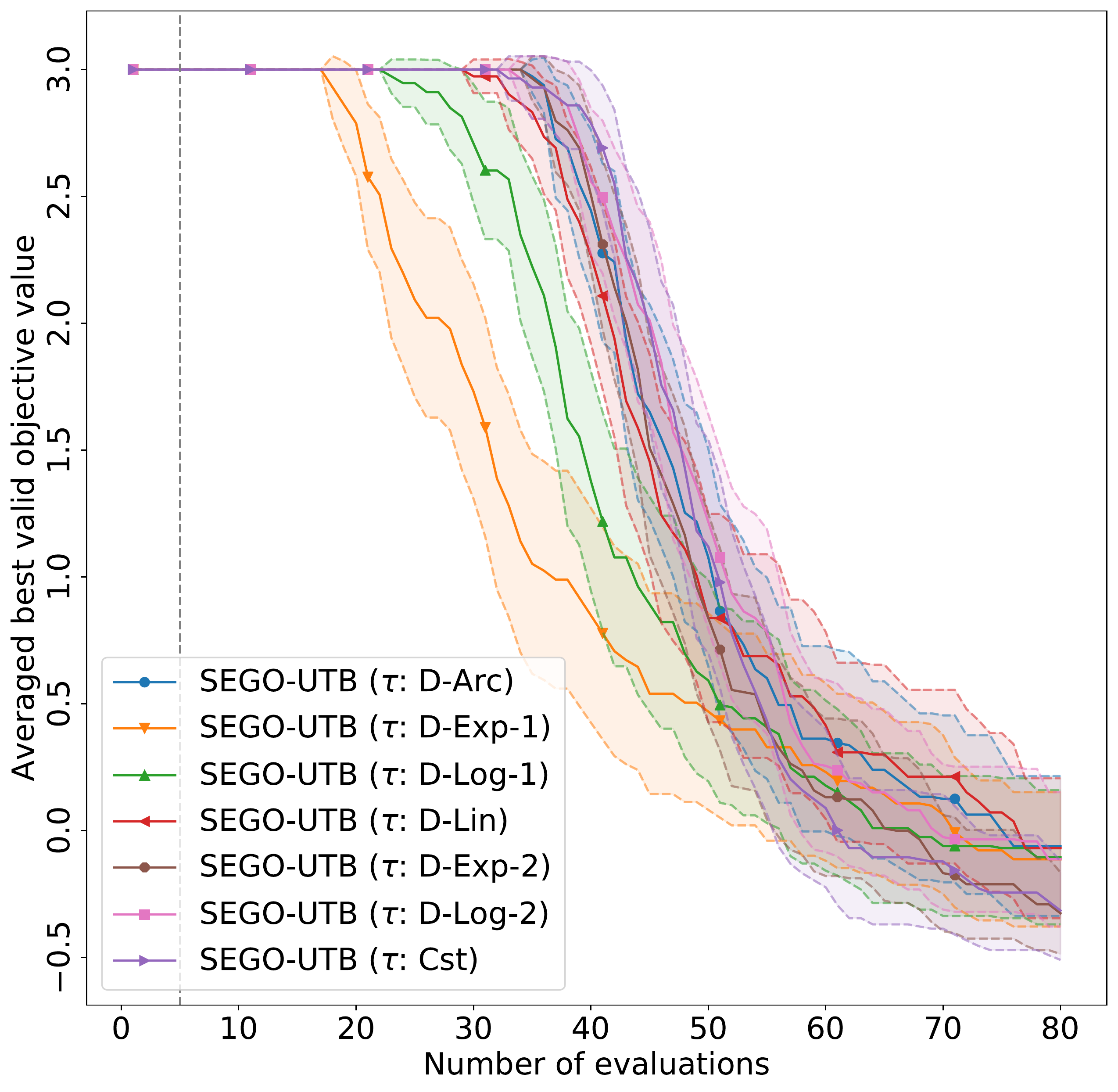}}
    \subfloat[Non-decreasing constraints learning rate strategies using $\epsilon_c=10^{-4}$. \label{fig:gbsp_ad:inc4}]{\includegraphics[width=0.45\textwidth]{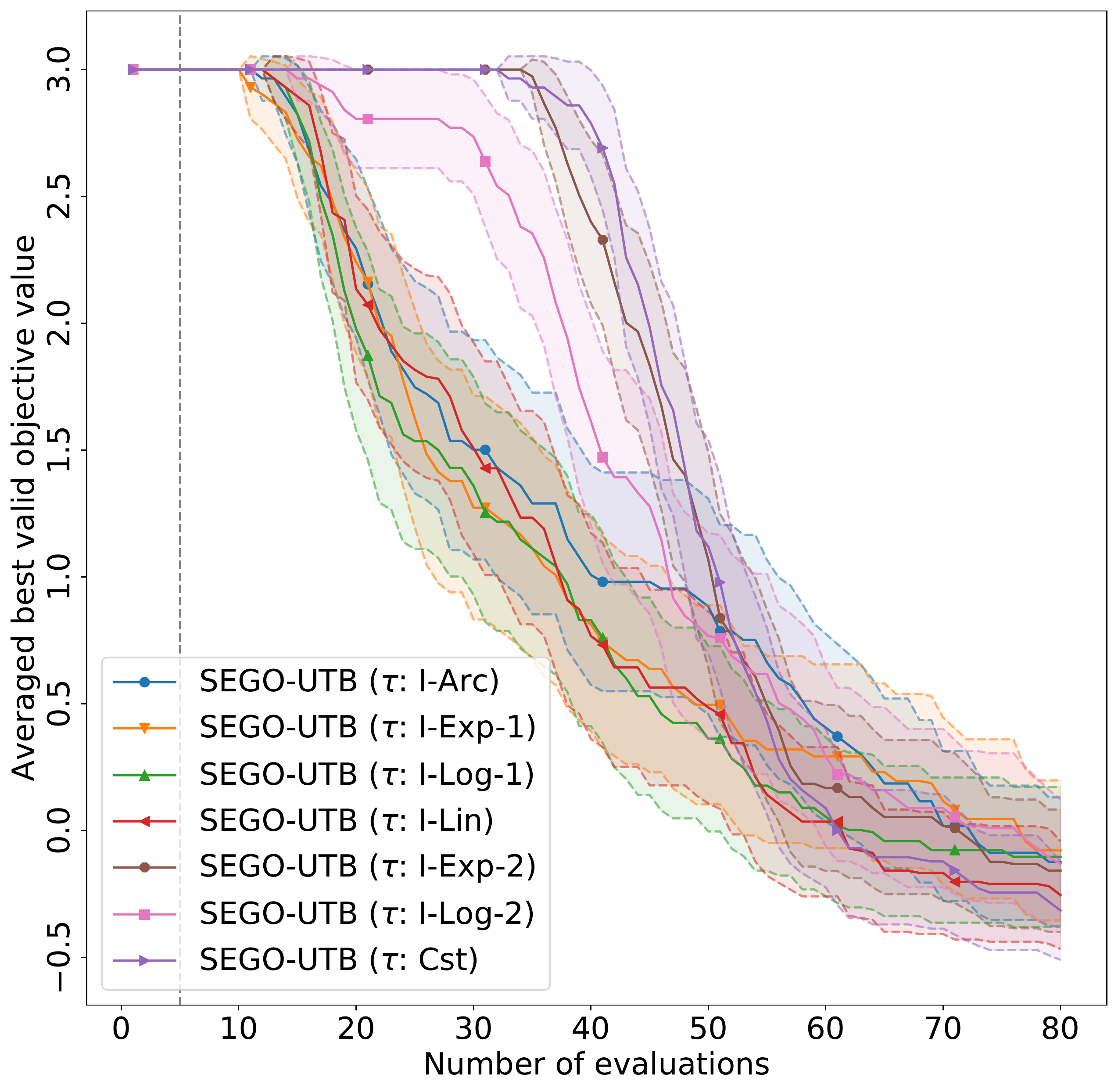}}
    
    \caption{Convergence plots for the GBSP problem, considering the two levels of constraints violation $10^{-4}$ and $10^{-2}$. The vertical grey-dashed line outlines the number of points in the initial DoEs.}
    \label{fig:gbsp_ad}
\end{figure}

Figure~\ref{fig:gbsp_ad} shows the averaged best valid value for non-decreasing number of evaluations of the GBSP problem \change{using the constraints violation of $10^{-2}$ and $10^{-4}$}. \change{Using both constraints violation, it appears that SEGO-UTB ($\tau$: D-Exp-2) (resp. ($\tau$: I-Exp-2)) shows the best compromise for the decreasing (resp. non-decreasing) constraints learning rate strategy.}

\begin{figure}[p]
    \centering
    
    \subfloat[Decreasing constraints learning rate strategies using $\epsilon_c=10^{-2}$. \label{fig:lah_ad:dec2}]{\includegraphics[width=0.45\textwidth]{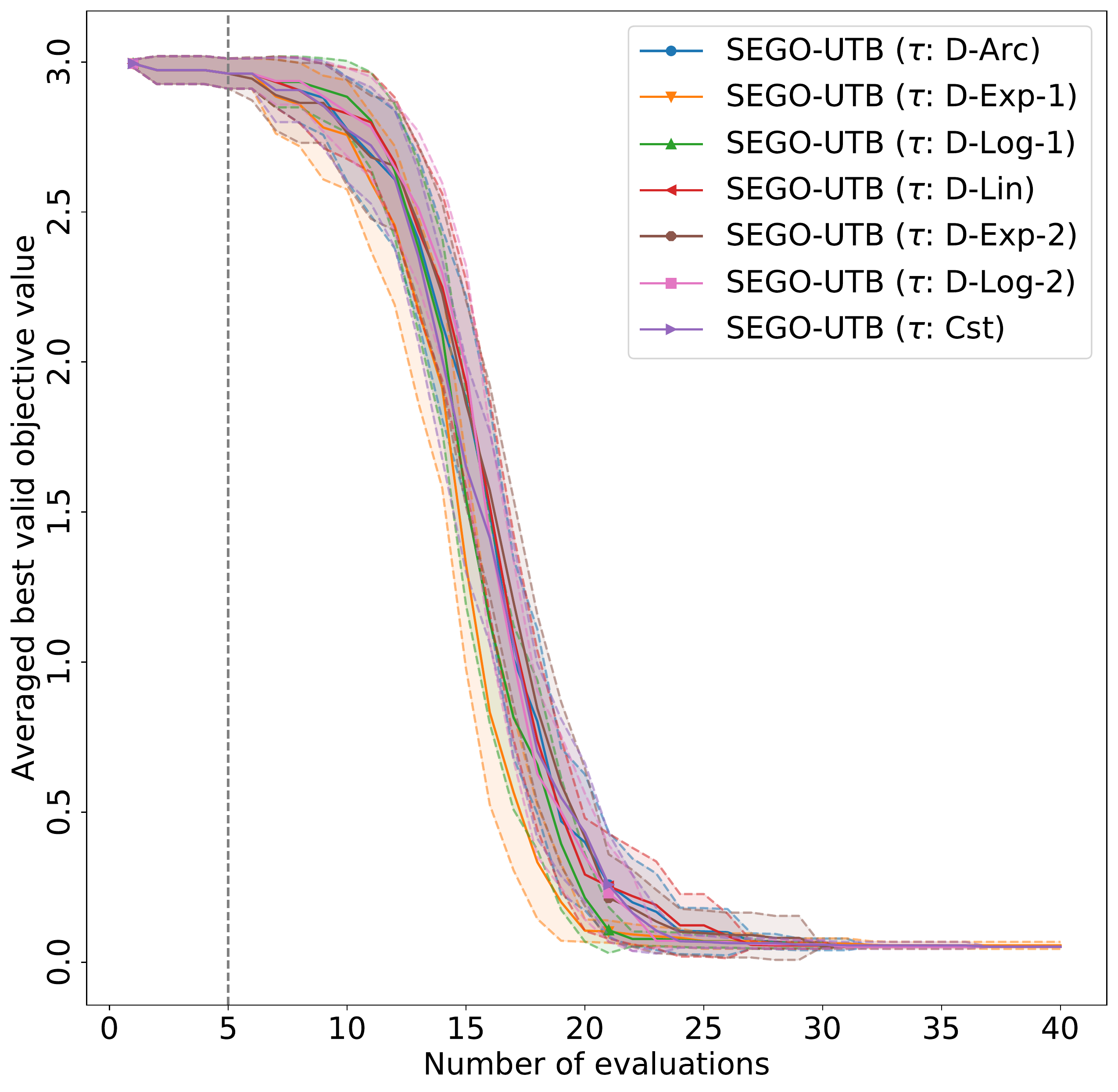}}
    \subfloat[Non-decreasing constraints learning rate strategies using $\epsilon_c=10^{-2}$. \label{fig:lah_ad:inc2}]{\includegraphics[width=0.45\textwidth]{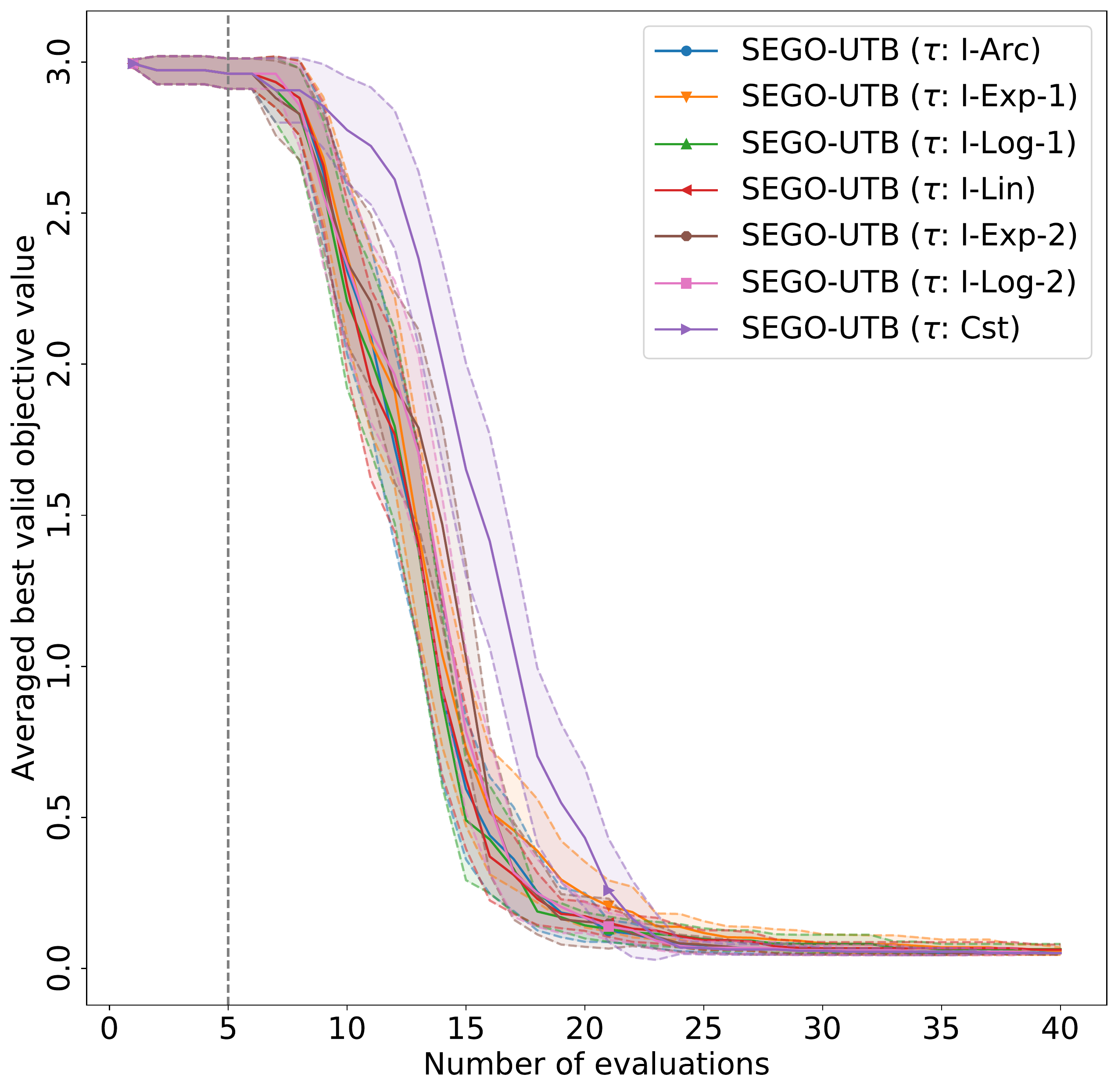}} \\
    
    \subfloat[Decreasing constraints learning rate strategies using $\epsilon_c=10^{-4}$. \label{fig:lah_ad:dec4}]{\includegraphics[width=0.45\textwidth]{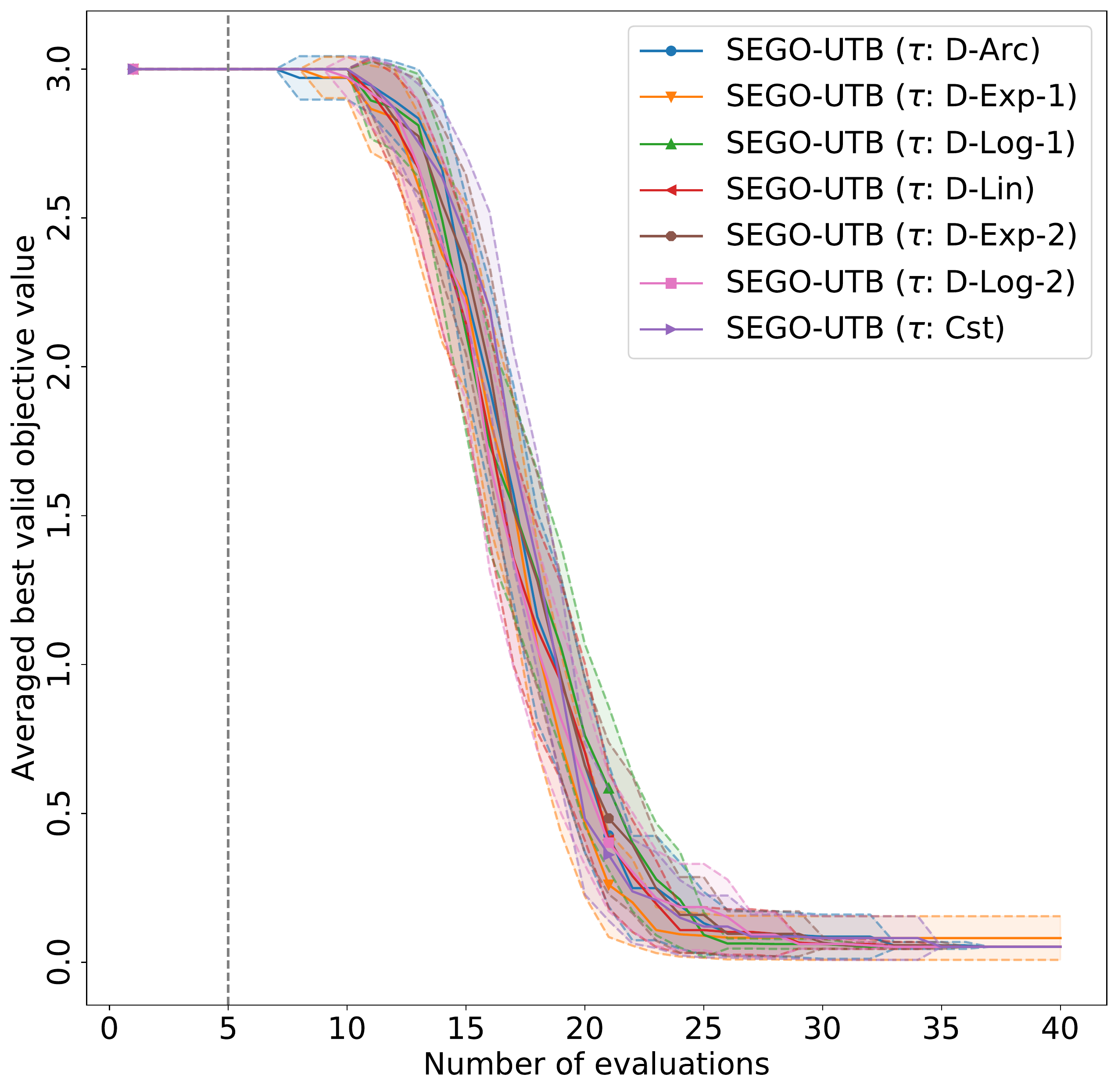}}
    \subfloat[Non-decreasing constraints learning rate strategies using $\epsilon_c=10^{-4}$. \label{fig:lah_ad:inc4}]{\includegraphics[width=0.45\textwidth]{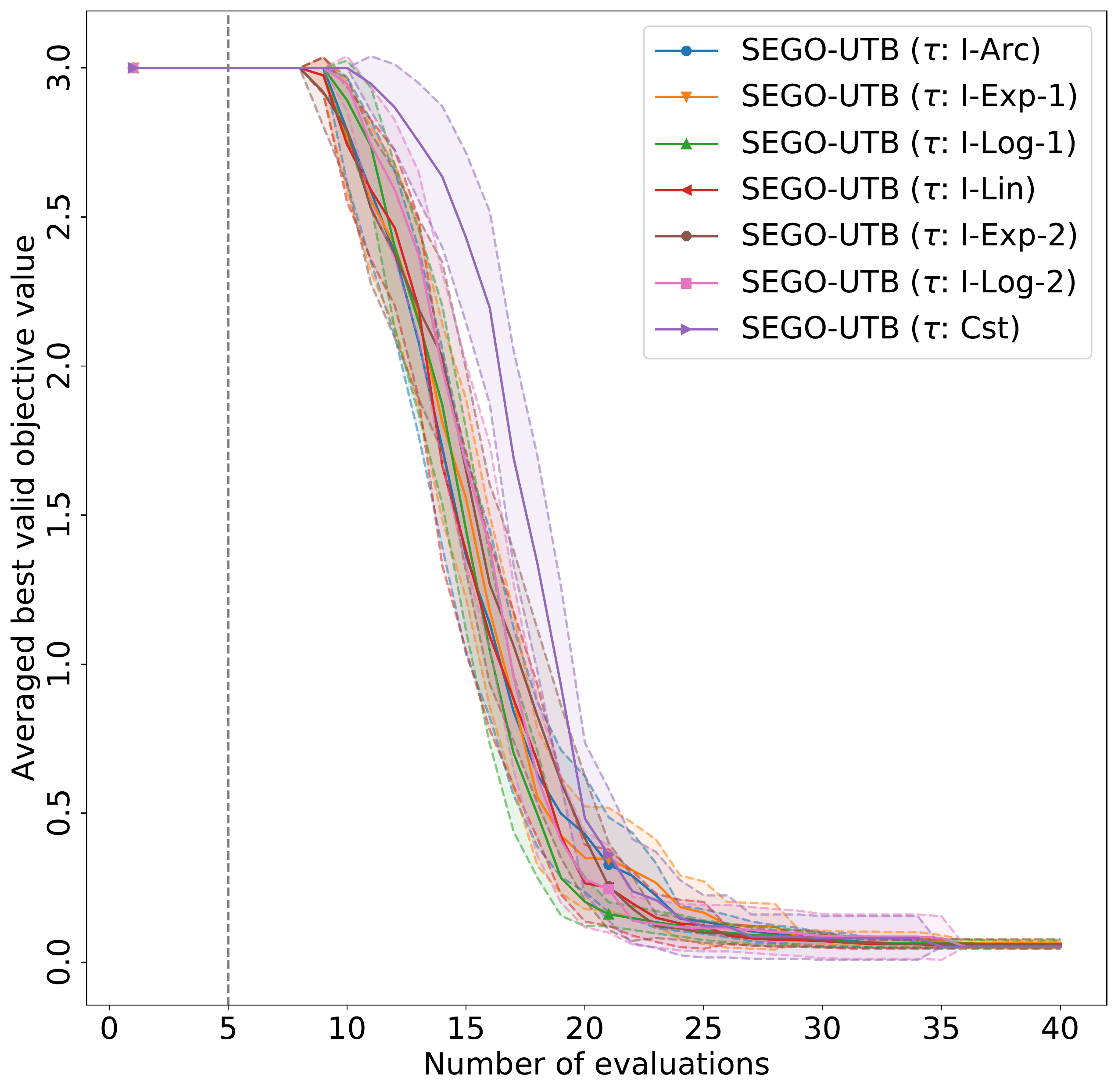}}
    
    \caption{Convergence plots for the LAH problem, considering the two levels of constraints violation $10^{-4}$ and $10^{-2}$. The vertical grey-dashed line outlines the number of points in the initial DoEs.}
    \label{fig:lah_ad}
\end{figure}

For the LAH problem, all the non-decreasing and decreasing learning rates are performing almost the same and are converging in less than 40 iterations as shown by Figure~\ref{fig:lah_ad}. \change{We choose arbitrarily SEGO-UTB ($\tau$: D-Exp-2) and resp. SEGO-UTB ($\tau$: I-Exp-2) as references constraints learning rate strategies for the LAH problem.}

\begin{figure}[p]
    \centering
    
    \subfloat[Decreasing constraints learning rate strategies using $\epsilon_c=10^{-2}$. \label{fig:lsq_ad:dec2}]{\includegraphics[width=0.45\textwidth]{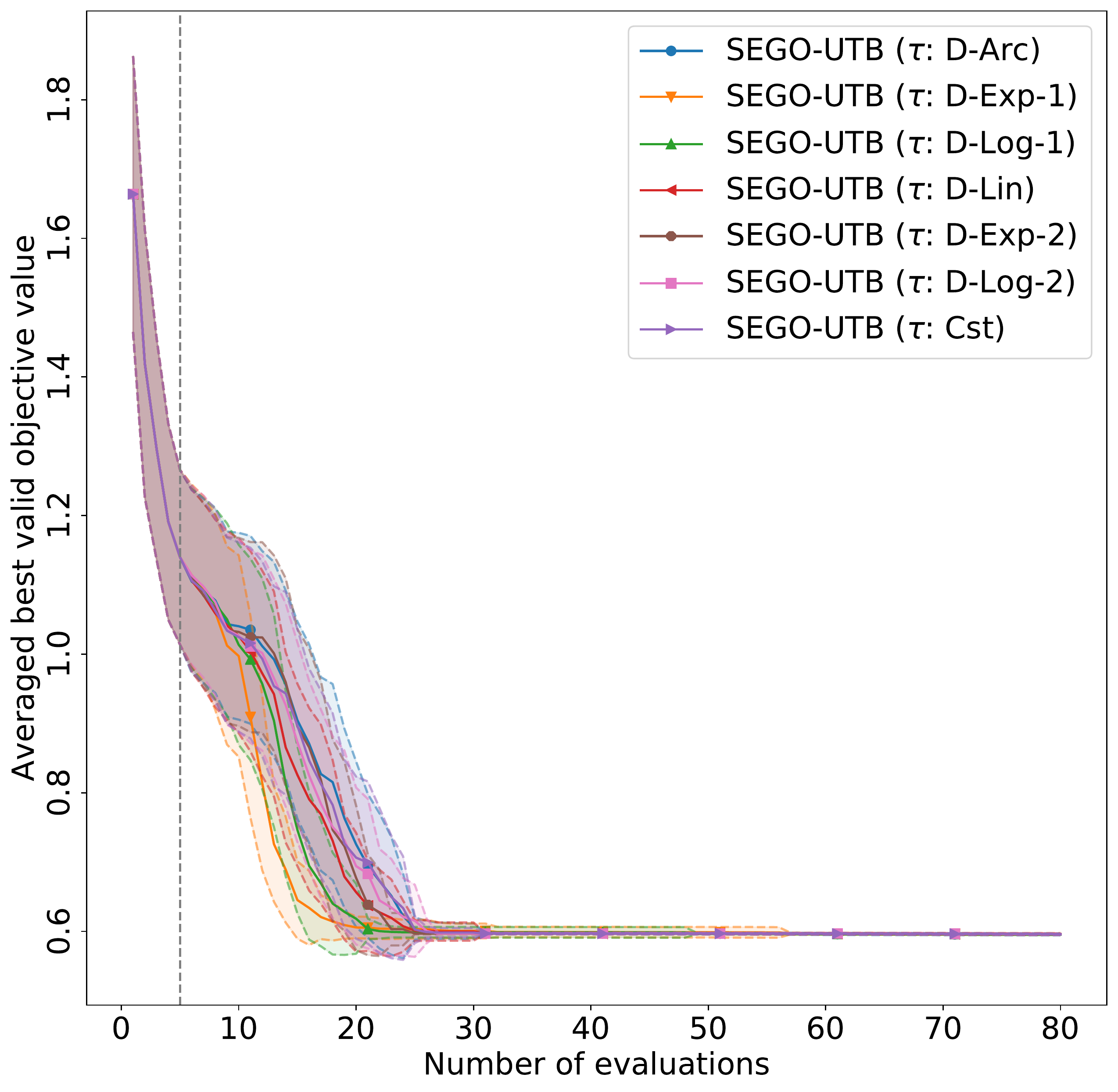}}
    \subfloat[Non-decreasing constraints learning rate strategies using $\epsilon_c=10^{-2}$. \label{fig:lsq_ad:inc2}]{\includegraphics[width=0.45\textwidth]{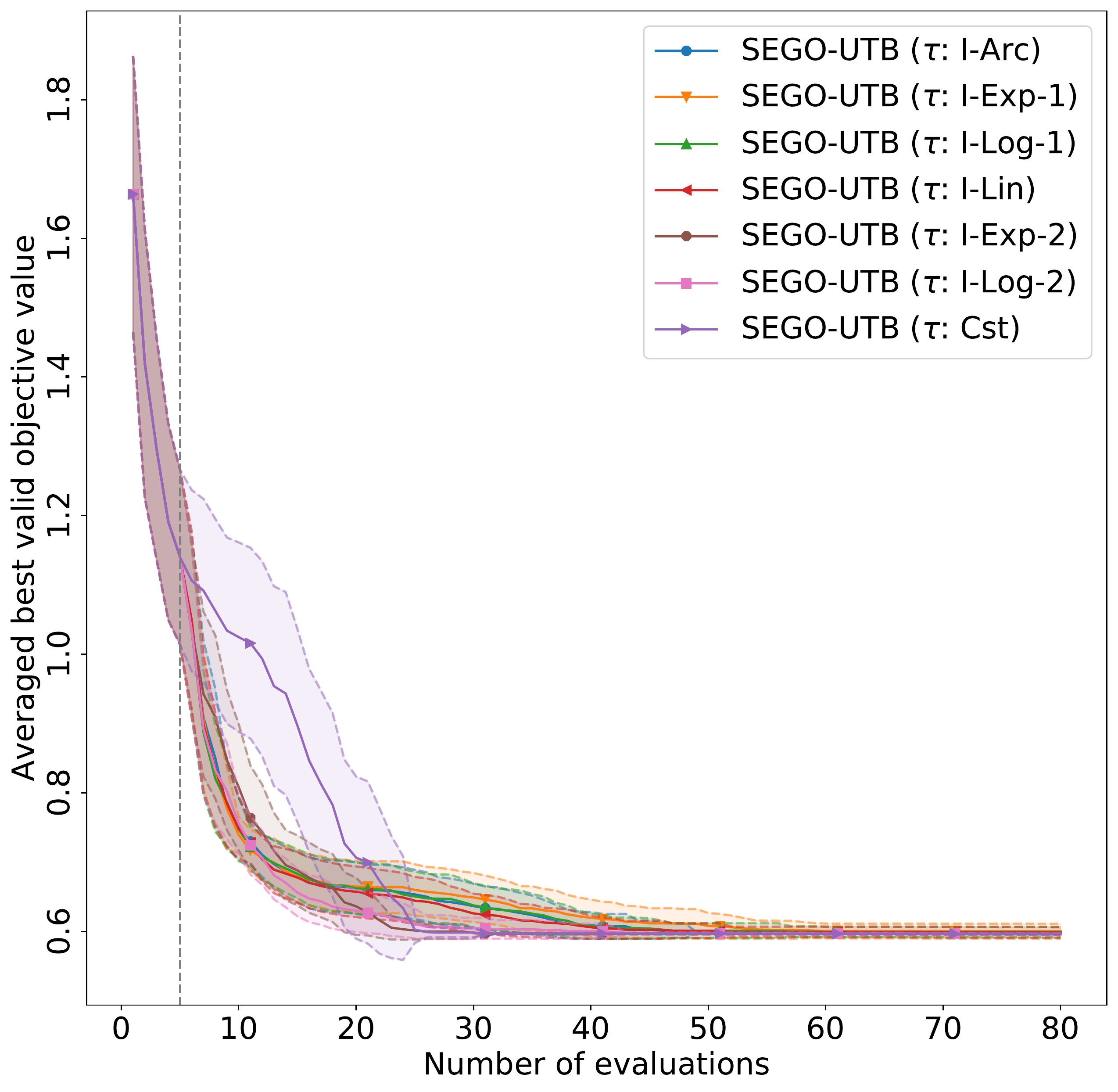}} \\
    
    \subfloat[Decreasing constraints learning rate strategies using $\epsilon_c=10^{-4}$. \label{fig:lsq_ad:dec4}]{\includegraphics[width=0.45\textwidth]{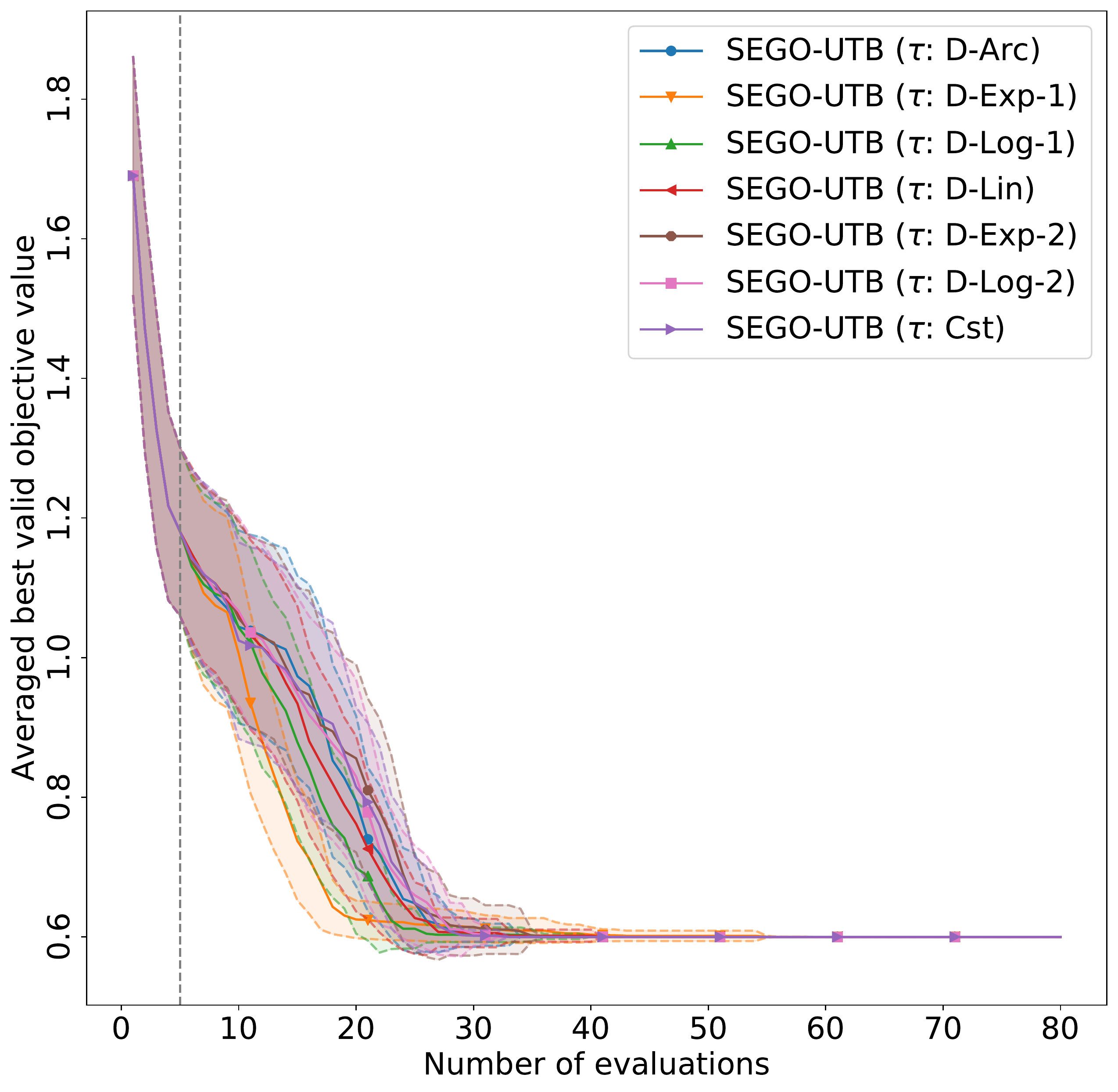}}
    \subfloat[Non-decreasing constraints learning rate strategies using $\epsilon_c=10^{-4}$. \label{fig:lsq_ad:inc4}]{\includegraphics[width=0.45\textwidth]{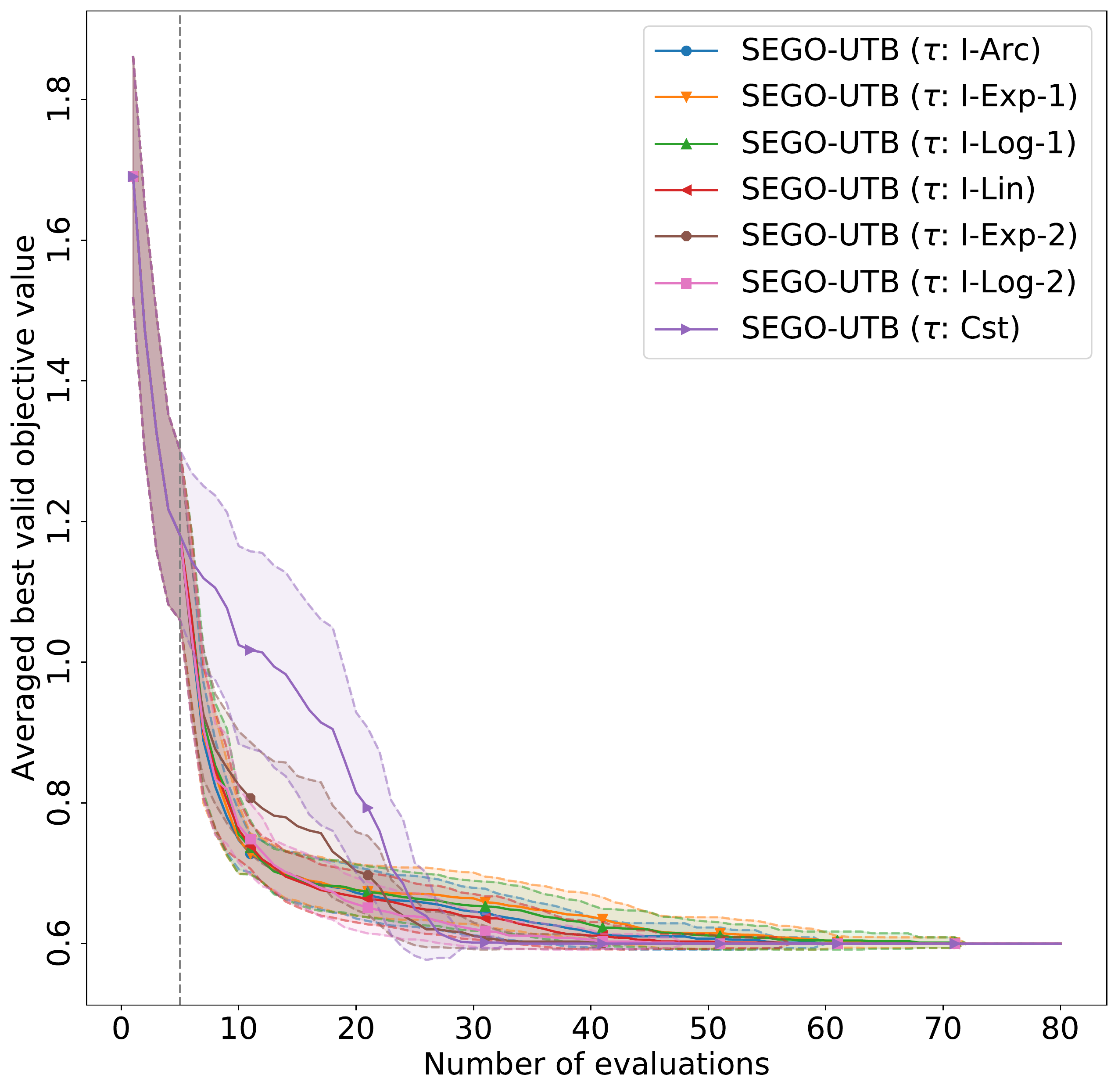}}
    
    \caption{Convergence plots for the LSQ problem, considering the two levels of constraints violation $10^{-4}$ and $10^{-2}$. The vertical grey-dashed line outlines the number of points in the initial DoEs.}
    \label{fig:lsq_ad}
\end{figure}

\change{SEGO-UTB ($\tau$: D-Exp-2) (resp. SEGO-UTB ($\tau$: I-Exp-2)) provides the best compromise on LSQ problem as implies by Figure~\ref{fig:lsq_ad}. Indeed, they both converge to the minimum value using the two levels of constraints violation and the standard deviation tend faster to zero than the other strategies.}

\begin{figure}[p]
    \centering
    
    \subfloat[Decreasing constraints learning rate strategies using $\epsilon_c=10^{-2}$. \label{fig:mb_ad:dec2}]{\includegraphics[width=0.45\textwidth]{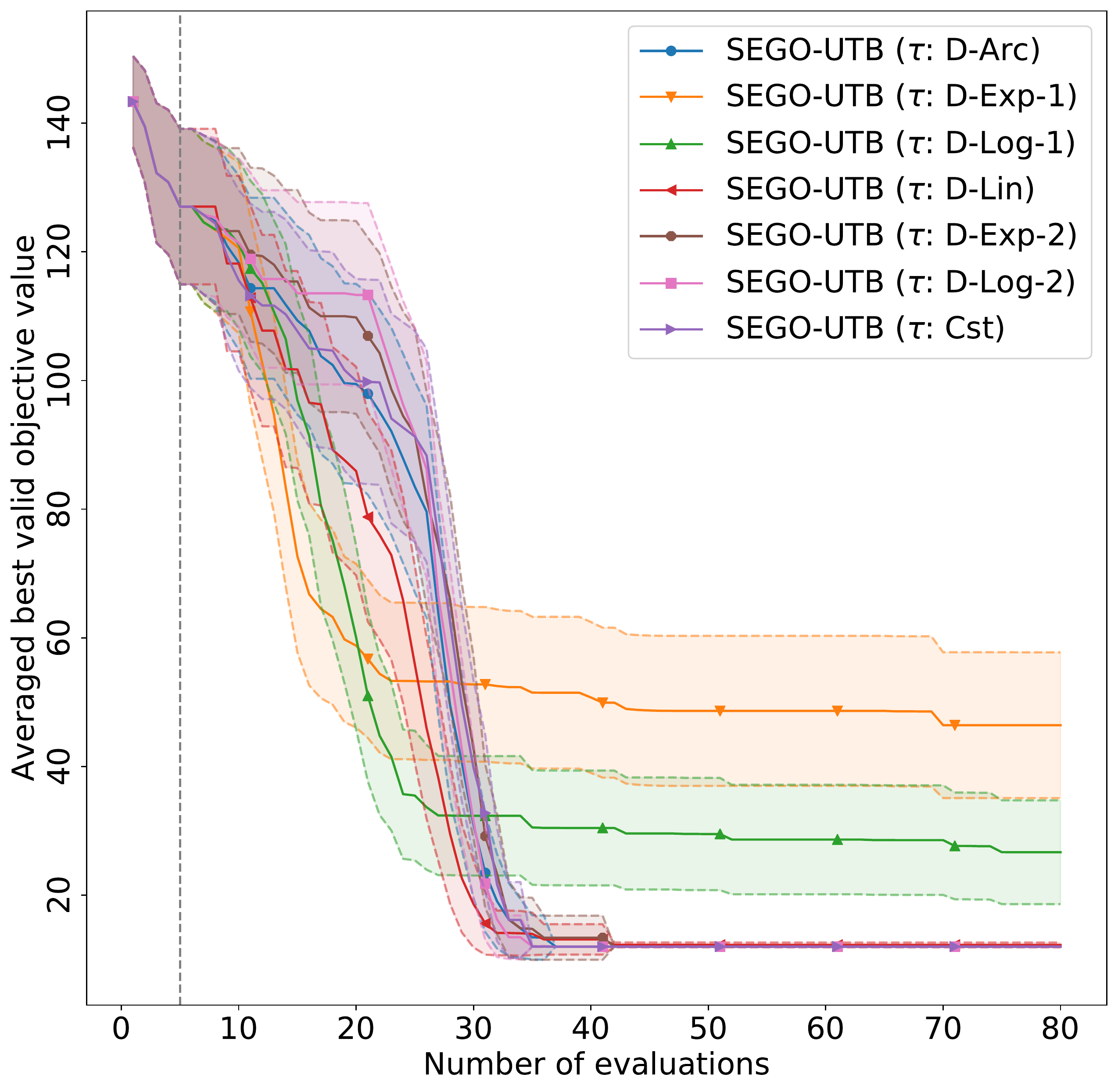}}
    \subfloat[Non-decreasing constraints learning rate strategies using $\epsilon_c=10^{-2}$. \label{fig:mb_ad:inc2}]{\includegraphics[width=0.45\textwidth]{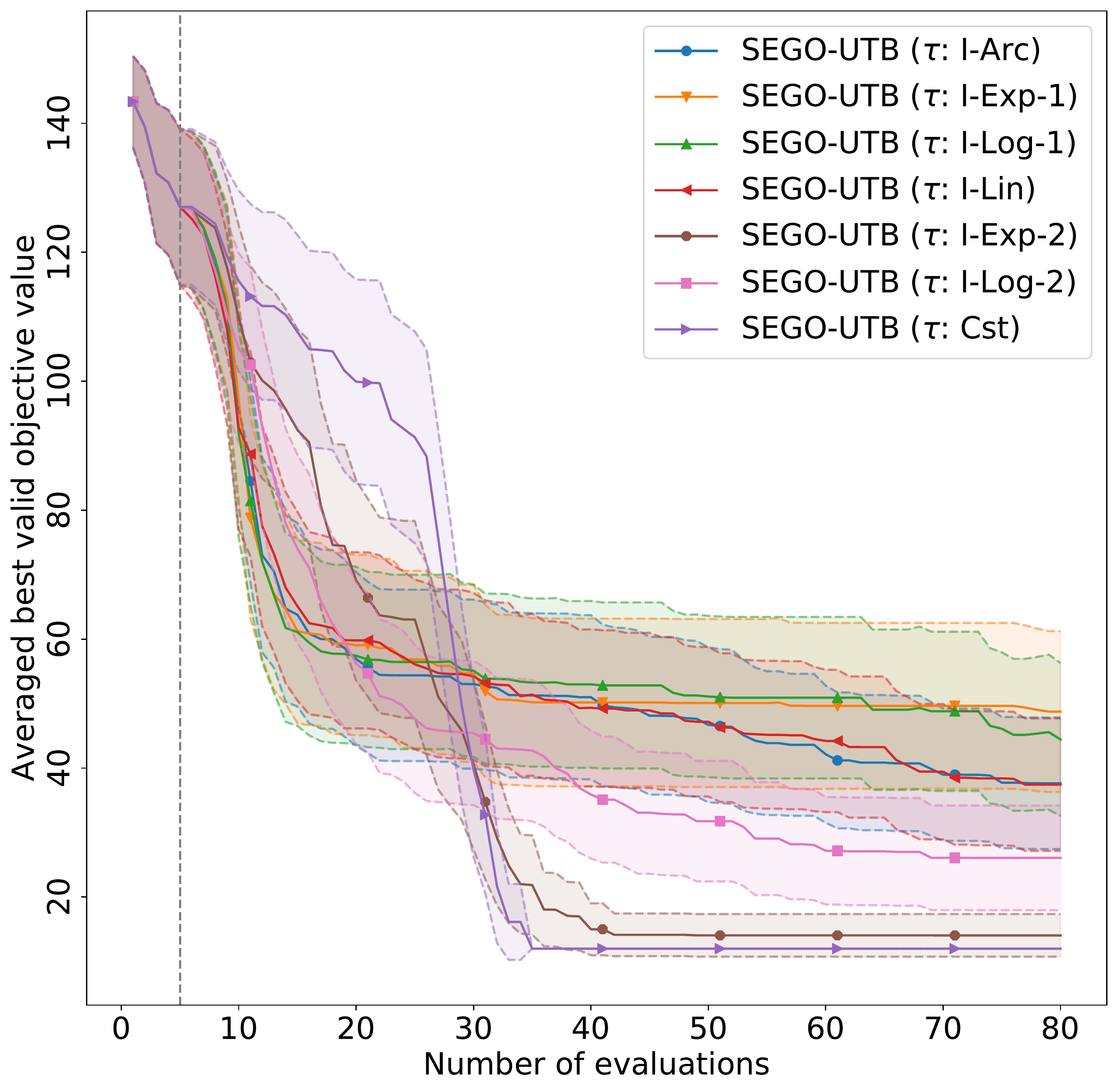}} \\
    
    \subfloat[Decreasing constraints learning rate strategies using $\epsilon_c=10^{-4}$. \label{fig:mb_ad:dec4}]{\includegraphics[width=0.45\textwidth]{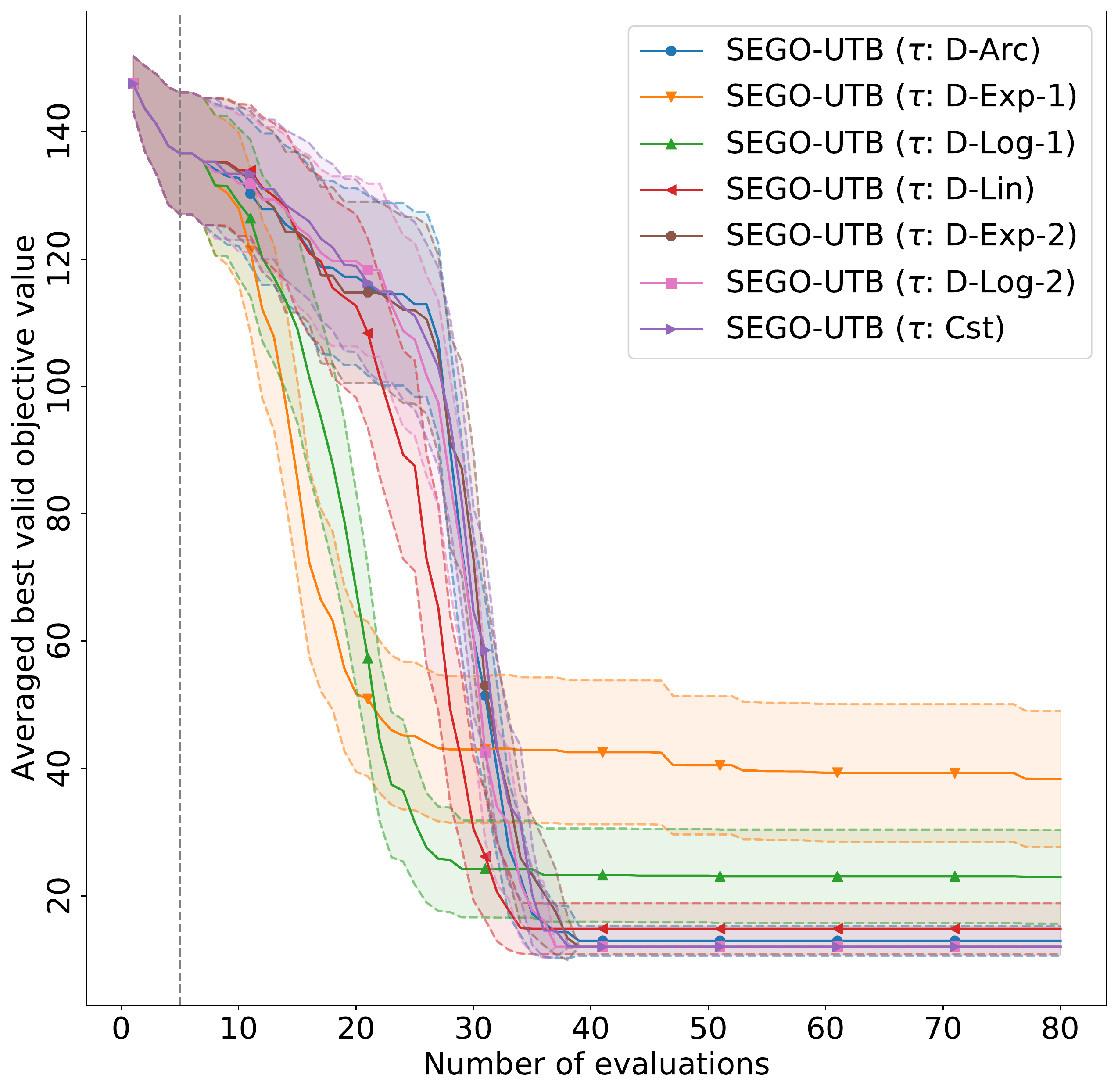}}
    \subfloat[Non-decreasing constraints learning rate strategies using $\epsilon_c=10^{-4}$. \label{fig:mb_ad:inc4}]{\includegraphics[width=0.45\textwidth]{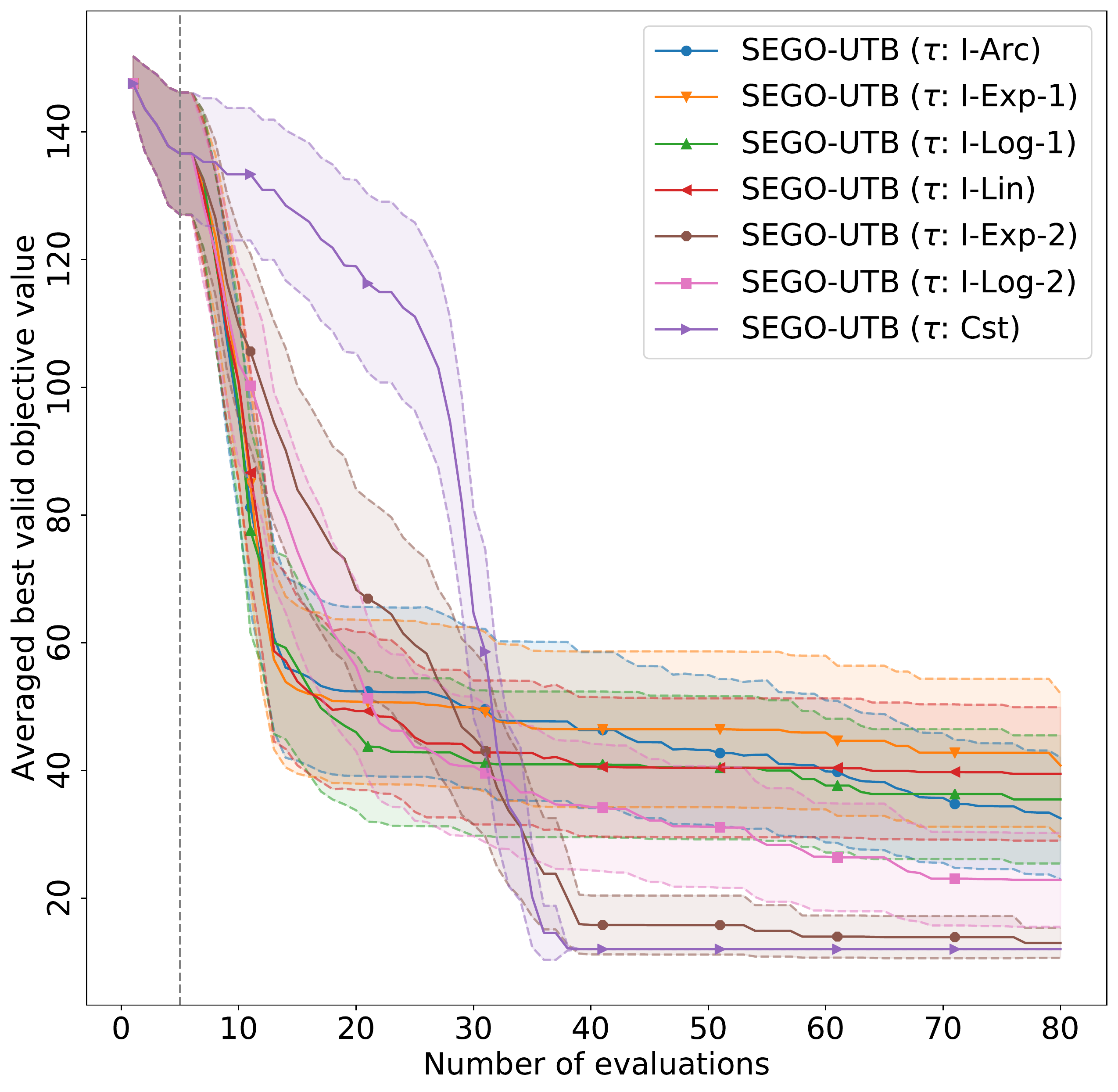}}
    
    \caption{Convergence plots for the MB problem, considering the two levels of constraints violation $10^{-4}$ and $10^{-2}$. The vertical grey-dashed line outlines the number of points in the initial DoEs.}
    \label{fig:mb_ad}
\end{figure}

\change{Figure~\ref{fig:mb_ad} provides the results for the non-decreasing and decreasing constraints learning rate strategies of SEGO-UTB for the MB problem using the two constraints violation $10^{-2}$ and $10^{-4}$. Again, SEGO-UTB ($\tau$: D-Exp-2) and SEGO-UTB ($\tau$: I-Exp-2) are a good compromise considering the two constraints violations. Note that none of the non-decreasing constraints learning rate are converging for the MB problem.}

\subsection{The 29 problem benchmark}

\begin{figure}[htb!]
    \centering
    \subfloat[Decreasing constraints learning rate strategies using $\epsilon_c=10^{-2}$. \label{fig:dp_ad:dec2}]{\includegraphics[width=0.45\textwidth]{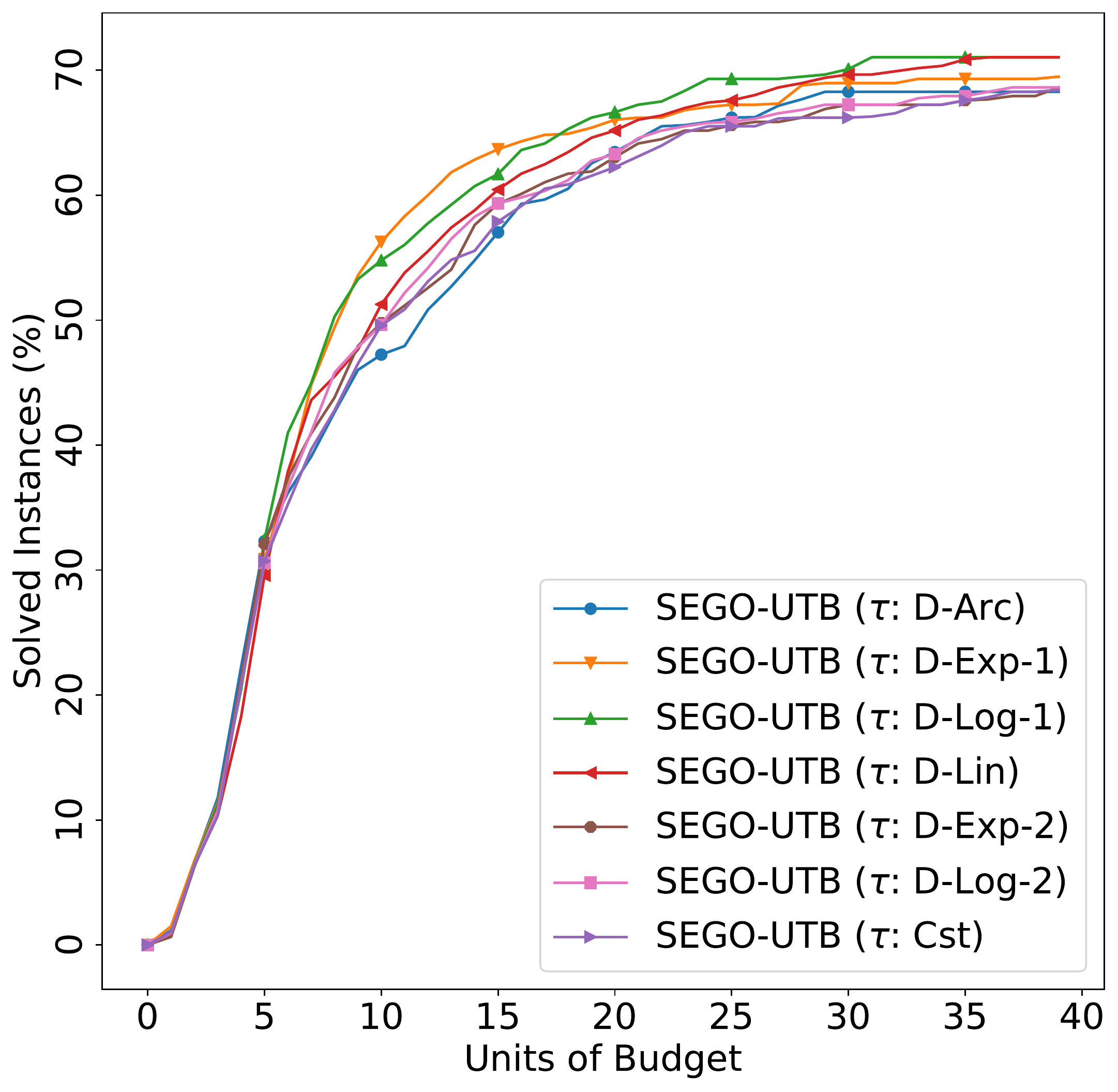}}
    \subfloat[Non-decreasing constraints learning rate strategies using $\epsilon_c=10^{-2}$. \label{fig:dp_ad:inc2}]{\includegraphics[width=0.45\textwidth]{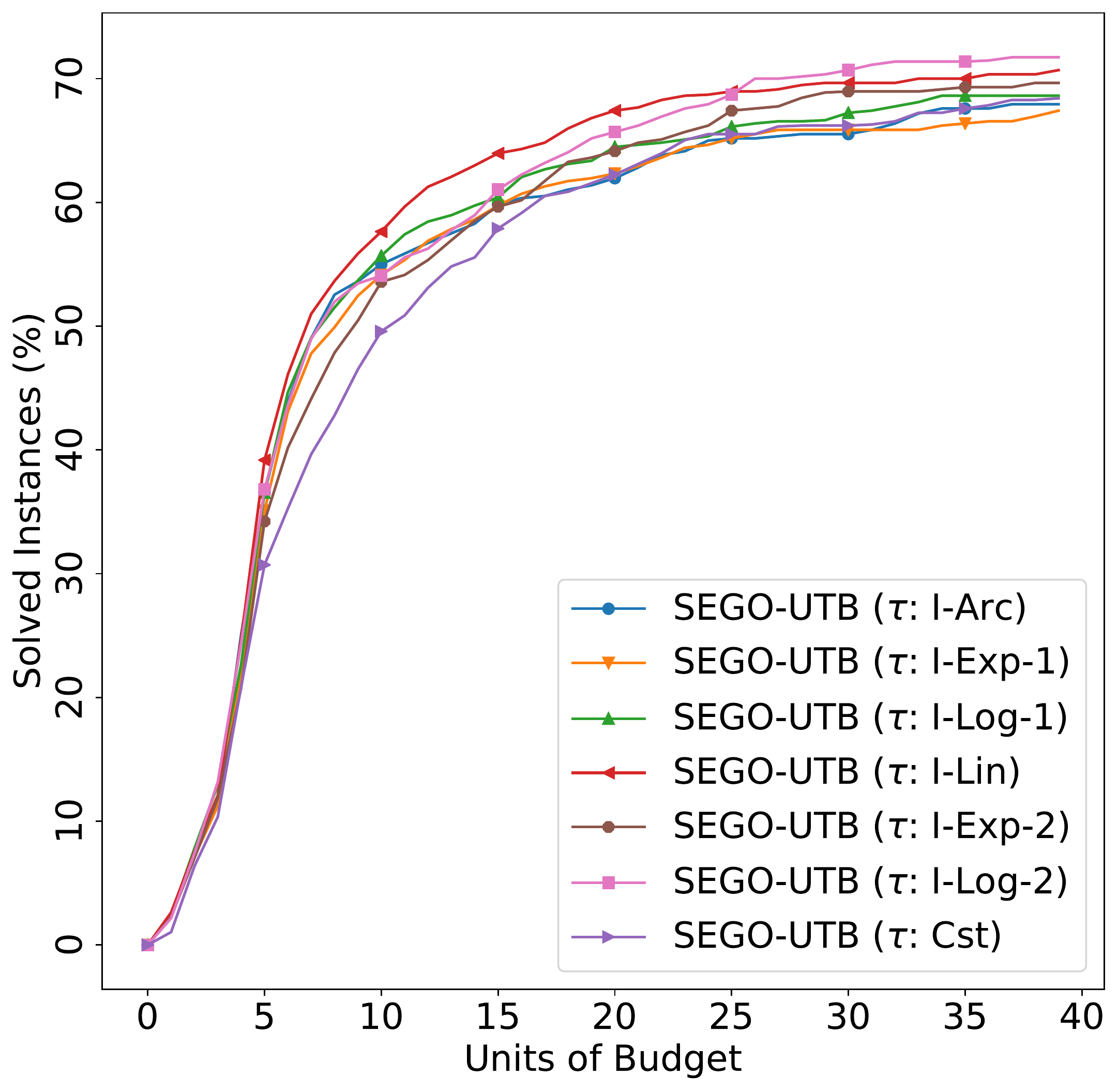}} \\
    
    \subfloat[Decreasing constraints learning rate strategies using $\epsilon_c=10^{-4}$. \label{fig:dp_ad:dec4}]{\includegraphics[width=0.45\textwidth]{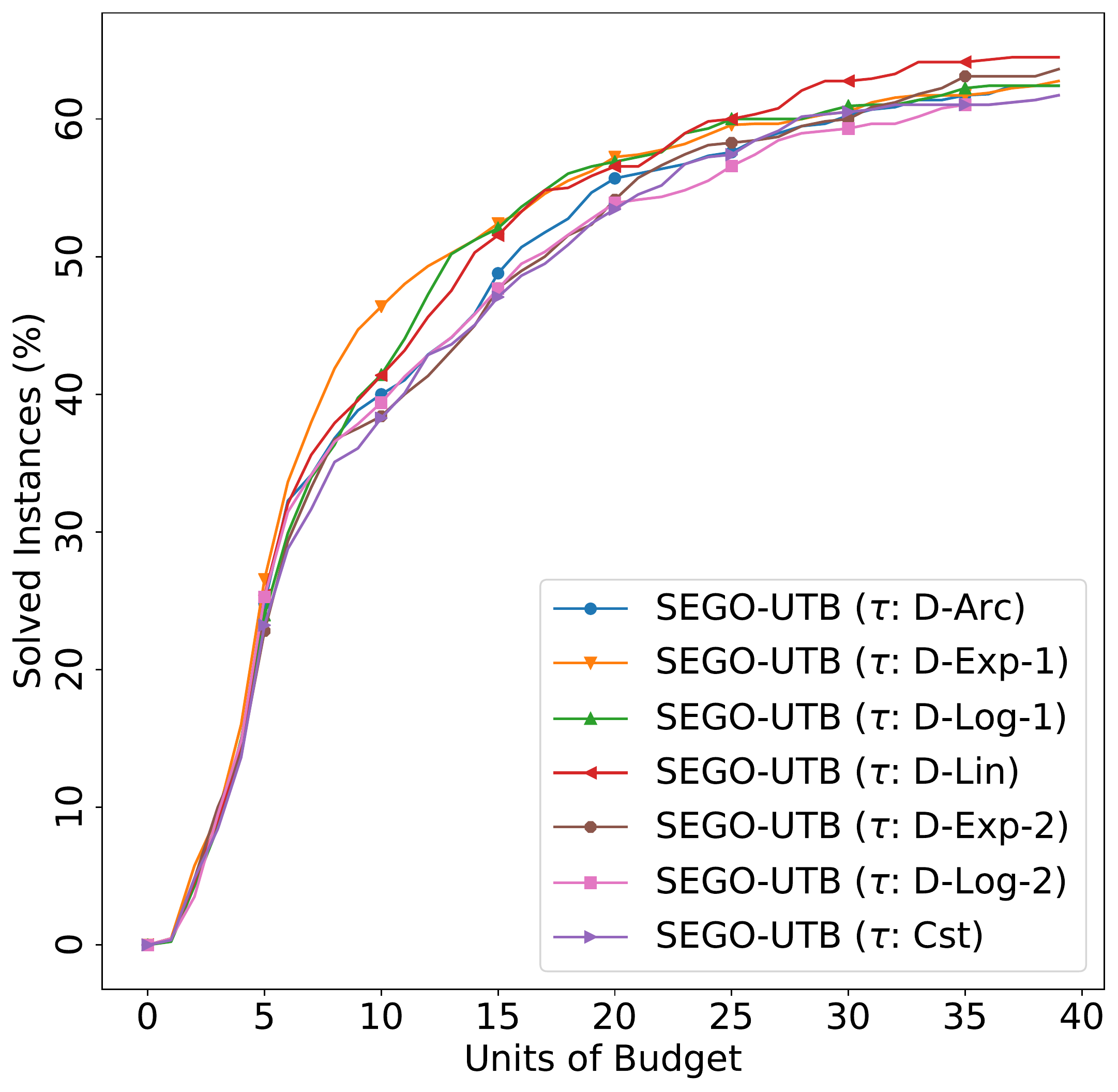}}
    \subfloat[Non-decreasing constraints learning rate strategies using $\epsilon_c=10^{-4}$. \label{fig:dp_ad:inc4}]{\includegraphics[width=0.45\textwidth]{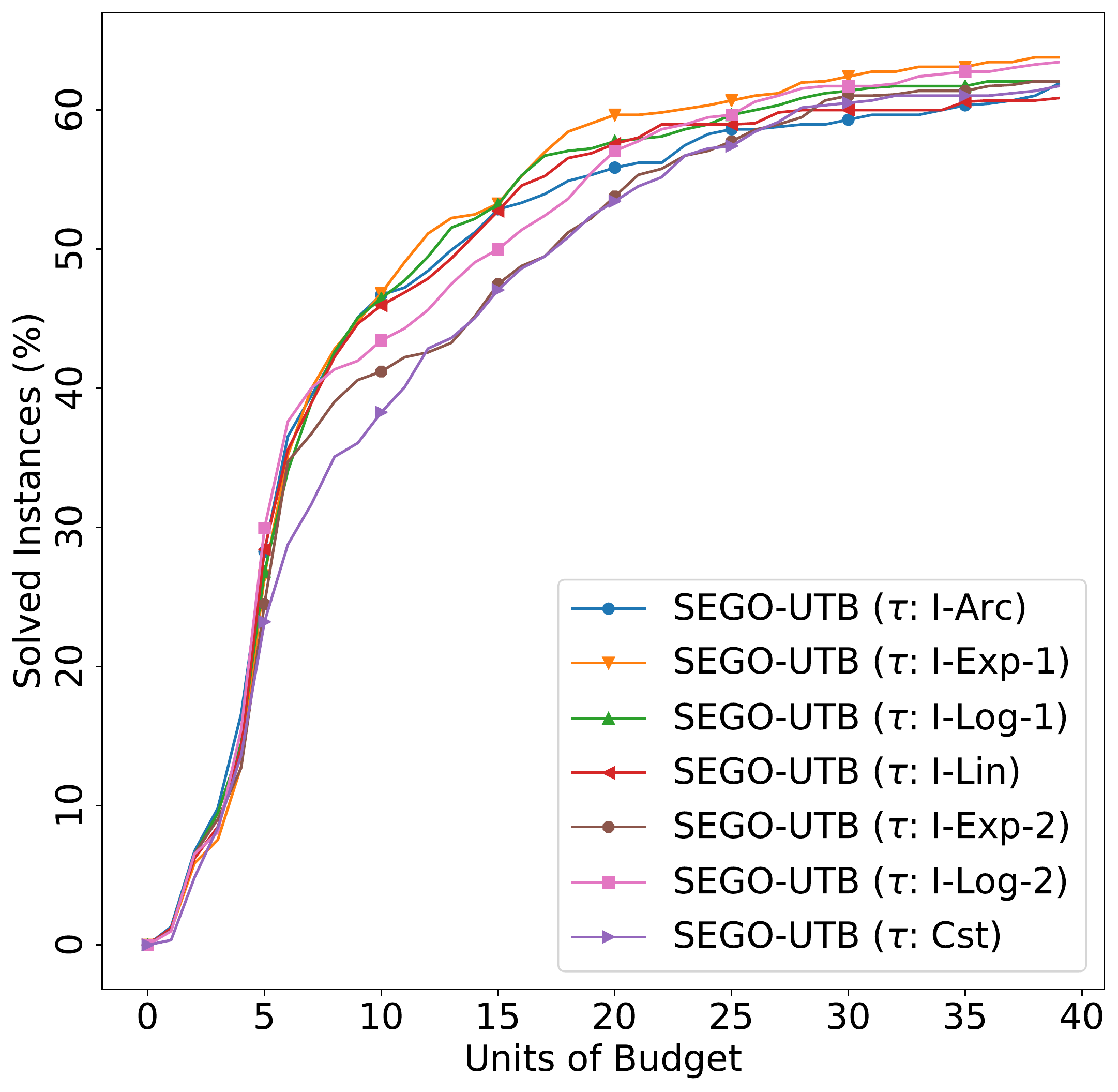}}
    \caption{Data Profiles for all the tested problems, considering the two levels of constraints violation $10^{-4}$ and $10^{-2}$.}
    \label{fig:dp_ad}
\end{figure}

\change{Figure \ref{fig:dp_ad} shows the data profiles for the non-decreasing and decreasing constraints learning rate strategies using the two levels of constraints violation $10^{-2}$ and $10^{-4}$. Concerning the decreasing strategies, one can clearly see that SEGO-UTB ($\tau$: D-Exp-1) is the best compromise as it solve 55\% (resp. 65\%) of the instances using $\epsilon_c=10^{-2}$ (resp. $\epsilon_c=10^{-4}$). SEGO-UTB ($\tau$: I-Log-1) offers the best performances on the both constraints violation for the non decreasing strategies.} 

\subsubsection{The FAST test case}

\begin{figure}[htb!]
    \centering
    \subfloat[Decreasing constraints learning rate strategies. \label{fig:fast_ad:dec}]{\includegraphics[width=0.45\textwidth]{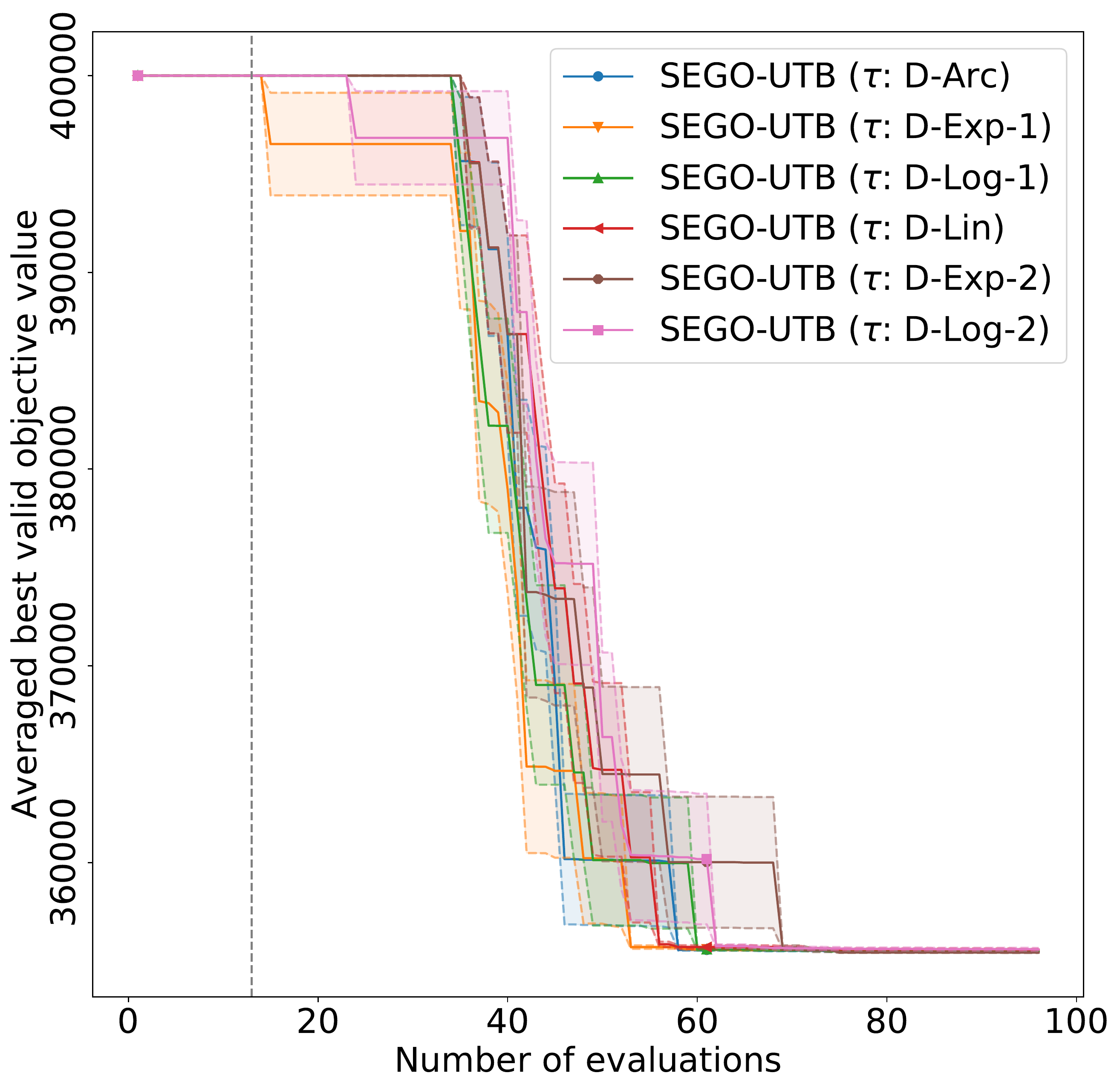}}
    \subfloat[Non-decreasing constraints learning rate strategies. \label{fig:fast_ad:inc}]{\includegraphics[width=0.45\textwidth]{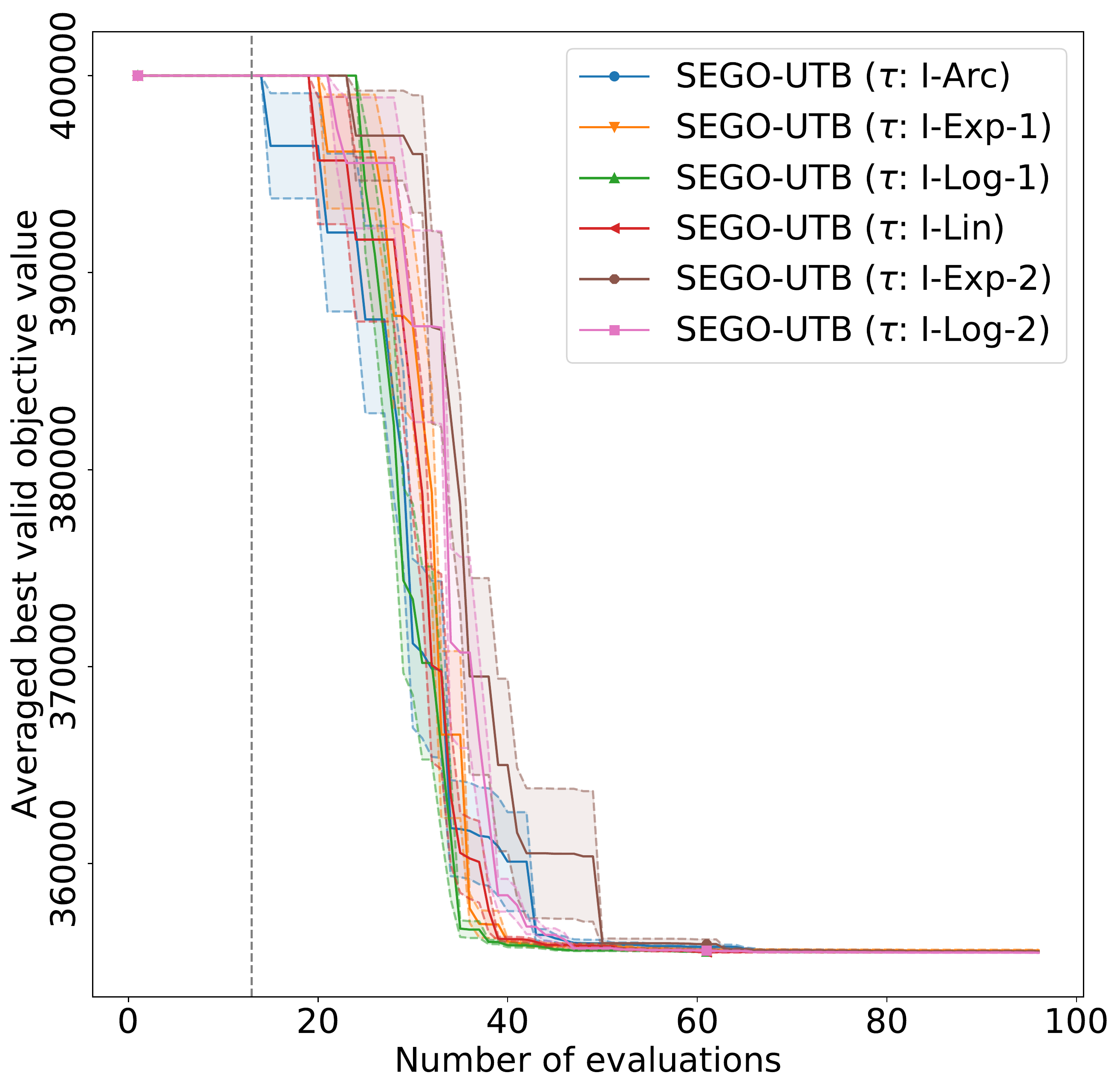}}
    \caption{Convergence plots for the FAST problem. The vertical grey-dashed line outlines the number of points in the initial DoEs.}
    \label{fig:fast_ad}
\end{figure}

For the FAST test case, the non-decreasing and decreasing constraints learning rates performance are given by Figure~\ref{fig:fast_ad}.
We notice that SEGO-UTB ($\tau$: D-Exp-1) is performing the best for the decreasing behaviors as it converges the fastest to the minimum value. 
For the non-decreasing behaviors, SEGO-UBT ($\tau$: I-Log-1) is providing the best results. 

\begin{figure}[p]
    \bigcentering
    \subfloat[SEGO \label{fig:pargraph:MC}]{\includegraphics[height=0.85\textheight]{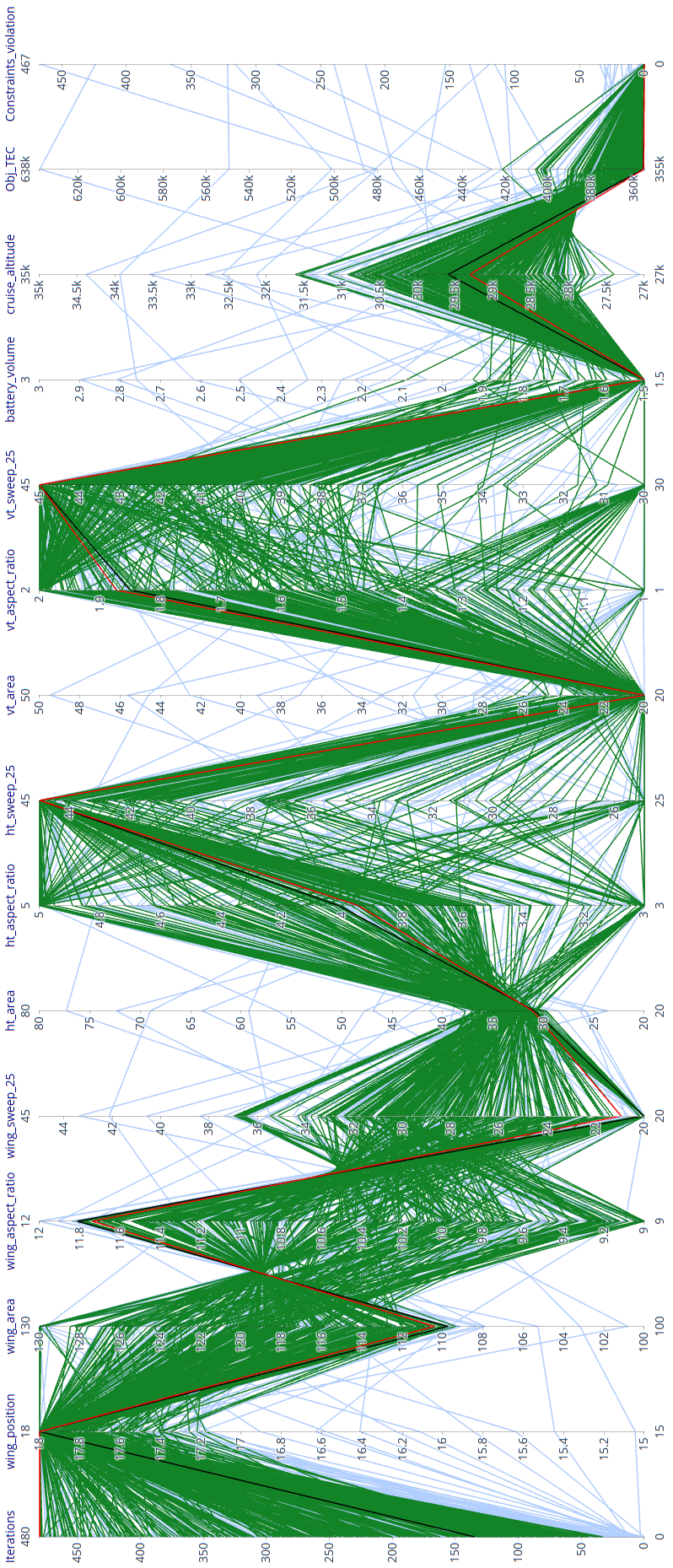}}
    \subfloat[SEGO-UTB ($\tau$: Cst) \label{fig:pargraph:C}]{\includegraphics[height=0.85\textheight]{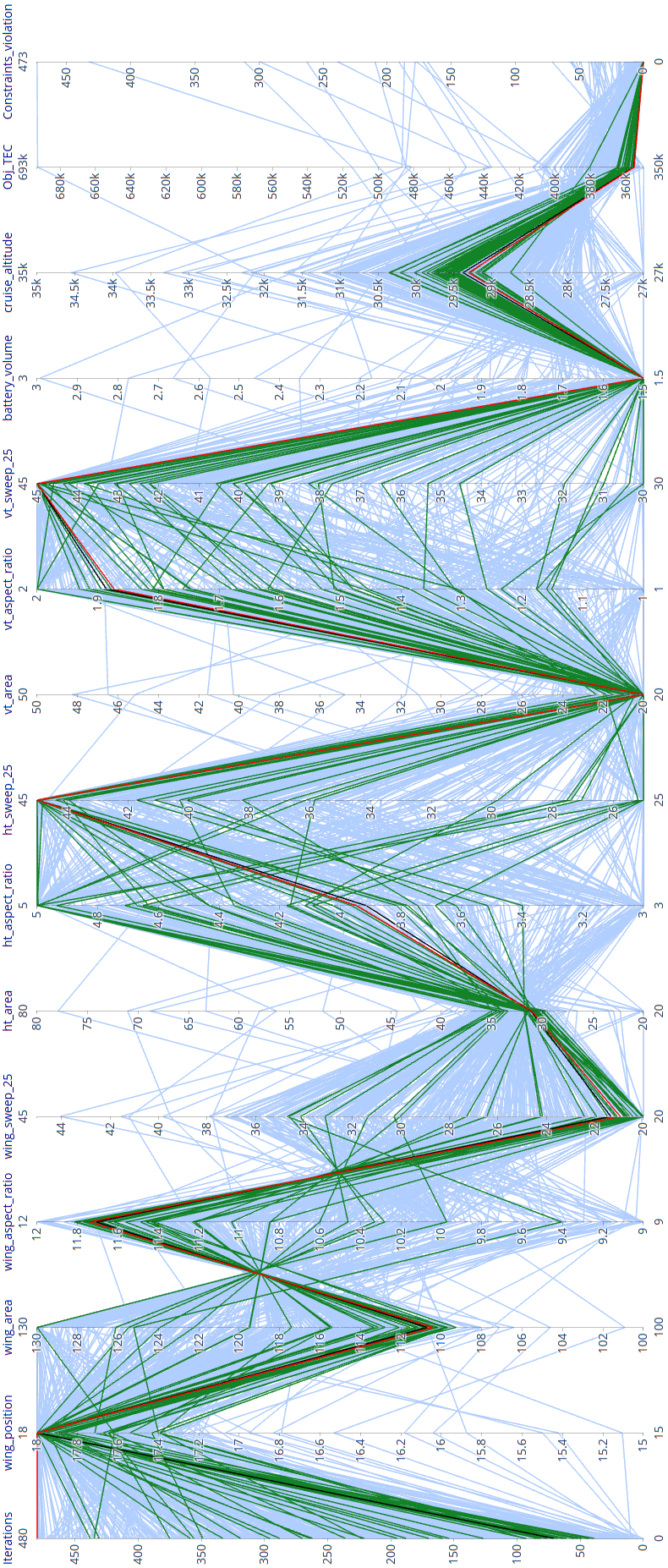}}
    \caption{Parallel plots of the median runs for FAST problem. In grey: the designs outside of the design space; in blue: the unfeasible designs; in green: the feasible designs; in black: the optimum; in red: the reference design.}
    \label{fig:pargraph:other1}
\end{figure}

\begin{figure}[p]
    \bigcentering
    \subfloat[SEGO-UTB ($\tau$: D-Exp-1) \label{fig:pargraph:DE}]{\includegraphics[height=0.85\textheight]{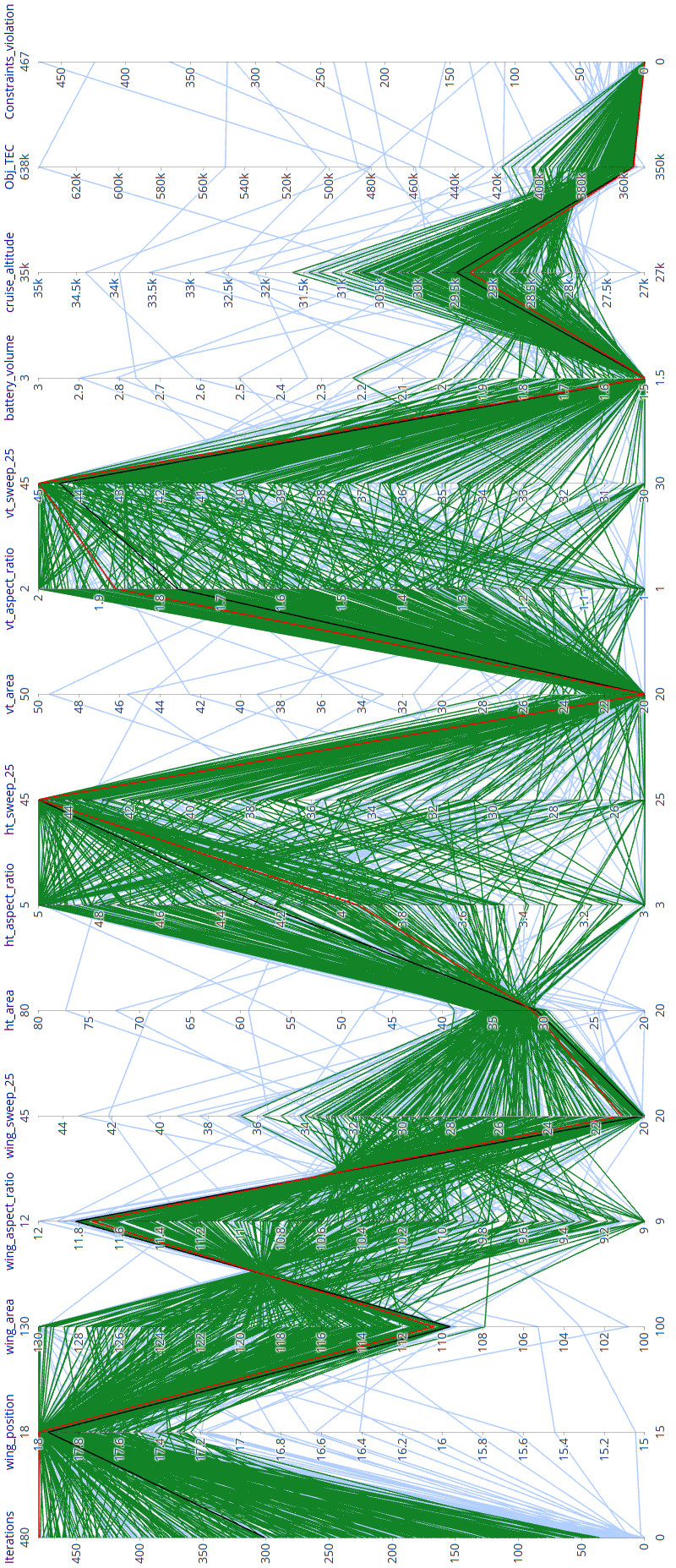}}
    \subfloat[NOMAD \label{fig:pargraph:Nomad}]{\includegraphics[height=0.85\textheight]{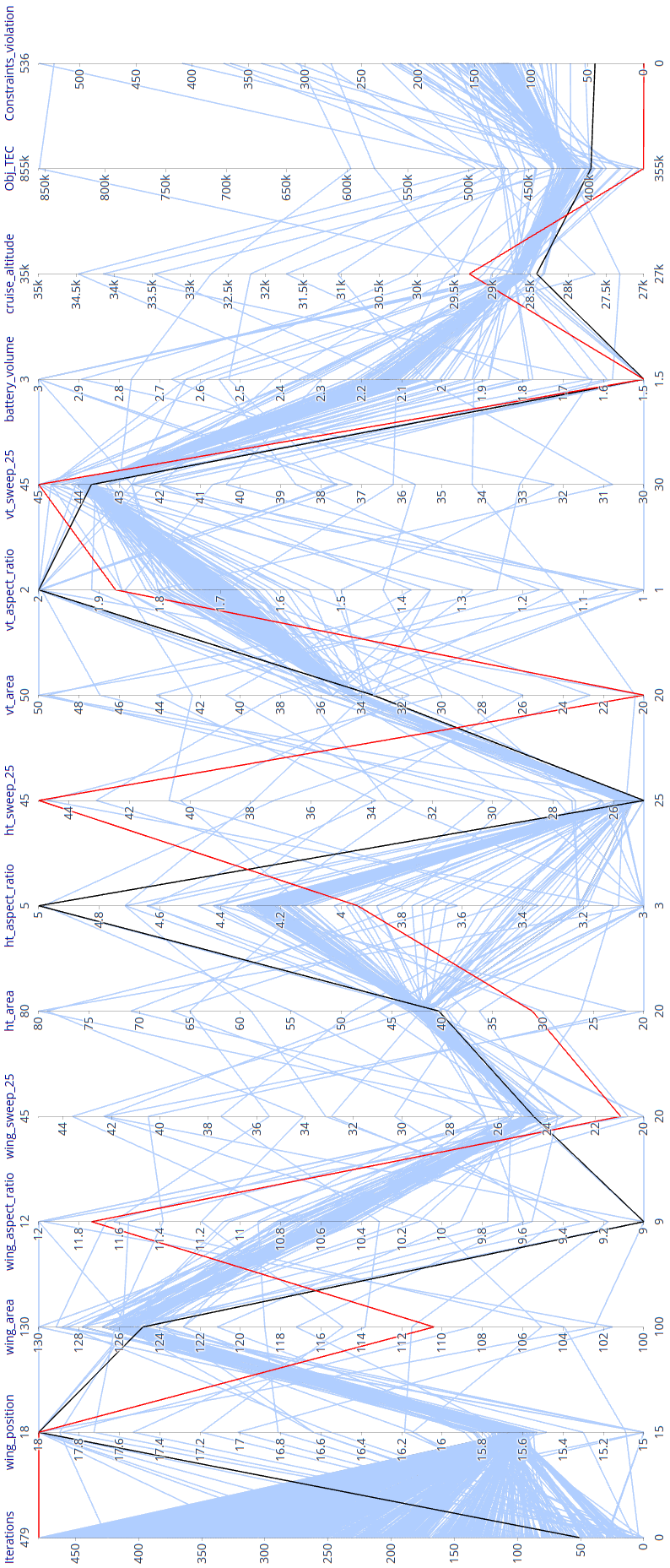}}
    \caption{Parallel plots of the median runs for FAST problem. In grey: the designs outside of the design space; in blue: the unfeasible designs; in green: the feasible designs; in black: the optimum; in red: the reference design.}
    \label{fig:pargraph:other2}
\end{figure}

Last, the parallel plots of the median run of SEGO, SE\change{G}O-UTB ($\tau$: Cst), SE\change{G}O-UTB ($\tau$: D-Exp-1) and NOMAD are introduced in Figure\change{s}~\ref{fig:pargraph:other1} and \ref{fig:pargraph:other2}." \ref{fig:pargraph:other2}.

First, NOMAD is not able to find any feasible point as implied by Figure~\ref{fig:pargraph:Nomad} and seems to get stuck in a zone with a violation around 150.
Then, SEGO-UTB ($\tau$: Cst), drawn in Figure~\ref{fig:pargraph:C}, highlights the exploration behavior of this constraints learning rate with an important number of unfeasible designs all along the optimization. 
This behavior allows the optimizer to detect the best feasible design which is very close to the reference optimum. 
Figure~\ref{fig:pargraph:MC} lastly displays that SEGO focuses on the exploitation of the constraints as demonstrated by the large number of feasible design points.
The optimum is thus quickly obtain.

\end{document}